\documentclass{article}
\usepackage{iclr2022_conference,times}

\usepackage{amsmath,amsfonts,bm}

\def\eqref#1{equation~\ref{#1}}

\def\floor#1{\lfloor #1 \rfloor}
\def\1{\bm{1}}

\DeclareMathAlphabet{\mathsfit}{\encodingdefault}{\sfdefault}{m}{sl}
\SetMathAlphabet{\mathsfit}{bold}{\encodingdefault}{\sfdefault}{bx}{n}

\usepackage[T5,T3,T1]{fontenc}

\usepackage[utf8]{inputenc}
\usepackage[mathletters]{ucs}
\usepackage{hyperref}
\usepackage{url}
\usepackage{graphicx}
\usepackage{tabularx} 
\usepackage{wrapfig}
\usepackage{booktabs}
\usepackage{multirow}
\usepackage{xspace}
\usepackage{tablefootnote}
\usepackage{caption}
\usepackage{subcaption}
\usepackage{mathtools}
\usepackage{algorithm}
\usepackage{algpseudocode}
\usepackage{inconsolata}
\usepackage{euflag}
\usepackage{wrapfig}
\usepackage{pifont}
\newcommand{\cmark}{\ding{51}}%
\newcommand{\xmark}{\ding{55}}%

\definecolor{darkblue}{rgb}{0, 0, 0.5}
\definecolor{gr}{HTML}{363636}

\usepackage{colortbl}
\colorlet{tableheadcolor}{gray!85}
\colorlet{tablerowcolor}{gray!20}
\newcommand{\rowcol}{\rowcolor{tablerowcolor}} %
\newcommand{\tc}{\textcolor{tableheadcolor}} %

\algdef{SE}{Input}{Output}{\textbf{Input: }}{\textbf{Output: }}

\usepackage{siunitx}
\newcommand{\copenhagen}{1}
\newcommand{\rff}{2}
\newcommand{\johnshopkins}{3}

\newcommand{\leuven}{5}
\newcommand{\pioneer}{6}

\title{Language Modelling with Pixels}

\author{Phillip Rust$^{\copenhagen}$ ~ Jonas F. Lotz$^{\copenhagen,\rff}$ ~ Emanuele Bugliarello$^{\copenhagen}$ \\ \textbf{Elizabeth Salesky}$^{\johnshopkins}$ ~ \textbf{Miryam de Lhoneux}$^{\leuven}$ ~ \textbf{Desmond Elliott}$^{\copenhagen,\pioneer}$\\
  $^{\copenhagen}$University of Copenhagen \quad
  $^{\rff}$ROCKWOOL Foundation Research Unit \\
  $^{\johnshopkins}$Johns Hopkins University\quad
  $^{\leuven}$KU Leuven \quad 
  $^{\pioneer}$Pioneer Centre for AI\\
\texttt{p.rust@di.ku.dk}
}

\newcommand{\model}{\textsc{pixel}\xspace}

\newcommand{\arabi}{\textsc{ara}\xspace}
\newcommand{\english}{\textsc{eng}\xspace}
\newcommand{\coptic}{\textsc{cop}\xspace}
\newcommand{\japanese}{\textsc{jpn}\xspace}
\newcommand{\hindi}{\textsc{hin}\xspace}
\newcommand{\korean}{\textsc{kor}\xspace}
\newcommand{\vietnamese}{\textsc{vie}\xspace}
\newcommand{\chinese}{\textsc{zho}\xspace}
\newcommand{\tamil}{\textsc{tam}\xspace}
\newcommand{\amharic}{\textsc{amh}\xspace}
\newcommand{\hausa}{\textsc{hau}\xspace}
\newcommand{\igbo}{\textsc{ibo}\xspace}
\newcommand{\kinyarwanda}{\textsc{kin}\xspace}
\newcommand{\luganda}{\textsc{lug}\xspace}
\newcommand{\luo}{\textsc{luo}\xspace}
\newcommand{\naija}{\textsc{pcm}\xspace}
\newcommand{\swahili}{\textsc{swa}\xspace}
\newcommand{\wolof}{\textsc{wol}\xspace}
\newcommand{\yoruba}{\textsc{yor}\xspace}
\newcommand{\french}{\textsc{fra}\xspace}
\newcommand{\spanish}{\textsc{spa}\xspace}
\newcommand{\german}{\textsc{deu}\xspace}
\newcommand{\bulgarian}{\textsc{bul}\xspace}
\newcommand{\greek}{\textsc{ell}\xspace}
\newcommand{\russian}{\textsc{rus}\xspace}
\newcommand{\thai}{\textsc{tha}\xspace}
\newcommand{\turkish}{\textsc{tur}\xspace}
\newcommand{\urdu}{\textsc{urd}\xspace}
\newcommand{\bengali}{\textsc{ben}\xspace}
\newcommand{\finnish}{\textsc{fin}\xspace}
\newcommand{\indonesian}{\textsc{ind}\xspace}
\newcommand{\telugu}{\textsc{tel}\xspace}

\newcommand{\modernarabic}{\textsc{msa}\xspace}
\newcommand{\egyptarabic}{\textsc{ea}\xspace}

\newcommand{\circa}{{\raise.17ex\hbox{$\scriptstyle\sim$}}}

\makeatletter
\g@addto@macro{\UrlBreaks}{\UrlOrds}
\makeatother

\iclrfinalcopy
\begin{document}

\maketitle

\begin{abstract}

Language models are defined over a finite set of inputs, which creates a \emph{vocabulary bottleneck} when we attempt to scale the number of supported languages. Tackling this bottleneck results in a trade-off between what can be represented in the embedding matrix and computational issues  in the output layer. This paper introduces \model, the \textbf{Pix}el-based \textbf{E}ncoder of \textbf{L}anguage, which suffers from neither of these issues. \model is a pretrained language model that renders text as images, making it possible to transfer representations across languages based on orthographic similarity or the co-activation of pixels. \model is trained to reconstruct the pixels of masked patches instead of predicting a distribution over tokens.\footnote{See Appendix~\ref{app:abstract_reconstructions} for reconstructions of this abstract.} We pretrain the $86$M parameter \model model on the same English data as \textsc{bert} and evaluate on syntactic and semantic tasks in typologically diverse languages, including various non-Latin scripts.
We find that \model substantially outperforms \textsc{bert} on syntactic and semantic processing tasks on scripts that are not found in the pretraining data, but \model is slightly weaker than \textsc{bert} when working with Latin scripts. 
Furthermore, we find that \model is more robust than \textsc{bert} to orthographic attacks and linguistic code-switching, further confirming the benefits of modelling language with pixels.

\end{abstract}

\section{Introduction}
\label{sec:intro}
\vspace{-2mm}

Natural language processing has rapidly progressed in recent years due to a combination of self-supervised representation learning, i.e.\ pretrained language models (PLMs) like \textsc{bert} \citep{devlin-etal-2019-bert}, GPT-3 \citep{brown-etal-2020-language}, and XLM-R \citep{conneau-etal-2020-unsupervised}; large unlabelled datasets; such as C4 \citep{raffel-etal-2020-t5}, The Pile \citep{gao2020pile}; and large-scale computing power \citep{hirschberg2015advances}.
Despite this progress, these models only cover a fraction of the world's languages, with large inequalities in performance \citep{pires-etal-2019-multilingual,lauscher-etal-2020-zero},   
and the majority of languages are falling behind English 
\citep{joshi-etal-2020-state,bugliarello-etal-2022-iglue}. Even within English, these models struggle when tasked with processing noisy inputs \citep{SunADV-BERT, eger-benz-2020-hero}.
In this paper, we show how to effectively support \textit{thousands} of written languages in a single model while being robust to variations caused by character-level noise.

Language models typically support a finite vocabulary of categorical inputs, e.g.\ characters, subwords or even words, and much effort has been devoted to vocabulary construction~\citep{wan2022fairness}.
On one end of the spectrum, a vocabulary over words has three problems: (i) it is not possible to encode out-of-vocabulary words because they lack an entry in a closed vocabulary, e.g.\ ``doxing'', (ii) there are too many parameters in the word embedding layer, and relatedly, (iii) the normalising constant for the softmax activation in the output layer is too expensive to compute. 
On the other end of the spectrum, vocabularies over bytes or characters are much smaller, which leads to increased sequence lengths \citep{keren-etal-2022-breaking}. 
In practice, most current models operate over inputs smaller than words but larger than characters: subword units \citep{sennrich-etal-2016-neural,kudo-2018-subword}. 
Subwords prevent the problem of extremely large embedding and output layers, and support open vocabulary processing. 
While this is a practical solution in a monolingual context and for some languages like English, dealing with many languages with a variety of scripts will either result in a very large vocabulary or a trade-off over what is represented within a fixed number of subwords (see \S \ref{sec:related_work}). 
Taken together, given a language model with a finite vocabulary, there is a bottleneck in two locations: at the level of the encoding of the inputs and at the level of estimating the probability distribution over the vocabulary. 
We call this the \textit{vocabulary bottleneck}. A language model that can handle thousands of languages needs to deal with this problem.

We propose to rethink language modelling as a visual recognition task, removing the need for a finite vocabulary. 
Our proposal is inspired by \citet{salesky-etal-2021-robust}, who showed how to train a machine translation model with ``visual text representations'' in the encoder instead of subwords.
Our \textbf{Pix}el-based \textbf{E}ncoder of \textbf{L}anguage (\textbf{\model}) is built on the Masked Autoencoding Visual Transformer \citep[ViT-MAE; ][]{he-etal-2022-mae}. ViT-MAE is a Transformer-based encoder-decoder trained to reconstruct the pixels in masked image patches.
\model does not have a vocabulary embedding layer; instead, text is rendered as a sequence of fixed-sized patches, which are processed using a Vision Transformer encoder \citep{dosovitskiy2021an}.
\model also does not have an expensive output layer
when it reconstructs the pixels of the masked patches. In effect, \model provides a solution to the vocabulary bottleneck without needing the prohibitively long sequences of character-based models. 

\model is pretrained on the same data as \textsc{bert}, given our computational resources. This means that it has encountered only \circa0.05\% non-English text \citep{blevins-zettlemoyer-2022-language}.\footnote{We do not claim that a language model designed to support thousands of languages should be pretrained only on English text. 
We expect that pretraining on an appropriate choice of another language or multilingually may provide more remarkable results. \model represents an initial effort at smaller scale.
}
We evaluate \model on a range of syntactic and semantic tasks in 32 typologically diverse languages across 14 scripts, showing that it can rapidly adapt to new languages and unseen scripts. 
\model is also evaluated on its ability to handle noisy text caused by orthographic attacks, where pixel-based encoding is a clear improvement over subword-based vocabularies. In lexical code-switching experiments, \model performs on-par with \textsc{bert} and sometimes outperforms the multilingually pretrained \textsc{mBert}.

\model is a new type of language model that can theoretically support any language that can be typeset by a modern computer. We make the implementation, the pretrained model including intermediate training checkpoints, and the fine-tuned models freely available for the community.\footnote{\url{https://github.com/xplip/pixel}}

\begin{figure*}[t]
    \centering
    \begin{subfigure}[b]{0.48\textwidth}\centering
    \includegraphics[width=0.85\textwidth]{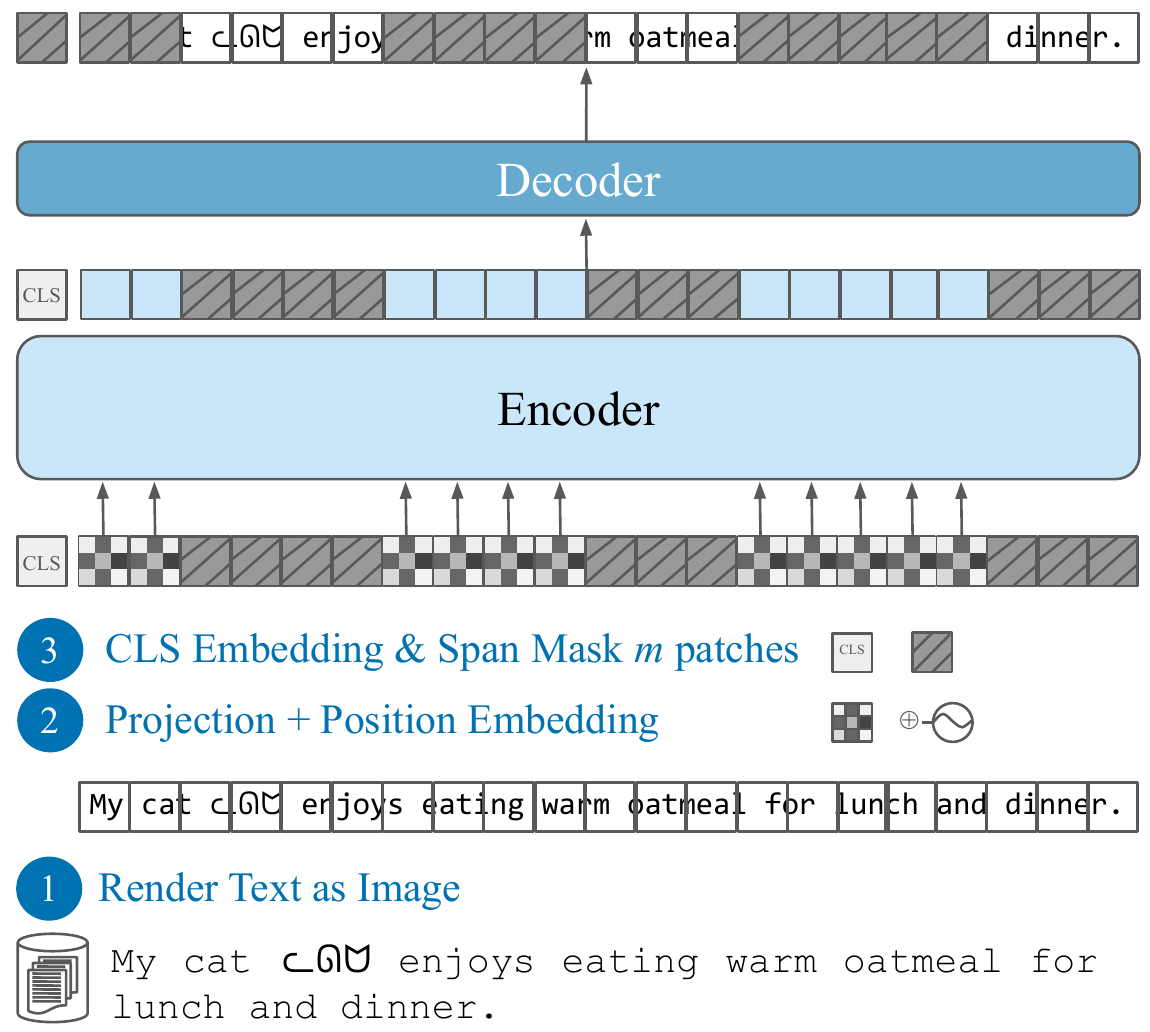}
    \caption{\model pretraining}
    \end{subfigure}
    \quad
    \begin{subfigure}[b]{0.48\textwidth}\centering
    \includegraphics[width=0.85\textwidth]{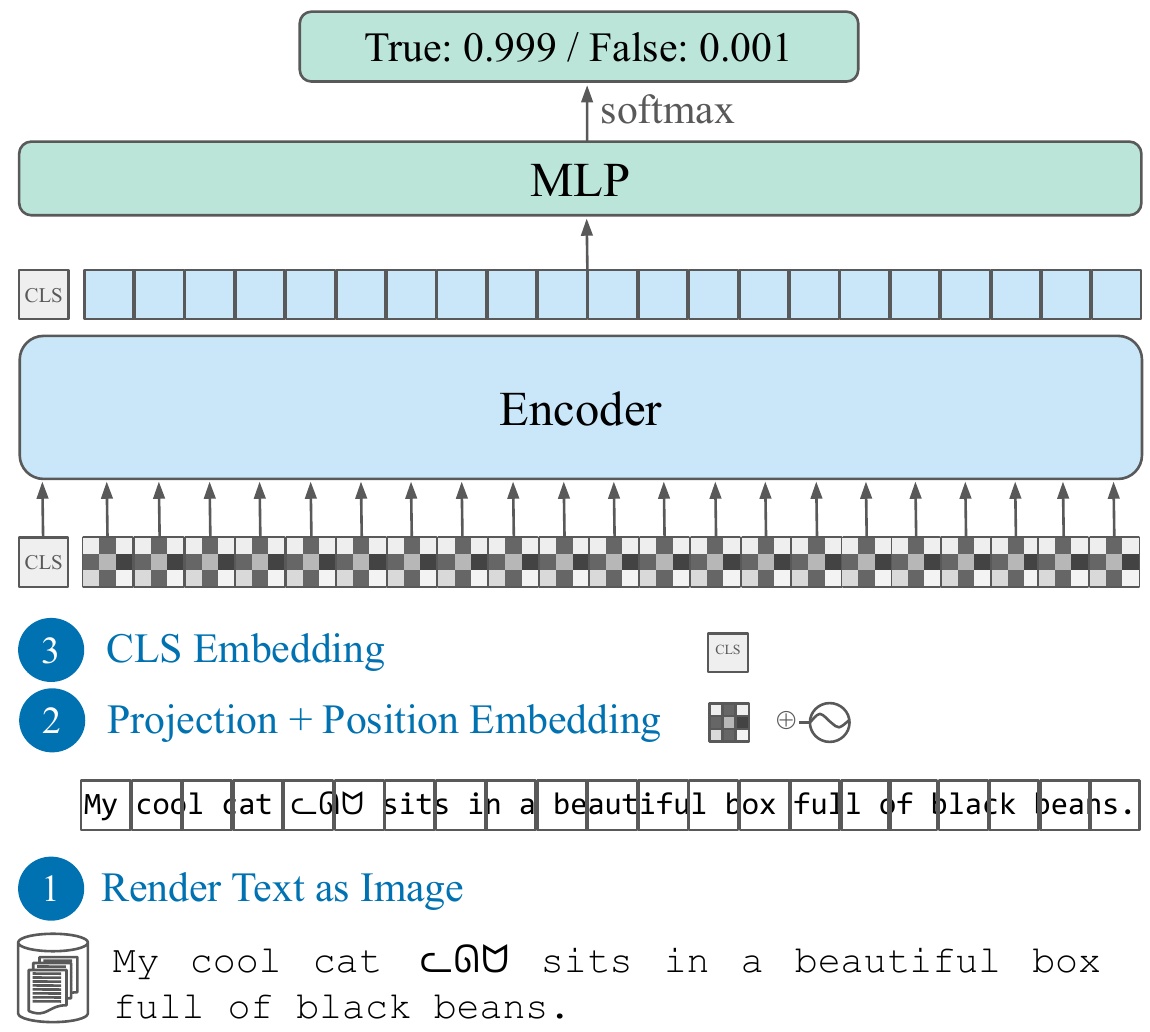}
    \caption{\model finetuning}
    \end{subfigure}
    \vspace{-2mm}
    \caption{Overview of \model's architecture. Following \cite{he-etal-2022-mae}, we use a masked autoencoder with a ViT architecture and a lightweight decoder for pretraining (left). At finetuning time (right), the decoder is replaced by a task-specific classification head that sits on top of the encoder.}
    \label{fig:architecture}
    \vspace{-4mm}
\end{figure*}

\vspace{-2mm}
\section{Approach}
\vspace{-3mm}

The Pixel-based Encoder of Language, \model, consists of three major components: a text renderer, which draws text as an image; an encoder, which encodes the unmasked regions of the image; and a decoder, which reconstructs the masked regions at the pixel level. Figure~\ref{fig:architecture} provides an illustration.

\vspace{-3mm}
\subsection{Text Renderer}
\vspace{-2mm}

\label{sec:renderer}

\begin{figure*}[t]
    \includegraphics[width=\textwidth]{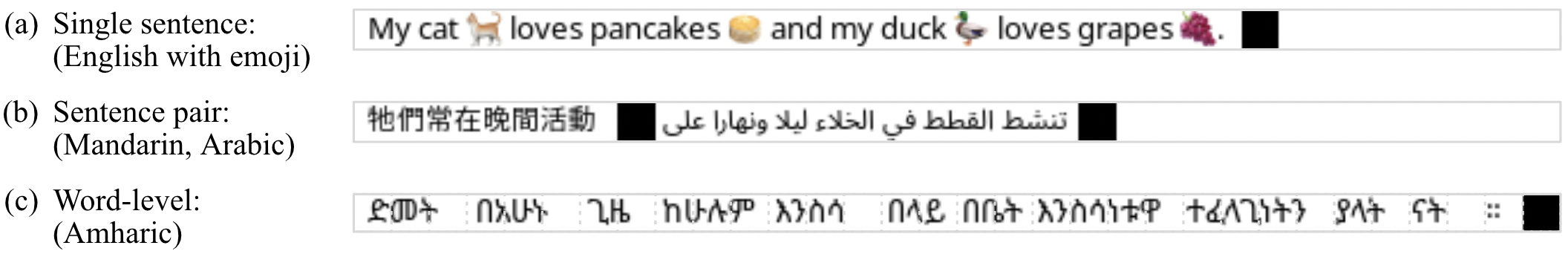}
    \caption{Illustrative examples of our rendered text. \model natively supports most writing systems, colour emoji (a), and complex text layouts such as right-to-left writing and ligatures (b). Black patches serve as separators and end-of-sequence markers. Blank patches to the right of the end-of-sequence marker are treated as sequence padding. For word-level tasks, horizontal spacing can be added between words (c) so that every patch can be assigned to exactly one word (dotted lines indicate patch boundaries for demonstration).} 
    \label{fig:rendered_examples}
    \vspace{-2mm}
\end{figure*}

The key component of \model is a text renderer that takes one or more pieces of text and renders them onto a blank RGB image $\bm{x} \in \mathbb{R}^{H \times W \times C}$. We set height $H=16$ and width $W=8464$ and choose $C=3$ RGB input channels, which is equivalent to a square colour image with a $368\times368$ resolution and corresponds to a sequence of $529$ image patches of size $16 \times 16$ pixels.\footnote{We chose a sequence length of 529 so that the memory requirements at maximum length are approx. equal to those of \textsc{bert}. Forward and backward passes of the transformer layers at equal length are also equally fast.}
Figure~\ref{fig:rendered_examples} shows examples of text inputs
rendered by the text renderer. 
The renderer supports (a) colour emoji and hieroglyphs scripts, (b) left-to-right and right-to-left writing systems, and (c) text that requires ligatures.  Analogous to \textsc{bert}, a sequence can either contain a single paragraph of text or a text pair; we use black $16 \times 16$ patches to serve as separators and end-of-sequence ({\footnotesize\texttt{EOS}}) markers. Blank (white) patches after the end-of-sequence marker are treated as padding by \model, where no attention scores or losses are computed. Sequences longer than the maximum length are either truncated or split into multiple sequences. Further technical details about the renderer are provided in Appendix~\ref{app:text_renderer}.

\vspace{-2mm}
\subsection{Architecture}
\vspace{-2mm}

\model-base is a 112M parameter ViT-MAE architecture \citep{he-etal-2022-mae} with a 12-layer ViT encoder \citep{dosovitskiy2021an} and an 8-layer Transformer decoder \citep{DBLP:conf/nips/VaswaniSPUJGKP17}. The encoder has $86$M parameters and the decoder has $26$M parameters, respectively. The 8-layer decoder is not used for downstream tasks. We give an overview of the architecture below, with more details in Appendix~\ref{app:architecture_details}. We did not train larger \model variants for lack of computational resources.

\vspace{-2mm}
\paragraph{Patch Embeddings} The images produced by the text renderer (\S\ref{sec:renderer}) are patch-wise linearly projected to obtain a sequence of patch embeddings with a 16 $\times$ 16 pixel resolution, to which fixed sinusoidal position embeddings are added.\footnote{This is a fast operation that does not require the large text embedding layer found in subword-based models, saving parameters which could in theory be re-allocated to the self-attention stack. We refer to \citet{xue-etal-2022-byt5} for a discussion regarding benefits and drawbacks of re-allocation of embedding layer weights.}

\begin{wrapfigure}{r}{0.35\textwidth}
    \vspace{-2.25em}
    \centering
    \begin{minipage}{\linewidth}
    \begin{algorithm}[H]
    \caption{\model Span Masking}\label{algo:span_masking}
    \begin{algorithmic}
        \scriptsize
        \Input{\#Image patches $N$, masking ratio $R$, maximum masked span length $S$, span length cumulative weights ${W=\{w_1,\ldots,w_S\}}$}
        \Output{Masked patches $\mathcal{M}$}
        \State $\mathcal{M} \leftarrow \emptyset $
        \Repeat{
            \State $s \leftarrow \text{randchoice}({\{1,\ldots, S\},W}$) 
            \State $l \leftarrow \text{randint}(0, \text{max}(0, N - s))$
            \State $r \leftarrow l + s$
            \If{$\mathcal{M} \cap \{l-s, \ldots, l-1\} = \emptyset $ \textbf{and} \\\hspace{8mm}$\mathcal{M} \cap \{r+1, \ldots, r+s\} = \emptyset $}
            \State $\mathcal{M} \leftarrow \mathcal{M} \cup \{l,\ldots,r\}$
            \EndIf
        }
        \Until{$\lvert \mathcal{M} \rvert > R \cdot N$} 
        \Return{$\mathcal{M}$}
    \end{algorithmic}
    \end{algorithm}
    \end{minipage}
    \vspace{-1em}
\end{wrapfigure}

\vspace{-2mm}
\paragraph{Patch Span Masking} Instead of the random masking procedure used in ViT-MAE or block-wise masking in BEiT \citep{bao2022beit}, \model uses span masking with a 25\% masking ratio as outlined in Algorithm~\ref{algo:span_masking}, which masks spans of up to $S=6$ consecutive image patches with a dynamic number of unmasked patches left between them. The idea behind the span masking approach, inspired by T5 \citep{raffel-etal-2020-t5} and SpanBERT \citep{joshi-etal-2020-spanbert}, is that it masks more meaningful units of text (full words or phrases) than random masking where the model more often has to fill in (parts of) individual characters, thereby encouraging \model to model a higher level of abstraction.
In practice, span masking was slightly more effective than random masking in early prototypes of \model. 
This effect may be less noticeable at higher masking ratios (such as the 75\% used in ViT-MAE), when random masking would more often masks consecutive patches.
We found 25\% masking ratio to work well for \model-base, which is in line with recent findings for \textsc{bert}-type models of similar size \citep{wettig-etal-2022-mask}. 
We mask spans of $s \in \{1,2,3,4\}$ patches in length, each with 20\% probability, and spans of $s \in \{5,6\}$ patches with 10\% probability each, so $\mathbb{E}(s)=3.1$.

\vspace{-2mm}
\paragraph{Encoder} Following ViT-MAE \citep{he-etal-2022-mae}, the \model encoder only processes unmasked patches (i.e., $\approx$ 396 ``visible'' patches at 25\% masking) rather than on a sequence including mask tokens, which not only reduces memory requirements and increases training speed, but also has the advantage of not creating a mismatch between pretraining and finetuning. This mismatch would occur when training the encoder with inserted mask tokens because they are not inserted during finetuning \citep{he-etal-2022-mae}. We also prepend the special \texttt{CLS} embedding to the unmasked patches.\footnote{In pretraining, no loss is computed for the \texttt{CLS} embedding but it can be used for finetuning.} The resulting \texttt{CLS} and unmasked patches are processed by a 12-layer Transformer encoder to produce a sequence of encoder output representations.

\vspace{-2mm}
\paragraph{Decoder} The \model decoder first projects the encoder outputs into the same space as the decoder model hidden size. It then inserts learnable mask embeddings at the masked positions; these are what \model tries to reconstruct at the pixel level. Fixed sinusoidal position embeddings \citep{DBLP:conf/nips/VaswaniSPUJGKP17} are added to inject order information. After processing this sequence via 8 Transformer layers, a linear projection yields patch logits. Note that the decoder does not have to compute an expensive softmax over a subword vocabulary and circumvents the question of whether to tie the subword embedding weights.
\model is trained with a normalised mean squared error (MSE) pixel reconstruction loss measuring the discrepancy between normalised target image patches and reconstructed patches. This loss is only computed for \emph{masked, non-blank (text)} patches. 

\vspace{-2mm}
\subsection{Pretraining}
\vspace{-2mm}

\model-base is pretrained on a rendered version of the English Wikipedia and the Bookcorpus \citep{Zhu_2015_ICCV}, which is roughly equivalent to the \textsc{bert} pretraining data.\footnote{We use a similar Wikipedia dump \citet{devlin-etal-2019-bert} used for \textsc{bert} (February 1, 2018) and a slightly newer version of the Bookcorpus available at \url{https://huggingface.co/datasets/bookcorpusopen}.} For better compute efficiency, we concatenate paragraphs until the maximum sequence length is reached, albeit not across document and book boundaries. Wikipedia has 2B words rendered into 11.4M examples and the Bookcorpus has 1.1B words rendered into 5.4M examples; in total \circa3.1B words (\textsc{bert} used 3.3B) rendered into 16.8M examples.\footnote{This rendering is quite compact; see Appendix~\ref{app:text_renderer}.} \model is pretrained for 1M steps with batch size 256 (i.e.\ \circa16 epochs) using the AdamW optimizer \citep{kingma-ba-2015-adam, loshchilov2018decoupled} with a linear warmup over the first 50k steps to a peak learning rate of \num{1.5e-4} and a cosine decay to a minimum learning rate of \num{1e-5}. Pretraining took 8 days on 8$\times$40GB Nvidia A100 GPUs. We show the loss curve
and additional pretraining details in Appendix~\ref{app:architecture_details}. We stored \model checkpoints every 10k steps and make them available alongside the fully trained model on the HuggingFace Hub \citep{wolf-etal-2020-transformers},
which we hope will be useful to analyze training dynamics of \model models \citep{sellam2022the}. Figure~\ref{fig:pretraining_dynamics} in Appendix~\ref{app:steps_reconstructions} shows, for three unseen examples, how \model learns to model language over the course of pretraining.

\vspace{-2mm}
\subsection{Finetuning}
\label{sec:finetuning}
\vspace{-2mm}

\model can be finetuned for downstream NLP tasks in a similar fashion to \textsc{bert}-like encoders by simply replacing the \model decoder with a suitable classification head. By truncating or interpolating the sinusoidal position embeddings, we can finetune with sequences shorter or longer than 529 patches, respectively. The latter, in particular, is common in computer vision applications to finetune on higher resolution images \citep{touvron-etal-2019, kolesnikov-etal-2020, dosovitskiy2021an, he-etal-2022-mae}. For most common NLP tasks, we can typically finetune with sequences shorter than 529 to accelerate training while retaining performance. To demonstrate that \model supports a variety of downstream tasks, we conduct finetuning experiments in four settings as follows:

\vspace{-2mm}
\paragraph{Word Classification} For word-level tasks like part-of-speech (POS) tagging and named entity recognition (NER), we render each word at the start of a new image patch so that we can create a bijective mapping between words and patches (see Figure~\ref{fig:rendered_examples} for an example).\footnote{This particular formulation assumes that word boundaries are available. We note that subword-based and character-based models also make this assumption. For further discussion on the implications, see Appendix~\ref{app:finetuning_details}.} To finetune \model on these images, we add a linear classifier with dropout. We assign the label of a word only to its first corresponding image patch and compute a cross-entropy loss with softmax.

\vspace{-2mm}
\paragraph{Dependency Parsing} For dependency parsing, we render text as above but obtain word-level representations by mean pooling over all corresponding image patches of a word and employ a biaffine parsing head \citep{DBLP:conf/iclr/DozatM17}, following the implementation from \cite{glavas-vulic-2021-supervised}.

\vspace{-2mm}
\paragraph{Sequence Classification} For sequence-level tasks, e.g.\ in GLUE \citep{wang-etal-2018-glue}, we render text as in pretraining. For sentence-pair tasks like natural language inference (NLI) we separate the sentences with a black patch. We finetune with different strategies, including training a classifier on top of
(1) the {\footnotesize\texttt{CLS}} embedding, (2) the mean-pooled or max-pooled representations of all patches, (3) a multi-head attention block. Although we did not notice significant performance differences between them in our experiments, we mainly used option (1), which is exactly the same as in \textsc{bert}, and (2), which has been shown to work well for image classification \citep{https://doi.org/10.48550/arxiv.2205.14540}.

\vspace{-2mm}
\paragraph{Extractive Question Answering (QA)} For extractive QA datasets like SQuAD \citep{rajpurkar-etal-2016-squad}, we render the question and context like in sequence-pair tasks above and, same as \cite{devlin-etal-2019-bert}, use a sliding window approach to extract answers for examples exceeding the maximum sequence length. We use a linear classifier to predict the start and end patches of the span containing the answer. Appendix~\ref{app:text_renderer} explains how we obtain the mapping between characters and rendered text.

\vspace{-2mm}
\section{Experiments}
\vspace{-2mm}

We finetune \model on common NLP tasks and evaluate its syntactic and semantic processing capabilities in English, as well as its adaptability to unseen languages. Table~\ref{tab:language_overview} (Appendix \ref{app:finetuning_details}) describes the languages used in these experiments, and our language and data selection is also motivated below.

\vspace{-2mm}
\subsection{Tasks and Languages}
\label{sec:tasks_languages}
\vspace{-2mm}

\paragraph{Syntactic Tasks} We evaluate \model on part-of-speech (POS) tagging and dependency parsing using data from Universal Dependencies v2.10 treebanks \citep{nivre-etal-2020-universal, zeman-etal-2022-ud} for a set of typologically diverse languages that captures a large variety of unseen scripts\footnote{By unseen, we mean not present in the pretraining data.}: Arabic (\arabi), Coptic (\coptic), English (\english), Hindi (\hindi), Japanese (\japanese), Korean (\korean), Tamil (\tamil), Vietnamese (\vietnamese), Chinese (\chinese).\footnote{Table~\ref{tab:treebank_overview} in Appendix~\ref{app:finetuning_details} gives an overview of the treebanks we use.}
We compare how well \model{} transfers to these languages compared to \textsc{bert}. Note that \textsc{bert} does not support all of these writing systems. However, both models have been trained on the same data. This comparison allows us to gauge the extent to which \model can overcome the script barrier and vocabulary bottleneck of subword-based models.

\vspace{-2mm}
\paragraph{Semantic Tasks} We evaluate both monolingual (\english) and cross-lingual \emph{word-level} understanding on MasakhaNER \citep{adelani-etal-2021-masakhaner}, a named entity recognition (NER) benchmark for 10 African languages (\amharic, \hausa, \igbo, \kinyarwanda, \luganda, \luo, \naija, \swahili, \wolof, \yoruba), which also includes a copy of the ConLL-2003 dataset \citep[\english; ][]{tjong-kim-sang-de-meulder-2003-introduction}. For monolingual \english \emph{sentence-level} understanding we rely on GLUE \citep{wang-etal-2018-glue} and SQuAD \citep{rajpurkar-etal-2016-squad}. 
Finally, we evaluate cross-lingual sentence-level understanding on TyDiQA-GoldP \citep{clark-etal-2020-tydi} in the \emph{in-language multitask} setting where we train on the combined gold data in all 9 target languages (\arabi, \bengali, \english, \finnish, \indonesian, \korean, \russian, \swahili, \telugu) at once, and on two additional larger monolingual extractive question answering (QA) corpora: KorQuAD 1.0 \citep[\korean; ][]{lim-etal-2019-korquad} and JaQuAD \citep[\japanese; ][]{so2022jaquad}.

\vspace{-2mm}
\subsection{Baselines and Finetuning protocols}
\vspace{-2mm}

We compare results to \textsc{bert}-base which is trained on the same data.\footnote{We use \textsc{bert} weights from \url{https://huggingface.co/bert-base-cased}.} We do not compare to newer monolingual English models like \textsc{roberta} \citep{DBLP:journals/corr/abs-1907-11692}, \textsc{T5} \citep{raffel-etal-2020-t5} or \textsc{deberta} \citep{he-etal-2020-deberta, he-etal-2021-debertav3} because these models have been pretrained longer on much larger corpora.\footnote{We do not intend to claim state-of-the-art performance, but to demonstrate that \model can overcome the vocabulary bottleneck and to provide a starting point for further research on pixel-based encoding of language.} 
Likewise, we do not compare against models trained on massively multilingual corpora.
However, to contextualise the performance of \model in cross-lingual settings, we report results for \textsc{mbert} and, if results are available, for \textsc{canine} \citep{DBLP:journals/tacl/ClarkGTW22}. For \textsc{bert}, we use the standard finetuning protocols used by \cite{devlin-etal-2019-bert} and the same biaffine classifier for parsing as for \model. We list finetuning details for all tasks in Appendix~\ref{app:finetuning_details}.

\vspace{-2mm}
\subsection{Results}
\label{sec:results}
\vspace{-2mm}

\paragraph{Syntactic Tasks} We present results for POS tagging and dependency parsing in Table~\ref{res:syntactic_task_results}. While \textsc{bert} is slightly better than \model in the monolingual setting (\english), \model clearly outperforms \textsc{bert} in the remaining languages. On the lower end, the accuracy gap in favor of \model in \arabi and \vietnamese, both languages covered by \textsc{bert}'s vocabulary, is relatively small (\circa1\%). On the higher end, in \coptic, where \textsc{bert} has an out-of-vocabulary ({\footnotesize{[\texttt{UNK}]}}) token ratio of 93\%, the gap is \circa70\% for both tasks. There is a strong correlation\footnote{Pearson correlation $r=0.9$, $p<0.001$ for POS tagging, $r=0.95$, $p<0.0001$ for dependency parsing.} between the proportion of {\footnotesize{[\texttt{UNK}]}}s (shown in Table~\ref{res:syntactic_task_results} on the right) and the performance gap, which shows that \model overcomes \textsc{bert}'s vocabulary bottleneck. These results are further analysed in Appendix~\ref{app:analysis}.

\begin{table*}[t]
\centering
\resizebox{0.765\textwidth}{!}{%
\begin{tabular}{@{}lcccccccccc@{}}
\toprule
 &
$\lvert\theta\rvert$ &
\english &
\arabi &
\coptic &
\hindi &
\japanese &
\korean &
\tamil &
\vietnamese &
\textsc{zho} \\ \midrule \addlinespace[0.3em] \rowcol\multicolumn{11}{c}{\textit{POS Tagging (Accuracy)}}                                                                           \\\addlinespace[0.2em]
\textsc{bert}  & 110M & \textbf{97.2}        & 95.4         & 26.5         &   86.4 & 87.9        & 60.0         & 45.4         & 84.5         & 58.6         \\
\model         & ~~86M & 96.7        & \textbf{95.7}         & \textbf{96.0}         &  \textbf{96.3} & \textbf{97.2}       & \textbf{94.2}         & \textbf{81.0}        & \textbf{85.7}         & \textbf{92.8}         \\ \midrule \addlinespace[0.3em] \rowcol\multicolumn{11}{c}{\textit{Dependency Parsing (LAS)}}                                                                    \\\addlinespace[0.2em]
\textsc{bert}  & 110M & \textbf{90.6} & \textbf{77.7} & 13.0 &  75.9 & 73.8 & 30.2 & 15.2 & 49.4 & 28.8 \\
\textsc{\model} & ~~86M  & 88.7 & 77.3 & \textbf{83.5} & \textbf{89.2} & \textbf{90.7} & \textbf{78.5} & \textbf{52.6} &\textbf{50.5} & \textbf{73.7} \\ \bottomrule
\end{tabular}%
}
\quad
\resizebox{0.195\textwidth}{!}{%
\begin{tabular}{@{}lrc@{}}
\toprule
         & {\footnotesize{[\texttt{UNK}]}}\% & Fertility \\ \midrule
\english    & 0 \qquad                                  & 1.2       \\
\arabi      & 1.8 \qquad                                & 3.7       \\
\coptic     & 93.6 \qquad                               & 1.0       \\
\hindi      & 32.6 \qquad                               & 2.7       \\
\japanese   & 45.5 \qquad                               & 1.5       \\
\korean     & 84.7 \qquad                               & 1.0       \\
\tamil      & 82.3 \qquad                               & 1.3       \\
\vietnamese & 4.5 \qquad                               & 2.5       \\
\chinese    & 73.2 \qquad                               & 1.5       \\ \bottomrule
\end{tabular}%
}
\caption{Results for \model and \textsc{bert} finetuned for POS tagging and dependency parsing on various Universal Dependencies treebanks. We report test set results averaged over 5 runs each. $\lvert\theta\rvert$ denotes the number of model parameters. The table on the right shows \textsc{bert}'s proportion of {\footnotesize{[\texttt{UNK}]}}s as a measure of (inverse) vocabulary coverage and fertility \citep[i.e., number of subwords per tokenized word; ][]{acs:2019, rust-etal-2021-good} as a measure of over-segmentation in respective UD treebanks. }
\label{res:syntactic_task_results}
\vspace{-2mm}
\end{table*}

\begin{table*}[t!]
\centering
\resizebox{\textwidth}{!}{%
\begin{tabular}{lccccccccccccc}
\toprule
&
\textsc{\#l} &
$\lvert\theta\rvert$ &
\english &
\amharic &
\hausa &
\igbo &
\kinyarwanda &
\luganda &
\luo &
\naija &
\swahili &
\wolof &
\yoruba \\ \midrule
\tc{\textsc{mbert*}} &
  \tc{104} &
  \tc{179M} &
  \tc{92.2} &
 \tc{0} &
  \tc{87.3} &
  \tc{85.3} &
  \tc{72.6} &
  \tc{79.3} &
  \tc{73.5} &
  \tc{86.4} &
  \tc{87.5} &
  \tc{62.2} &
  \tc{80.0} \\
\tc{\textsc{canine-c} + n-gram*} &
\tc{104} &
\tc{167M} &
\tc{89.8} &
\tc{50.0} &
\tc{88.0} &
\tc{85.0} &
\tc{72.8} &
\tc{79.6} &
\tc{74.2} &
\tc{88.7} &
\tc{83.7} &
\tc{66.5} &
\tc{79.1} \\
\tc{\textsc{canine-c*}} &
  \tc{104} &
  \tc{127M} &
  \tc{79.8} &
  \tc{44.6} &
  \tc{76.1} &
  \tc{75.6} &
  \tc{58.3} &
  \tc{69.4} &
  \tc{63.4} &
  \tc{66.6} &
  \tc{72.7} &
  \tc{60.7}&
  \tc{67.9} \\
 \midrule
\textsc{bert} &
  \quad 1 &
  110M &
  \textbf{92.9} &
  0             &
  \textbf{86.6} &
  \textbf{83.5} &
  \textbf{72.0} &
  \textbf{78.4} &
  \textbf{73.2} &
  \textbf{87.0} &
  \textbf{83.3} &
  \textbf{62.2} &
  \textbf{73.8} \\
\model &
  \quad 1 & 
  ~~86M &
  89.5 &
  \textbf{47.7} &
  82.4 &
  79.9 &
  64.2 &
  76.5 &
  66.6 &
  78.7 &
  79.8 &
  59.7 &
  70.7 \\ \bottomrule
\end{tabular}%
}
\caption{Results for \model and \textsc{bert} finetuned for NER on MasakhaNER. We report test set $F_1$ scores averaged over 5 runs each. \textsc{bert} outperforms \model in all of the languages that use Latin script, whereas \model does better on \amharic, whose script is not covered by \textsc{bert}'s vocabulary. The performance gap is smaller for languages heavier in diacritics, e.g.\ \yoruba. It is larger for languages closer to English such as Naija Pidgin (\naija), an English-based creole. \textsc{\#l} denotes the number of pretraining languages and \textbf{*} indicates results taken from \cite{DBLP:journals/tacl/ClarkGTW22} for additional context.}
\label{res:ner_results}
\vspace{-2mm}
\end{table*}

\vspace{-2mm}
\paragraph{Semantic Tasks} We present results for NER in Table~\ref{res:ner_results}, for GLUE in Table~\ref{res:glue_results}, for QA in Table~\ref{res:qa_results}.
We also conduct experiments on XNLI in the \emph{translate-train-all} setting which we present in Table~\ref{res:xnli} in Appendix~\ref{app:analysis}, for brevity.
We find that \textsc{bert} consistently achieves higher performance than \model in its pretraining language \english. Likewise, it often outperforms on languages using the Latin writing system; for instance in NER where all languages besides \amharic use Latin script, in QA for \finnish, \indonesian and \swahili.
Although \textsc{bert} has more trainable parameters, this finding indicates that a \model model pretrained for the same number of steps as \textsc{bert} is slightly worse at semantic tasks, and it may require longer pretraining or an additional inductive bias to close the performance gap. 
Similarly, character-based models also tend to underperform subword-based models on NER \citep{keren-etal-2022-breaking}, here seen by the \textsc{canine-c} results.
Since the addition of n-gram embeddings improves the performance of \textsc{canine-c}, likely due to boosting entity memorisation capabilities \citep{DBLP:journals/tacl/ClarkGTW22}, we hypothesize that \model may benefit from equivalent enhancements.

For languages where \textsc{bert} only partially covers the script, such as \korean, \japanese and \telugu in QA,
\model consistently outperforms \textsc{bert}, sometimes by large amounts (e.g.\ , +63 $F_1$ points better on KorQuAD). In the extreme case where \textsc{bert} has no coverage of the script whatsoever, seen in NER for \amharic, \textsc{bert} fails completely (0~$F_1$) while \model outperforms the larger, multilingually trained \textsc{canine} and performs competitively with its n-gram variant. In other words, \model also overcomes the vocabulary bottleneck of subword-based PLMs in semantics-driven tasks.
Note that although \textsc{bert} was trained on English, its vocabulary has a high coverage of the Arabic script, explaining its good performance in \arabi and \urdu.\footnotemark

While the same may apply to languages like \bengali and \russian in QA, where one may otherwise expect \model to outperform \textsc{bert}, there is an external factor at play; in the standard QA task formulation used by \textsc{bert}, answer spans are extracted by predicting start and end tokens. We adopt this procedure in \model for simplicity. However, an image patch will often overlap two words at variable positions, so the answer may actually start or end mid-patch. By only predicting on a full-patch level, and extracting the entire content of the patch, \model will sometimes extract leading and trailing characters that should not be part of the answer, which degrades the $F_1$ score---even though the model may have correctly identified the span. Languages not using whitespace to delimit words are particularly affected, which also explains why \model is only slightly better than \textsc{bert} in \japanese.

\begin{table*}[t]
\centering
\resizebox{\textwidth}{!}{%
\begin{tabular}{@{}lccccccccccc@{}}
\toprule
&
$\lvert\theta\rvert$ &
  \begin{tabular}[c]{@{}c@{}}\textsc{MNLI-M/MM}\\ 393k\end{tabular} &
  \begin{tabular}[c]{@{}c@{}}\textsc{QQP}\\ 364k\end{tabular} &
  \begin{tabular}[c]{@{}c@{}}\textsc{QNLI}\\ 105k\end{tabular} &
  \begin{tabular}[c]{@{}c@{}}\textsc{SST-2}\\ 67k\end{tabular} &
  \begin{tabular}[c]{@{}c@{}}\textsc{COLA}\\ 8.6k\end{tabular} &
  \begin{tabular}[c]{@{}c@{}}\textsc{STS-B}\\ 5.8k\end{tabular} &
  \begin{tabular}[c]{@{}c@{}}\textsc{MRPC}\\ 3.7k\end{tabular} &
  \begin{tabular}[c]{@{}c@{}}\textsc{RTE}\\ 2.5k\end{tabular} &
  \begin{tabular}[c]{@{}c@{}}\textsc{WNLI}\\ 635\end{tabular} &
  \textsc{AVG} \\ \midrule
\textsc{bert} &
  110M &
  \textbf{84.0} / \textbf{84.2} &
   \textbf{87.6} &
   \textbf{91.0} &
   \textbf{92.6} &
   \textbf{60.3} &
   \textbf{88.8} &
   \textbf{90.2} &
  \textbf{69.5} &
  51.8 &
  \textbf{80.0} \\
\model &
  ~~86M &
  78.1 / 78.9 &
  84.5 &
  87.8 &
  89.6 &
  38.4 &
  81.1 &
  88.2 &
  60.5 &
  \textbf{53.8} &
  74.1 \\ \bottomrule
\end{tabular}%
}
\caption{Results for \model and \textsc{bert} finetuned on \textsc{glue}. We report \emph{validation} set performance averaged over 5 runs. The metrics are $F_1$ score for \textsc{qqp} and \textsc{mrpc}, Matthew's correlation for \textsc{cola}, Spearman's $\rho$ for \textsc{sts-b}, and accuracy for the remaining datasets. \model achieves non-trivial performance scores on \textsc{glue}, indicating \emph{pixel-based encoders can learn higher-level semantic tasks}, but performs worse overall than \textsc{bert}, so it may require (a) more pretraining steps than subword-tokenized PLMs or (b) additional inductive bias to acquire the same level of monolingual abstraction.}
\label{res:glue_results}
\vspace{-2mm}
\end{table*}

\begin{table*}[bt!]
\centering
\resizebox{\textwidth}{!}{%
\begin{tabular}{lccccccccccccccc}
\toprule
\multirow{2}{*}{} &
  \multirow{2}{*}{\textsc{\#l}} &
  \multicolumn{1}{c}{\multirow{2}{*}{$\lvert\theta\rvert$}} &
  \multicolumn{10}{c}{TyDiQA-GoldP} &
  \multicolumn{1}{c}{SQuAD} &
  KorQuAD &
  JaQuAD \\
 &
   &
  \multicolumn{1}{l}{} &
  \english &
  \arabi &
  \bengali &
  \finnish &
  \indonesian &
  \korean &
  \russian &
  \swahili &
  \telugu &
  \textbf{\textsc{AVG}} &
  \english &
  \korean &
  \japanese \\ \midrule
\tc{\textsc{mbert}} &
  \tc{104} &
  \tc{179M} &
  \tc{75.6} &
  \tc{78.1} &
  \tc{74.7} &
  \tc{75.5} &
  \tc{84.3} &
  \tc{64.8} &
  \tc{74.9} &
  \tc{83.1} &
  \tc{81.6} &
  \tc{77.1} &
  \tc{88.6} &
  \tc{90.0} &
  \tc{76.4} \\ \midrule
\textsc{bert} &
  \quad 1 &
  110M &
  \textbf{68.5} &
  \textbf{58.0} &
  \textbf{43.2} &
  \textbf{58.3} &
  \textbf{67.1} &
  12.4 &
  \textbf{53.2} &
  \textbf{71.3} &
  48.2 &
  51.5 &
  \textbf{88.2} &
  14.9 &
  28.8 \\
\model &
  \quad 1 &
  ~~86M &
  59.6 &
  57.3 &
  36.3 &
  57.1 &
  63.6 &
  \textbf{26.1} &
  50.5 &
  65.9 &
  \textbf{61.7} &
  \textbf{52.3} &
  81.4 &
  \textbf{78.0} &
  \textbf{34.1} \\ \bottomrule
\end{tabular}%
}
\caption{Results for \model and \textsc{bert} finetuned on extractive QA datasets. We report validation set $F_1$ scores averaged over 5 runs each. Average (\textsc{avg}) scores for TyDiQA-GoldP exclude \english as customary \citep{clark-etal-2020-tydi}. While \textsc{bert} clearly outperforms \model in \english, \model is much better in \korean, \telugu, and \japanese---a consequence of the vocabulary bottleneck in \textsc{bert}---thereby gaining an edge on average. In some languages, answer span extraction adversely affects results (see \S\ref{sec:results}).}
\label{res:qa_results}
\vspace{-4mm} 
\end{table*}

Generally, and in particular when transferring to unseen scripts, we find that \model performs best when finetuning on larger corpora. An example of this behaviour can be seen in QA, where \model performs significantly better on KorQuAD (60k examples) than the \korean subset of TyDi (1.6k examples). While large corpora may often not be available when dealing with unseen scripts, we hypothesize that multilingual pretraining will alleviate the need for long finetuning, while potentially being even more conducive to \emph{positive transfer} \citep{conneau-etal-2020-unsupervised, chau-etal-2020-parsing, pfeiffer-etal-2021-unks} by not being vocabulary-bottlenecked.

\footnotetext{Arabic is lexically sparse \citep{antoun-etal-2020-arabert, 10.1145/3086575}, so the characters can be covered in the vocabulary. However, it is morphologically complex, which leads to over-segmentation, as the fertility of 3.7 in Table~\ref{res:syntactic_task_results} shows. This over-segmentation is not necessarily problematic in our selection of tasks \citep{keren-etal-2022-breaking}, e.g.\ due to the sliding window in QA, but can be a disadvantage in others \citep{rust-etal-2021-good}.}

\vspace{-2mm}
\section{Robustness to Orthographic Attacks and Code-Switching}
\vspace{-2mm}

Informal text, commonly found on social media, often contains orthographic noise such as typos and other variations \citep{baldwin-etal-2015-shared, EschWritingAcrossWorld, caswell2020language}.
Previous work has demonstrated the vulnerability of pretrained language models to 
character-level adversarial attacks and noise  \citep{SunADV-BERT, eger-benz-2020-hero}, with text normalization typically required to maintain performance \citep{pruthi-etal-2019-combating, keller-etal-2021-bert}.
To evaluate \model{}'s robustness to textual noise and variation and inspired by the robustness tests of \citet{salesky-etal-2021-robust}, we experiment with the \textit{Zeroé} benchmark \citep{eger-benz-2020-hero,keller-etal-2021-bert} which covers a variety of low-level orthographic attacks as illustrated in Table \ref{tab:zeroe-examples}.
We replace their version of visual attacks with the Unicode Technical Standard \#39 set of visually-confusable characters.\footnote{\url{https://util.unicode.org/UnicodeJsps/confusables.jsp}} 
We apply \textit{Zeroé} attacks during finetuning and evaluation of two English downstream tasks, POS tagging and NLI \citep{bowman-etal-2015-large}, where we expect models to rely on different levels of abstraction.

\begin{figure*}[!t]
    \centering
    \begin{subfigure}[]{0.31\textwidth}
        {%
        \setlength{\fboxsep}{0pt}%
        \setlength{\fboxrule}{1pt}%
        \fbox{\includegraphics[width=\textwidth]{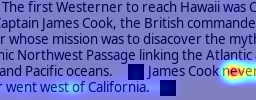}}%
        }%
        \caption{0\%, \emph{contradiction}}
    \end{subfigure}
    \quad
    \begin{subfigure}[]{0.31\textwidth}
        {%
        \setlength{\fboxsep}{0pt}%
        \setlength{\fboxrule}{1pt}%
        \fbox{\includegraphics[width=\textwidth]{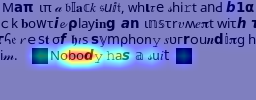}}%
        }%
        \caption{80\%, \emph{contradiction}}
    \end{subfigure}
    \quad
    \begin{subfigure}[]{0.31\textwidth}
        {%
        \setlength{\fboxsep}{0pt}%
        \setlength{\fboxrule}{1pt}%
        \fbox{\includegraphics[width=\textwidth]{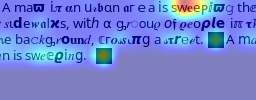}}%
        }%
        \caption{80\%, \emph{entailment}}
    \end{subfigure}
    \caption{Visual explanations of correct \model predictions (for classes \emph{contradiction} and \emph{entailment}) for NLI examples with 0\% and 80\% \textsc{confusable} substitutions using method by \cite{Chefer_2021_ICCV}, providing qualitative evidence for \model's robustness to character-level noise and the interpretability of its predictions.
    Red heatmap regions represent high relevancy.}
    \label{fig:pixel_interpretability}
    \vspace{-2mm}
\end{figure*}

Figures~\ref{fig:robustness_snli} and \ref{fig:robustness_pos} in Appendix~\ref{app:zeroe} compare \model and \textsc{bert} across three levels of token-level noise for POS tagging and NLI. 
There is little impact on POS tagging performance with either model from most low-level attacks, with the exception of visually-confusable character substitutions (\textsc{confusable}); here \model expectedly maintains performance above 92\% as it generalizes across orthographic similarities but \textsc{bert} drops to 38\%. 
For NLI, both models are negatively affected, but \model exhibits less degradation than \textsc{bert} with higher proportions of noise, with the impact varying across the types of attacks which each affect subword tokenization differently. Figure~\ref{fig:pixel_interpretability} shows relevancy heatmaps \citep{Chefer_2021_ICCV} for SNLI predictions made with and without \textsc{confusable} substitutions. The heatmaps are similarly clear with and without noise, providing qualitative evidence that \model is indeed robust to the noise. 
The illustrated robustness may be dependent upon finetuning, however; we find that \model can struggle in zero-shot applications when text is rendered differently from observed during pretraining (see Appendix~\ref{app:text_renderer} on using different fonts). 
Future work could explore the impact of data augmentation during pretraining on \model's robustness and ability to transfer across scripts. Furthermore, it would be interesting to investigate how the choice of font influences the search space during reconstruction of masked patches \citep{Bland2022StoryBT}.

\begin{wraptable}{r}{0.45\textwidth}
\vspace{-1em}
\setlength\tabcolsep{2pt}
\centering
\resizebox{0.43\textwidth}{!}{%
\begin{tabular}{lccccc}
\toprule
\multirow{2}{*}{} &
  \multicolumn{2}{c}{POS Tagging} &
  \multicolumn{3}{c}{Named Entity Recognition} \\
 &
  \spanish-\english &
  \hindi-\english &
  \spanish-\english &
  \hindi-\english &
  \modernarabic-\egyptarabic \\ 
\midrule
\tc{\textsc{mbert}} &
  \tc{97.1} &
  \tc{86.3} &
  \tc{64.0} &
  \tc{72.6} &
  \tc{65.4} \\
\midrule
\textsc{bert} &
  \textbf{96.9} &
  {87.0} &
  \textbf{61.1} &
  \textbf{74.5} &
  {59.4} \\
\model &
  {96.8} &
  \textbf{88.2} &
  {61.0} &
  {73.0} &
  \textbf{63.7} \\ 
\bottomrule
\end{tabular}%
}
\vspace{-0.15cm}
\caption{Code-switching results on \textsc{LinCE}.}
\label{res:cs_results}
\vspace{-0.3cm}
\end{wraptable}

In addition to robustness to orthographic noise, dealing with character-level substitutions is important for effectively modelling different morphological forms.
There are also many types of higher-level token, phrase or sequence-level variations such as code-switching---when a speaker alternates between two or more languages in the same utterance, while being grammatically consistent in each language~\citep{joshi-1982-processing}---or the lexical substitutions in social media text.
We evaluate \model{} on the LinCE benchmark~\citep{aguilar-etal-2020-lince}, which includes core tasks and downstream applications for linguistic code-switching.
\model{} is fine-tuned on POS Tagging and NER in Spanish-English, Hindi-English and Modern Standard Arabic-Egyptian Arabic.
Table~\ref{res:cs_results} shows that \model{} and \textsc{bert} perform similarly on \spanish-\english tasks, with \textsc{bert} outperforming \model{} on NER for (romanised) \hindi-\english.
On the other tasks, \model{} performs better than \textsc{bert} and even outperforms \textsc{mbert} on \hindi-\english POS tagging. The gap between \textsc{mbert} and \model{} is larger on Arabic scripts, which were extensively seen by \textsc{mbert} during pretraining.

\vspace{-2mm}
\section{Related Work}
\label{sec:related_work}
\vspace{-2mm}

The question of vocabulary construction is an open problem in NLP, especially in a multilingual context.\footnote{See \citet{DBLP:journals/corr/abs-2112-10508} for a recent, comprehensive survey on open-vocabulary modeling and tokenization.}
The most widely used language models, e.g.\ BERT, RoBERTa, T5, GPT-2 \textit{inter alia}, rely on different tokenizers, such as WordPiece \citep{devlin-etal-2019-bert}, Byte-Pair Encoding \citep[BPE; ][]{sennrich-etal-2016-neural} and Unigram LM \citep{kudo-2018-subword}. There is an established ecosystem around subword tokenizers, such as the SentencePiece \citep{kudo-richardson-2018-sentencepiece} and HuggingFace Tokenizers.

In a monolingual context and for some languages like English, vocabularies of subwords are a good tradeoff between vocabularies of characters and vocabularies of words.
When representing a large number of languages in multilingual PLMs like mBERT and XLM-R, adequately representing the vocabulary of each individual language would be computationally prohibitive. 
The tokenization then becomes a bottleneck when trying to scale up to a large number of languages \citep{conneau-etal-2020-unsupervised, rust-etal-2021-good}, which manifests itself in degraded cross-lingual performance to languages and language families that are underrepresented in the data used for training multilingual PLMs. There are large inequalities in the performance of these models across typologically diverse languages \citep{wu-dredze-2020-languages, lauscher-etal-2020-zero}. 
This issue is further exacerbated by tokenizations out-of-the-box not being compatible across languages \citep{maronikolakis-etal-2021-wine-v}. Language imbalance and poor character coverage in the vocabulary can also decrease downstream performance \citep{zhang2022robust}.
To some extent, these problems can be attenuated through techniques such as subword mapping \citep{vernikos-popescu-belis-2021-subword-mapping}, transliteration \citep{DBLP:journals/corr/abs-2201-12501}, leveraging lexical overlap \citep{patil-etal-2022-overlap}, vocabulary clustering and reallocation \citep{chung-etal-2020-improving}, continued or language-adaptive pretraining \citep{ebrahimi-kann-2021-adapt}, adaptation via bilingual lexica \citep{wang-etal-2022-expanding}, and embedding matrix adaptation \citep{artetxe-etal-2020-cross}. However, these are post-hoc workarounds to expand model vocabularies after training. They do not provide a direct solution to the vocabulary bottleneck problem.

Some subword-based algorithms can also produce undesirable segmentations for morphologically rich languages \citep{klein-tsarfaty-2020-getting, amrhein-sennrich-2021-suitable-subword}, so dedicated morphologically-aware tokenizers have been developed (e.g.\ \citet{smit-etal-2014-morfessor}), but this process often requires expert-level knowledge and may only work for individual languages.

Due to the limitations of subword vocabularies in multilingual language modelling, some works have used vocabularies over characters \citep[][\emph{inter alia}]{lee-etal-2017-fully, ma-etal-2020-charbert} or bytes \citep{DBLP:conf/aaai/WangCG20, DBLP:journals/corr/abs-2101-09469}. These provide benefits over purely subword-based models in terms of robustness and most of them are readily applicable in a multilingual context,\footnote{Character-aware models are not directly applicable to languages that do not use whitespace to delimit sentences \citep{tay2021charformer}, for example.} but they typically come at the cost of increased sequence lengths or latency. Also, such models cannot exploit orthographic similarities between characters across and within scripts and do not account for the fact that meaning of language may be carried visually such as in writing systems that are (partially) logographic like Chinese, in ancient hieroglyphs, or when using emoji.

Finally, some works have developed pixel-based approaches. \cite{broscheit-2018-learning} embedded images of Chinese glyphs but still relied on a fixed vocabulary. 
\cite{Wu2019GlyceGF} combined character-level images and embeddings for a variety of Chinese tasks. 
\cite{DBLP:conf/icml/RadfordKHRGASAM21} trained a linear probe for CLIP, which also incorporates a tokenizer,
on a rendered version of SST-2 \citep{socher-etal-2013-recursive}. Other works have trained pixel-based models that removed the need for a fixed vocabulary: 
\cite{Sun2019SquaredEW} trained a convolutional sentiment classifier on pixels. 
\cite{mansimov-etal-2020-towards} used images of text for in-image MT. 
\cite{salesky-etal-2021-robust} employed a convolutional embedder for a Transformer-based MT system with a subword-based decoder. Our method differs from these in that it provides a general-purpose language encoder that completely removes the need for a vocabulary.

\vspace{-2mm}
\section{Conclusion}
\vspace{-2mm}

This paper introduced \model, a pretrained language model that renders text as images, which allows it to represent any written language that can be typeset using its text renderer. \model was pretrained on the predominantly English Wikipedia and Bookcorpus datasets, and evaluated on part-of-speech tagging, dependency parsing, question answering, and language understanding tasks. The results demonstrate that \model readily transfers to unseen scripts, as shown by its performance on 14 scripts across 32 languages. \model currently lags behind \textsc{bert} when processing languages with a Latin script, including English; however, \model is more robust than \textsc{bert} against low-level orthographic attacks and performs competitively to \textsc{bert} and \textsc{mBert} on linguistic code-switching tasks. Overall, these results show that pixel-based representations are a strong backbone for cross-lingual and cross-script transfer learning. The limitations of this work are discussed in Appendix~\ref{app:limitations}.

In future work, we will investigate inductive biases and additional objectives that can better capture long-range dependencies in \model models. We hope that this will help overcome the limits of \model in semantic processing.
We also plan to pretrain \model on multilingual text with a view to further improving its cross-script and cross-lingual abilities. This will also allow us to more fairly compare pixel-based models against larger subword-based and tokenization-free \textit{multilingual} models. Finally, we will also develop new rendering and finetuning formulations that are better tailored to pixel-based models, e.g. for improving downstream question answering.

\section*{Acknowledgments}
We thank Ákos Kádár, Barbara Plank, and Kris Cao for their comments on an earlier draft. We also thank Davide Rigoni, Rita Ramos, Stella Frank, and members of the CoAStaL and LAMP groups for discussions. Miryam de Lhoneux is funded by the Swedish Research Council (grant 2020-00437). Phillip Rust is funded by the Novo Nordisk Foundation (grant NNF 20SA0066568). Jonas F. Lotz is funded by the ROCKWOOL Foundation (grant 1242). {\scriptsize\euflag} Emanuele Bugliarello is supported by funding from the European Union's Horizon 2020 research and innovation programme under the Marie Sk\l{}odowska-Curie grant agreement No 801199. Elizabeth Salesky is supported by the Apple Scholars in AI/ML fellowship. Desmond Elliott is partially supported by the Innovation Foundation (grant 0176-00013B) and the Novo Nordisk Foundation (grant NNF 20SA0066568). This work was supported by a research grant (VIL53122) from VILLUM FONDEN. The computing power was generously supported by EuroHPC grants 2010PA5869, 2021D02-068, and 2021D05-141, and with Cloud TPUs from Google's TPU Research Cloud (TRC).

\bibliography{bibliography}

\begin{thebibliography}{107}
\providecommand{\natexlab}[1]{#1}
\providecommand{\url}[1]{\texttt{#1}}
\expandafter\ifx\csname urlstyle\endcsname\relax
  \providecommand{\doi}[1]{doi: #1}\else
  \providecommand{\doi}{doi: \begingroup \urlstyle{rm}\Url}\fi

\bibitem[\'Acs(2019)]{acs:2019}
Judit \'Acs.
\newblock Exploring {BERT}'s {V}ocabulary.
\newblock \emph{Blog Post}, 2019.
\newblock URL
  \url{http://juditacs.github.io/2019/02/19/bert-tokenization-stats.html}.

\bibitem[Adelani et~al.(2021)Adelani, Abbott, Neubig, D’souza, Kreutzer,
  Lignos, Palen-Michel, Buzaaba, Rijhwani, Ruder, Mayhew, Azime, Muhammad,
  Emezue, Nakatumba-Nabende, Ogayo, Anuoluwapo, Gitau, Mbaye, Alabi, Yimam,
  Gwadabe, Ezeani, Niyongabo, Mukiibi, Otiende, Orife, David, Ngom, Adewumi,
  Rayson, Adeyemi, Muriuki, Anebi, Chukwuneke, Odu, Wairagala, Oyerinde, Siro,
  Bateesa, Oloyede, Wambui, Akinode, Nabagereka, Katusiime, Awokoya, MBOUP,
  Gebreyohannes, Tilaye, Nwaike, Wolde, Faye, Sibanda, Ahia, Dossou, Ogueji,
  DIOP, Diallo, Akinfaderin, Marengereke, and
  Osei]{adelani-etal-2021-masakhaner}
David~Ifeoluwa Adelani, Jade Abbott, Graham Neubig, Daniel D’souza, Julia
  Kreutzer, Constantine Lignos, Chester Palen-Michel, Happy Buzaaba, Shruti
  Rijhwani, Sebastian Ruder, Stephen Mayhew, Israel~Abebe Azime, Shamsuddeen~H.
  Muhammad, Chris~Chinenye Emezue, Joyce Nakatumba-Nabende, Perez Ogayo, Aremu
  Anuoluwapo, Catherine Gitau, Derguene Mbaye, Jesujoba Alabi, Seid~Muhie
  Yimam, Tajuddeen~Rabiu Gwadabe, Ignatius Ezeani, Rubungo~Andre Niyongabo,
  Jonathan Mukiibi, Verrah Otiende, Iroro Orife, Davis David, Samba Ngom, Tosin
  Adewumi, Paul Rayson, Mofetoluwa Adeyemi, Gerald Muriuki, Emmanuel Anebi,
  Chiamaka Chukwuneke, Nkiruka Odu, Eric~Peter Wairagala, Samuel Oyerinde,
  Clemencia Siro, Tobius~Saul Bateesa, Temilola Oloyede, Yvonne Wambui, Victor
  Akinode, Deborah Nabagereka, Maurice Katusiime, Ayodele Awokoya, Mouhamadane
  MBOUP, Dibora Gebreyohannes, Henok Tilaye, Kelechi Nwaike, Degaga Wolde,
  Abdoulaye Faye, Blessing Sibanda, Orevaoghene Ahia, Bonaventure F.~P. Dossou,
  Kelechi Ogueji, Thierno~Ibrahima DIOP, Abdoulaye Diallo, Adewale Akinfaderin,
  Tendai Marengereke, and Salomey Osei.
\newblock {MasakhaNER: Named Entity Recognition for African Languages}.
\newblock \emph{Transactions of the Association for Computational Linguistics},
  9:\penalty0 1116--1131, 10 2021.
\newblock ISSN 2307-387X.
\newblock \doi{10.1162/tacl_a_00416}.
\newblock URL \url{https://doi.org/10.1162/tacl\_a\_00416}.

\bibitem[Aguilar et~al.(2020)Aguilar, Kar, and
  Solorio]{aguilar-etal-2020-lince}
Gustavo Aguilar, Sudipta Kar, and Thamar Solorio.
\newblock {L}in{CE}: {A} {C}entralized {B}enchmark for {L}inguistic
  {C}ode-switching {E}valuation.
\newblock In \emph{Proceedings of The 12th Language Resources and Evaluation
  Conference}, pp.\  1803--1813, Marseille, France, May 2020. European Language
  Resources Association.
\newblock ISBN 979-10-95546-34-4.
\newblock URL \url{https://www.aclweb.org/anthology/2020.lrec-1.223}.

\bibitem[Al-Sallab et~al.(2017)Al-Sallab, Baly, Hajj, Shaban, El-Hajj, and
  Badaro]{10.1145/3086575}
Ahmad Al-Sallab, Ramy Baly, Hazem Hajj, Khaled~Bashir Shaban, Wassim El-Hajj,
  and Gilbert Badaro.
\newblock Aroma: A recursive deep learning model for opinion mining in arabic
  as a low resource language.
\newblock \emph{ACM Trans. Asian Low-Resour. Lang. Inf. Process.}, 16\penalty0
  (4), jul 2017.
\newblock ISSN 2375-4699.
\newblock \doi{10.1145/3086575}.
\newblock URL \url{https://doi.org/10.1145/3086575}.

\bibitem[Amrhein \& Sennrich(2021)Amrhein and
  Sennrich]{amrhein-sennrich-2021-suitable-subword}
Chantal Amrhein and Rico Sennrich.
\newblock How suitable are subword segmentation strategies for translating
  non-concatenative morphology?
\newblock In \emph{Findings of the Association for Computational Linguistics:
  EMNLP 2021}, pp.\  689--705, Punta Cana, Dominican Republic, November 2021.
  Association for Computational Linguistics.
\newblock \doi{10.18653/v1/2021.findings-emnlp.60}.
\newblock URL \url{https://aclanthology.org/2021.findings-emnlp.60}.

\bibitem[Antoun et~al.(2020)Antoun, Baly, and Hajj]{antoun-etal-2020-arabert}
Wissam Antoun, Fady Baly, and Hazem Hajj.
\newblock {A}ra{BERT}: Transformer-based model for {A}rabic language
  understanding.
\newblock In \emph{Proceedings of the 4th Workshop on Open-Source Arabic
  Corpora and Processing Tools, with a Shared Task on Offensive Language
  Detection}, pp.\  9--15, Marseille, France, May 2020. European Language
  Resource Association.
\newblock ISBN 979-10-95546-51-1.
\newblock URL \url{https://aclanthology.org/2020.osact-1.2}.

\bibitem[Artetxe et~al.(2020)Artetxe, Ruder, and
  Yogatama]{artetxe-etal-2020-cross}
Mikel Artetxe, Sebastian Ruder, and Dani Yogatama.
\newblock On the cross-lingual transferability of monolingual representations.
\newblock In \emph{Proceedings of the 58th Annual Meeting of the Association
  for Computational Linguistics}, pp.\  4623--4637, Online, July 2020.
  Association for Computational Linguistics.
\newblock \doi{10.18653/v1/2020.acl-main.421}.
\newblock URL \url{https://aclanthology.org/2020.acl-main.421}.

\bibitem[Asahara et~al.(2018)Asahara, Kanayama, Tanaka, Miyao, Uematsu, Mori,
  Matsumoto, Omura, and Murawaki]{asahara-etal-2018-universal}
Masayuki Asahara, Hiroshi Kanayama, Takaaki Tanaka, Yusuke Miyao, Sumire
  Uematsu, Shinsuke Mori, Yuji Matsumoto, Mai Omura, and Yugo Murawaki.
\newblock {U}niversal {D}ependencies version 2 for {J}apanese.
\newblock In \emph{Proceedings of the Eleventh International Conference on
  Language Resources and Evaluation ({LREC} 2018)}, Miyazaki, Japan, May 2018.
  European Language Resources Association (ELRA).
\newblock URL \url{https://aclanthology.org/L18-1287}.

\bibitem[Ba et~al.(2016)Ba, Kiros, and Hinton]{DBLP:journals/corr/BaKH16}
Lei~Jimmy Ba, Jamie~Ryan Kiros, and Geoffrey~E. Hinton.
\newblock Layer normalization.
\newblock \emph{arXiv preprint}, 2016.
\newblock URL \url{http://arxiv.org/abs/1607.06450}.

\bibitem[Baldwin et~al.(2015)Baldwin, de~Marneffe, Han, Kim, Ritter, and
  Xu]{baldwin-etal-2015-shared}
Timothy Baldwin, Marie~Catherine de~Marneffe, Bo~Han, Young-Bum Kim, Alan
  Ritter, and Wei Xu.
\newblock Shared tasks of the 2015 workshop on noisy user-generated text:
  {T}witter lexical normalization and named entity recognition.
\newblock In \emph{Proceedings of the Workshop on Noisy User-generated Text},
  pp.\  126--135, Beijing, China, July 2015. Association for Computational
  Linguistics.
\newblock \doi{10.18653/v1/W15-4319}.
\newblock URL \url{https://aclanthology.org/W15-4319}.

\bibitem[Bao et~al.(2022)Bao, Dong, Piao, and Wei]{bao2022beit}
Hangbo Bao, Li~Dong, Songhao Piao, and Furu Wei.
\newblock {BE}i{T}: {BERT} pre-training of image transformers.
\newblock In \emph{International Conference on Learning Representations}, 2022.
\newblock URL \url{https://openreview.net/forum?id=p-BhZSz59o4}.

\bibitem[Bland et~al.(2022)Bland, Iyer, and Levchenko]{Bland2022StoryBT}
Maxwell~Troy Bland, Anushya Iyer, and Kirill Levchenko.
\newblock Story beyond the eye: Glyph positions break {PDF} text redaction.
\newblock \emph{arXiv preprint}, 2022.
\newblock URL \url{https://arxiv.org/abs/2206.02285}.

\bibitem[Blevins \& Zettlemoyer(2022)Blevins and
  Zettlemoyer]{blevins-zettlemoyer-2022-language}
Terra Blevins and Luke Zettlemoyer.
\newblock Language contamination explains the cross-lingual capabilities of
  english pretrained models.
\newblock \emph{arXiv preprint}, 2022.
\newblock URL \url{https://doi.org/10.48550/arXiv.2204.08110}.

\bibitem[Blevins et~al.(2022)Blevins, Gonen, and
  Zettlemoyer]{blevins2022analyzing}
Terra Blevins, Hila Gonen, and Luke Zettlemoyer.
\newblock Analyzing the mono-and cross-lingual pretraining dynamics of
  multilingual language models.
\newblock \emph{arXiv preprint}, 2022.
\newblock URL \url{https://arxiv.org/abs/2205.11758}.

\bibitem[Bowman et~al.(2015)Bowman, Angeli, Potts, and
  Manning]{bowman-etal-2015-large}
Samuel~R. Bowman, Gabor Angeli, Christopher Potts, and Christopher~D. Manning.
\newblock A large annotated corpus for learning natural language inference.
\newblock In \emph{Proceedings of the 2015 Conference on Empirical Methods in
  Natural Language Processing}, pp.\  632--642, Lisbon, Portugal, September
  2015. Association for Computational Linguistics.
\newblock \doi{10.18653/v1/D15-1075}.
\newblock URL \url{https://aclanthology.org/D15-1075}.

\bibitem[Broscheit(2018)]{broscheit-2018-learning}
Samuel Broscheit.
\newblock Learning distributional token representations from visual features.
\newblock In \emph{Proceedings of The Third Workshop on Representation Learning
  for {NLP}}, pp.\  187--194, Melbourne, Australia, July 2018. Association for
  Computational Linguistics.
\newblock \doi{10.18653/v1/W18-3025}.
\newblock URL \url{https://aclanthology.org/W18-3025}.

\bibitem[Brown et~al.(2020)Brown, Mann, Ryder, Subbiah, Kaplan, Dhariwal,
  Neelakantan, Shyam, Sastry, Askell, Agarwal, Herbert-Voss, Krueger, Henighan,
  Child, Ramesh, Ziegler, Wu, Winter, Hesse, Chen, Sigler, Litwin, Gray, Chess,
  Clark, Berner, McCandlish, Radford, Sutskever, and
  Amodei]{brown-etal-2020-language}
Tom Brown, Benjamin Mann, Nick Ryder, Melanie Subbiah, Jared~D Kaplan, Prafulla
  Dhariwal, Arvind Neelakantan, Pranav Shyam, Girish Sastry, Amanda Askell,
  Sandhini Agarwal, Ariel Herbert-Voss, Gretchen Krueger, Tom Henighan, Rewon
  Child, Aditya Ramesh, Daniel Ziegler, Jeffrey Wu, Clemens Winter, Chris
  Hesse, Mark Chen, Eric Sigler, Mateusz Litwin, Scott Gray, Benjamin Chess,
  Jack Clark, Christopher Berner, Sam McCandlish, Alec Radford, Ilya Sutskever,
  and Dario Amodei.
\newblock Language models are few-shot learners.
\newblock In H.~Larochelle, M.~Ranzato, R.~Hadsell, M.F. Balcan, and H.~Lin
  (eds.), \emph{Advances in Neural Information Processing Systems}, volume~33,
  pp.\  1877--1901. Curran Associates, Inc., 2020.
\newblock URL
  \url{https://proceedings.neurips.cc/paper/2020/file/1457c0d6bfcb4967418bfb8ac142f64a-Paper.pdf}.

\bibitem[Bugliarello et~al.(2022)Bugliarello, Liu, Pfeiffer, Reddy, Elliott,
  Ponti, and Vuli{\'c}]{bugliarello-etal-2022-iglue}
Emanuele Bugliarello, Fangyu Liu, Jonas Pfeiffer, Siva Reddy, Desmond Elliott,
  Edoardo~Maria Ponti, and Ivan Vuli{\'c}.
\newblock {IGLUE}: {A} benchmark for transfer learning across modalities,
  tasks, and languages.
\newblock In \emph{Proceedings of the 39th International Conference on Machine
  Learning}, Balitmore, MA, July 2022. PMLR.
\newblock URL \url{https://arxiv.org/abs/2201.11732}.

\bibitem[Caswell et~al.(2020)Caswell, Breiner, van Esch, and
  Bapna]{caswell2020language}
Isaac Caswell, Theresa Breiner, Daan van Esch, and Ankur Bapna.
\newblock Language id in the wild: Unexpected challenges on the path to a
  thousand-language web text corpus.
\newblock In \emph{Proceedings of the 28th International Conference on
  Computational Linguistics}, pp.\  6588--6608, Barcelona, Spain (Online),
  2020. Association for Computational Linguistics.
\newblock URL \url{https://aclanthology.org/2020.coling-main.579.pdf}.

\bibitem[Chau et~al.(2020)Chau, Lin, and Smith]{chau-etal-2020-parsing}
Ethan~C. Chau, Lucy~H. Lin, and Noah~A. Smith.
\newblock Parsing with multilingual {BERT}, a small corpus, and a small
  treebank.
\newblock In \emph{Findings of the Association for Computational Linguistics:
  EMNLP 2020}, pp.\  1324--1334, Online, November 2020. Association for
  Computational Linguistics.
\newblock \doi{10.18653/v1/2020.findings-emnlp.118}.
\newblock URL \url{https://aclanthology.org/2020.findings-emnlp.118}.

\bibitem[Chefer et~al.(2021)Chefer, Gur, and Wolf]{Chefer_2021_ICCV}
Hila Chefer, Shir Gur, and Lior Wolf.
\newblock Generic attention-model explainability for interpreting bi-modal and
  encoder-decoder transformers.
\newblock In \emph{Proceedings of the IEEE/CVF International Conference on
  Computer Vision (ICCV)}, pp.\  397--406, October 2021.
\newblock URL
  \url{https://openaccess.thecvf.com/content/ICCV2021/papers/Chefer_Generic_Attention-Model_Explainability_for_Interpreting_Bi-Modal_and_Encoder-Decoder_Transformers_ICCV_2021_paper.pdf}.

\bibitem[Chun et~al.(2018)Chun, Han, Hwang, and Choi]{chun-etal-2018-building}
Jayeol Chun, Na-Rae Han, Jena~D. Hwang, and Jinho~D. Choi.
\newblock Building {U}niversal {D}ependency treebanks in {K}orean.
\newblock In \emph{Proceedings of the Eleventh International Conference on
  Language Resources and Evaluation ({LREC} 2018)}, Miyazaki, Japan, May 2018.
  European Language Resources Association (ELRA).
\newblock URL \url{https://aclanthology.org/L18-1347}.

\bibitem[Chung et~al.(2020)Chung, Garrette, Tan, and
  Riesa]{chung-etal-2020-improving}
Hyung~Won Chung, Dan Garrette, Kiat~Chuan Tan, and Jason Riesa.
\newblock Improving multilingual models with language-clustered vocabularies.
\newblock In \emph{Proceedings of the 2020 Conference on Empirical Methods in
  Natural Language Processing (EMNLP)}, pp.\  4536--4546, Online, November
  2020. Association for Computational Linguistics.
\newblock \doi{10.18653/v1/2020.emnlp-main.367}.
\newblock URL \url{https://aclanthology.org/2020.emnlp-main.367}.

\bibitem[Clark et~al.(2020)Clark, Choi, Collins, Garrette, Kwiatkowski,
  Nikolaev, and Palomaki]{clark-etal-2020-tydi}
Jonathan~H. Clark, Eunsol Choi, Michael Collins, Dan Garrette, Tom Kwiatkowski,
  Vitaly Nikolaev, and Jennimaria Palomaki.
\newblock {T}y{D}i {QA}: A benchmark for information-seeking question answering
  in typologically diverse languages.
\newblock \emph{Transactions of the Association for Computational Linguistics},
  8:\penalty0 454--470, 2020.
\newblock \doi{10.1162/tacl_a_00317}.
\newblock URL \url{https://aclanthology.org/2020.tacl-1.30}.

\bibitem[Clark et~al.(2022)Clark, Garrette, Turc, and
  Wieting]{DBLP:journals/tacl/ClarkGTW22}
Jonathan~H. Clark, Dan Garrette, Iulia Turc, and John Wieting.
\newblock Canine: Pre-training an efficient tokenization-free encoder for
  language representation.
\newblock \emph{Trans. Assoc. Comput. Linguistics}, 10:\penalty0 73--91, 2022.
\newblock \doi{10.1162/tacl\_a\_00448}.
\newblock URL \url{https://doi.org/10.1162/tacl\_a\_00448}.

\bibitem[Conneau et~al.(2018)Conneau, Rinott, Lample, Williams, Bowman,
  Schwenk, and Stoyanov]{conneau-etal-2018-xnli}
Alexis Conneau, Ruty Rinott, Guillaume Lample, Adina Williams, Samuel Bowman,
  Holger Schwenk, and Veselin Stoyanov.
\newblock {XNLI}: Evaluating cross-lingual sentence representations.
\newblock In \emph{Proceedings of the 2018 Conference on Empirical Methods in
  Natural Language Processing}, pp.\  2475--2485, Brussels, Belgium,
  October-November 2018. Association for Computational Linguistics.
\newblock \doi{10.18653/v1/D18-1269}.
\newblock URL \url{https://aclanthology.org/D18-1269}.

\bibitem[Conneau et~al.(2020)Conneau, Khandelwal, Goyal, Chaudhary, Wenzek,
  Guzm{\'a}n, Grave, Ott, Zettlemoyer, and
  Stoyanov]{conneau-etal-2020-unsupervised}
Alexis Conneau, Kartikay Khandelwal, Naman Goyal, Vishrav Chaudhary, Guillaume
  Wenzek, Francisco Guzm{\'a}n, Edouard Grave, Myle Ott, Luke Zettlemoyer, and
  Veselin Stoyanov.
\newblock Unsupervised cross-lingual representation learning at scale.
\newblock In \emph{Proceedings of the 58th Annual Meeting of the Association
  for Computational Linguistics}, pp.\  8440--8451, Online, July 2020.
  Association for Computational Linguistics.
\newblock \doi{10.18653/v1/2020.acl-main.747}.
\newblock URL \url{https://aclanthology.org/2020.acl-main.747}.

\bibitem[Devlin et~al.(2019)Devlin, Chang, Lee, and
  Toutanova]{devlin-etal-2019-bert}
Jacob Devlin, Ming-Wei Chang, Kenton Lee, and Kristina Toutanova.
\newblock {BERT}: Pre-training of deep bidirectional transformers for language
  understanding.
\newblock In \emph{Proceedings of the 2019 Conference of the North {A}merican
  Chapter of the Association for Computational Linguistics: Human Language
  Technologies, Volume 1 (Long and Short Papers)}, pp.\  4171--4186,
  Minneapolis, Minnesota, June 2019. Association for Computational Linguistics.
\newblock \doi{10.18653/v1/N19-1423}.
\newblock URL \url{https://aclanthology.org/N19-1423}.

\bibitem[Dosovitskiy et~al.(2021)Dosovitskiy, Beyer, Kolesnikov, Weissenborn,
  Zhai, Unterthiner, Dehghani, Minderer, Heigold, Gelly, Uszkoreit, and
  Houlsby]{dosovitskiy2021an}
Alexey Dosovitskiy, Lucas Beyer, Alexander Kolesnikov, Dirk Weissenborn,
  Xiaohua Zhai, Thomas Unterthiner, Mostafa Dehghani, Matthias Minderer, Georg
  Heigold, Sylvain Gelly, Jakob Uszkoreit, and Neil Houlsby.
\newblock An image is worth 16x16 words: Transformers for image recognition at
  scale.
\newblock In \emph{International Conference on Learning Representations}, 2021.
\newblock URL \url{https://openreview.net/forum?id=YicbFdNTTy}.

\bibitem[Dozat \& Manning(2017)Dozat and Manning]{DBLP:conf/iclr/DozatM17}
Timothy Dozat and Christopher~D. Manning.
\newblock Deep biaffine attention for neural dependency parsing.
\newblock In \emph{5th International Conference on Learning Representations,
  {ICLR} 2017, Toulon, France, April 24-26, 2017, Conference Track
  Proceedings}. OpenReview.net, 2017.
\newblock URL \url{https://openreview.net/forum?id=Hk95PK9le}.

\bibitem[Ebrahimi \& Kann(2021)Ebrahimi and Kann]{ebrahimi-kann-2021-adapt}
Abteen Ebrahimi and Katharina Kann.
\newblock How to adapt your pretrained multilingual model to 1600 languages.
\newblock In \emph{Proceedings of the 59th Annual Meeting of the Association
  for Computational Linguistics and the 11th International Joint Conference on
  Natural Language Processing (Volume 1: Long Papers)}, pp.\  4555--4567,
  Online, August 2021. Association for Computational Linguistics.
\newblock \doi{10.18653/v1/2021.acl-long.351}.
\newblock URL \url{https://aclanthology.org/2021.acl-long.351}.

\bibitem[Eger \& Benz(2020)Eger and Benz]{eger-benz-2020-hero}
Steffen Eger and Yannik Benz.
\newblock From hero to z{\'e}roe: A benchmark of low-level adversarial attacks.
\newblock In \emph{Proceedings of the 1st Conference of the Asia-Pacific
  Chapter of the Association for Computational Linguistics and the 10th
  International Joint Conference on Natural Language Processing}, pp.\
  786--803, Suzhou, China, December 2020. Association for Computational
  Linguistics.
\newblock URL \url{https://aclanthology.org/2020.aacl-main.79}.

\bibitem[Gao et~al.(2020)Gao, Biderman, Black, Golding, Hoppe, Foster, Phang,
  He, Thite, Nabeshima, et~al.]{gao2020pile}
Leo Gao, Stella Biderman, Sid Black, Laurence Golding, Travis Hoppe, Charles
  Foster, Jason Phang, Horace He, Anish Thite, Noa Nabeshima, et~al.
\newblock The pile: An 800gb dataset of diverse text for language modeling.
\newblock \emph{arXiv preprint}, 2020.
\newblock URL \url{https://arxiv.org/abs/2101.00027}.

\bibitem[Glava{\v{s}} \& Vuli{\'c}(2021)Glava{\v{s}} and
  Vuli{\'c}]{glavas-vulic-2021-supervised}
Goran Glava{\v{s}} and Ivan Vuli{\'c}.
\newblock Is supervised syntactic parsing beneficial for language understanding
  tasks? an empirical investigation.
\newblock In \emph{Proceedings of the 16th Conference of the European Chapter
  of the Association for Computational Linguistics: Main Volume}, pp.\
  3090--3104, Online, April 2021. Association for Computational Linguistics.
\newblock \doi{10.18653/v1/2021.eacl-main.270}.
\newblock URL \url{https://aclanthology.org/2021.eacl-main.270}.

\bibitem[Haji{\v{c}} et~al.(2009)Haji{\v{c}}, Smr{\v{z}}, Zem{\'a}nek, Pajas,
  {\v{S}}naidauf, Be{\v{s}}ka, Kracmar, and Hassanov{\'a}]{hajivc2009prague}
Jan Haji{\v{c}}, Otakar Smr{\v{z}}, Petr Zem{\'a}nek, Petr Pajas, Jan
  {\v{S}}naidauf, Emanuel Be{\v{s}}ka, Jakub Kracmar, and Kamila Hassanov{\'a}.
\newblock Prague arabic dependency treebank 1.0, 2009.
\newblock URL \url{https://ufal.mff.cuni.cz/padt/PADT_1.0/docs/index.html}.

\bibitem[He et~al.(2022)He, Chen, Xie, Li, Doll{\'a}r, and
  Girshick]{he-etal-2022-mae}
Kaiming He, Xinlei Chen, Saining Xie, Yanghao Li, Piotr Doll{\'a}r, and Ross
  Girshick.
\newblock Masked autoencoders are scalable vision learners.
\newblock In \emph{Proceedings of the IEEE/CVF Conference on Computer Vision
  and Pattern Recognition}, pp.\  16000--16009, 2022.
\newblock URL
  \url{https://openaccess.thecvf.com/content/CVPR2022/papers/He_Masked_Autoencoders_Are_Scalable_Vision_Learners_CVPR_2022_paper.pdf}.

\bibitem[He et~al.(2021{\natexlab{a}})He, Gao, and
  Chen]{he-etal-2021-debertav3}
Pengcheng He, Jianfeng Gao, and Weizhu Chen.
\newblock Debertav3: Improving deberta using electra-style pre-training with
  gradient-disentangled embedding sharing.
\newblock \emph{arXiv preprint}, 2021{\natexlab{a}}.
\newblock URL \url{https://arxiv.org/abs/2111.09543}.

\bibitem[He et~al.(2021{\natexlab{b}})He, Liu, Gao, and
  Chen]{he-etal-2020-deberta}
Pengcheng He, Xiaodong Liu, Jianfeng Gao, and Weizhu Chen.
\newblock Deberta: Decoding-enhanced {BERT} with disentangled attention.
\newblock In \emph{International Conference on Learning Representations},
  2021{\natexlab{b}}.
\newblock URL \url{https://openreview.net/forum?id=XPZIaotutsD}.

\bibitem[Hendrycks \& Gimpel(2016)Hendrycks and
  Gimpel]{DBLP:journals/corr/HendrycksG16}
Dan Hendrycks and Kevin Gimpel.
\newblock Bridging nonlinearities and stochastic regularizers with gaussian
  error linear units.
\newblock \emph{arXiv preprint}, 2016.
\newblock URL \url{http://arxiv.org/abs/1606.08415}.

\bibitem[Hirschberg \& Manning(2015)Hirschberg and
  Manning]{hirschberg2015advances}
Julia Hirschberg and Christopher~D Manning.
\newblock Advances in natural language processing.
\newblock \emph{Science}, 349\penalty0 (6245):\penalty0 261--266, 2015.
\newblock URL \url{https://cs224d.stanford.edu/papers/advances.pdf}.

\bibitem[Joshi(1982)]{joshi-1982-processing}
Aravind~K. Joshi.
\newblock Processing of sentences with intra-sentential code-switching.
\newblock In \emph{{C}oling 1982: Proceedings of the {N}inth {I}nternational
  {C}onference on {C}omputational {L}inguistics}, 1982.
\newblock URL \url{https://aclanthology.org/C82-1023}.

\bibitem[Joshi et~al.(2020{\natexlab{a}})Joshi, Chen, Liu, Weld, Zettlemoyer,
  and Levy]{joshi-etal-2020-spanbert}
Mandar Joshi, Danqi Chen, Yinhan Liu, Daniel~S. Weld, Luke Zettlemoyer, and
  Omer Levy.
\newblock {S}pan{BERT}: Improving pre-training by representing and predicting
  spans.
\newblock \emph{Transactions of the Association for Computational Linguistics},
  8:\penalty0 64--77, 2020{\natexlab{a}}.
\newblock \doi{10.1162/tacl_a_00300}.
\newblock URL \url{https://aclanthology.org/2020.tacl-1.5}.

\bibitem[Joshi et~al.(2020{\natexlab{b}})Joshi, Santy, Budhiraja, Bali, and
  Choudhury]{joshi-etal-2020-state}
Pratik Joshi, Sebastin Santy, Amar Budhiraja, Kalika Bali, and Monojit
  Choudhury.
\newblock The state and fate of linguistic diversity and inclusion in the {NLP}
  world.
\newblock In \emph{Proceedings of the 58th Annual Meeting of the Association
  for Computational Linguistics}, pp.\  6282--6293, Online, July
  2020{\natexlab{b}}. Association for Computational Linguistics.
\newblock \doi{10.18653/v1/2020.acl-main.560}.
\newblock URL \url{https://aclanthology.org/2020.acl-main.560}.

\bibitem[Keller et~al.(2021)Keller, Mackensen, and Eger]{keller-etal-2021-bert}
Yannik Keller, Jan Mackensen, and Steffen Eger.
\newblock {BERT}-defense: A probabilistic model based on {BERT} to combat
  cognitively inspired orthographic adversarial attacks.
\newblock In \emph{Findings of the Association for Computational Linguistics:
  ACL-IJCNLP 2021}, pp.\  1616--1629, Online, August 2021. Association for
  Computational Linguistics.
\newblock \doi{10.18653/v1/2021.findings-acl.141}.
\newblock URL \url{https://aclanthology.org/2021.findings-acl.141}.

\bibitem[Keren et~al.(2022)Keren, Avinari, Tsarfaty, and
  Levy]{keren-etal-2022-breaking}
Omri Keren, Tal Avinari, Reut Tsarfaty, and Omer Levy.
\newblock Breaking character: Are subwords good enough for mrls after all?
\newblock \emph{arXiv preprint}, 2022.
\newblock URL \url{https://arxiv.org/abs/2204.04748}.

\bibitem[Kingma \& Ba(2015)Kingma and Ba]{kingma-ba-2015-adam}
Diederik~P. Kingma and Jimmy Ba.
\newblock Adam: {A} method for stochastic optimization.
\newblock In Yoshua Bengio and Yann LeCun (eds.), \emph{Proceedings of the 3rd
  International Conference on Learning Representations ({ICLR})}, San Diego,
  CA, USA, 2015.
\newblock URL \url{http://arxiv.org/abs/1412.6980}.

\bibitem[Klein \& Tsarfaty(2020)Klein and
  Tsarfaty]{klein-tsarfaty-2020-getting}
Stav Klein and Reut Tsarfaty.
\newblock Getting the {\#}{\#}life out of living: How adequate are word-pieces
  for modelling complex morphology?
\newblock In \emph{Proceedings of the 17th SIGMORPHON Workshop on Computational
  Research in Phonetics, Phonology, and Morphology}, pp.\  204--209, Online,
  July 2020. Association for Computational Linguistics.
\newblock \doi{10.18653/v1/2020.sigmorphon-1.24}.
\newblock URL \url{https://aclanthology.org/2020.sigmorphon-1.24}.

\bibitem[Kolesnikov et~al.(2020)Kolesnikov, Beyer, Zhai, Puigcerver, Yung,
  Gelly, and Houlsby]{kolesnikov-etal-2020}
Alexander Kolesnikov, Lucas Beyer, Xiaohua Zhai, Joan Puigcerver, Jessica Yung,
  Sylvain Gelly, and Neil Houlsby.
\newblock Big transfer (bit): General visual representation learning.
\newblock In Andrea Vedaldi, Horst Bischof, Thomas Brox, and Jan-Michael Frahm
  (eds.), \emph{Computer Vision -- ECCV 2020}, pp.\  491--507, Cham, 2020.
  Springer International Publishing.
\newblock ISBN 978-3-030-58558-7.
\newblock URL
  \url{https://link.springer.com/chapter/10.1007/978-3-030-58558-7_29}.

\bibitem[Kudo(2018)]{kudo-2018-subword}
Taku Kudo.
\newblock Subword regularization: Improving neural network translation models
  with multiple subword candidates.
\newblock In \emph{Proceedings of the 56th Annual Meeting of the Association
  for Computational Linguistics (Volume 1: Long Papers)}, pp.\  66--75,
  Melbourne, Australia, July 2018. Association for Computational Linguistics.
\newblock \doi{10.18653/v1/P18-1007}.
\newblock URL \url{https://aclanthology.org/P18-1007}.

\bibitem[Kudo \& Richardson(2018)Kudo and
  Richardson]{kudo-richardson-2018-sentencepiece}
Taku Kudo and John Richardson.
\newblock {S}entence{P}iece: A simple and language independent subword
  tokenizer and detokenizer for neural text processing.
\newblock In \emph{Proceedings of the 2018 Conference on Empirical Methods in
  Natural Language Processing: System Demonstrations}, pp.\  66--71, Brussels,
  Belgium, November 2018. Association for Computational Linguistics.
\newblock \doi{10.18653/v1/D18-2012}.
\newblock URL \url{https://aclanthology.org/D18-2012}.

\bibitem[Lauscher et~al.(2020)Lauscher, Ravishankar, Vuli{\'c}, and
  Glava{\v{s}}]{lauscher-etal-2020-zero}
Anne Lauscher, Vinit Ravishankar, Ivan Vuli{\'c}, and Goran Glava{\v{s}}.
\newblock From zero to hero: {O}n the limitations of zero-shot language
  transfer with multilingual {T}ransformers.
\newblock In \emph{Proceedings of the 2020 Conference on Empirical Methods in
  Natural Language Processing (EMNLP)}, pp.\  4483--4499, Online, November
  2020. Association for Computational Linguistics.
\newblock \doi{10.18653/v1/2020.emnlp-main.363}.
\newblock URL \url{https://aclanthology.org/2020.emnlp-main.363}.

\bibitem[Lee et~al.(2017)Lee, Cho, and Hofmann]{lee-etal-2017-fully}
Jason Lee, Kyunghyun Cho, and Thomas Hofmann.
\newblock Fully character-level neural machine translation without explicit
  segmentation.
\newblock \emph{Transactions of the Association for Computational Linguistics},
  5:\penalty0 365--378, 2017.
\newblock \doi{10.1162/tacl_a_00067}.
\newblock URL \url{https://aclanthology.org/Q17-1026}.

\bibitem[Liang et~al.(2022)Liang, Li, and
  Marculescu]{https://doi.org/10.48550/arxiv.2205.14540}
Feng Liang, Yangguang Li, and Diana Marculescu.
\newblock Supmae: Supervised masked autoencoders are efficient vision learners.
\newblock \emph{arXiv preprint}, 2022.
\newblock URL \url{https://arxiv.org/abs/2205.14540}.

\bibitem[Lim et~al.(2019)Lim, Kim, and Lee]{lim-etal-2019-korquad}
Seungyoung Lim, Myungji Kim, and Jooyoul Lee.
\newblock Korquad1.0: Korean {QA} dataset for machine reading comprehension.
\newblock \emph{arXiv preprint}, 2019.
\newblock URL \url{http://arxiv.org/abs/1909.07005}.

\bibitem[Liu et~al.(2019)Liu, Ott, Goyal, Du, Joshi, Chen, Levy, Lewis,
  Zettlemoyer, and Stoyanov]{DBLP:journals/corr/abs-1907-11692}
Yinhan Liu, Myle Ott, Naman Goyal, Jingfei Du, Mandar Joshi, Danqi Chen, Omer
  Levy, Mike Lewis, Luke Zettlemoyer, and Veselin Stoyanov.
\newblock Roberta: {A} robustly optimized {BERT} pretraining approach.
\newblock \emph{arXiv preprint}, 2019.
\newblock URL \url{http://arxiv.org/abs/1907.11692}.

\bibitem[Loshchilov \& Hutter(2017)Loshchilov and
  Hutter]{DBLP:conf/iclr/LoshchilovH17}
Ilya Loshchilov and Frank Hutter.
\newblock {SGDR:} stochastic gradient descent with warm restarts.
\newblock In \emph{5th International Conference on Learning Representations,
  {ICLR} 2017, Toulon, France, April 24-26, 2017, Conference Track
  Proceedings}. OpenReview.net, 2017.
\newblock URL \url{https://openreview.net/forum?id=Skq89Scxx}.

\bibitem[Loshchilov \& Hutter(2019)Loshchilov and
  Hutter]{loshchilov2018decoupled}
Ilya Loshchilov and Frank Hutter.
\newblock Decoupled weight decay regularization.
\newblock In \emph{Proceedings of the 7th International Conference on Learning
  Representations ({ICLR})}, New Orleans, LA, USA, 2019. OpenReview.net.
\newblock URL \url{https://openreview.net/forum?id=Bkg6RiCqY7}.

\bibitem[Ma et~al.(2020)Ma, Cui, Si, Liu, Wang, and Hu]{ma-etal-2020-charbert}
Wentao Ma, Yiming Cui, Chenglei Si, Ting Liu, Shijin Wang, and Guoping Hu.
\newblock {C}har{BERT}: Character-aware pre-trained language model.
\newblock In \emph{Proceedings of the 28th International Conference on
  Computational Linguistics}, pp.\  39--50, Barcelona, Spain (Online), December
  2020. International Committee on Computational Linguistics.
\newblock \doi{10.18653/v1/2020.coling-main.4}.
\newblock URL \url{https://aclanthology.org/2020.coling-main.4}.

\bibitem[Mansimov et~al.(2020)Mansimov, Stern, Chen, Firat, Uszkoreit, and
  Jain]{mansimov-etal-2020-towards}
Elman Mansimov, Mitchell Stern, Mia Chen, Orhan Firat, Jakob Uszkoreit, and
  Puneet Jain.
\newblock Towards end-to-end in-image neural machine translation.
\newblock In \emph{Proceedings of the First International Workshop on Natural
  Language Processing Beyond Text}, pp.\  70--74, Online, November 2020.
  Association for Computational Linguistics.
\newblock \doi{10.18653/v1/2020.nlpbt-1.8}.
\newblock URL \url{https://aclanthology.org/2020.nlpbt-1.8}.

\bibitem[Maronikolakis et~al.(2021)Maronikolakis, Dufter, and
  Sch{\"u}tze]{maronikolakis-etal-2021-wine-v}
Antonis Maronikolakis, Philipp Dufter, and Hinrich Sch{\"u}tze.
\newblock Wine is not v i n. on the compatibility of tokenizations across
  languages.
\newblock In \emph{Findings of the Association for Computational Linguistics:
  EMNLP 2021}, pp.\  2382--2399, Punta Cana, Dominican Republic, November 2021.
  Association for Computational Linguistics.
\newblock \doi{10.18653/v1/2021.findings-emnlp.205}.
\newblock URL \url{https://aclanthology.org/2021.findings-emnlp.205}.

\bibitem[Mielke et~al.(2021)Mielke, Alyafeai, Salesky, Raffel, Dey,
  Gall{\'{e}}, Raja, Si, Lee, Sagot, and
  Tan]{DBLP:journals/corr/abs-2112-10508}
Sabrina~J. Mielke, Zaid Alyafeai, Elizabeth Salesky, Colin Raffel, Manan Dey,
  Matthias Gall{\'{e}}, Arun Raja, Chenglei Si, Wilson~Y. Lee, Beno{\^{\i}}t
  Sagot, and Samson Tan.
\newblock Between words and characters: {A} brief history of open-vocabulary
  modeling and tokenization in {NLP}.
\newblock \emph{arXiv preprint}, 2021.
\newblock URL \url{https://arxiv.org/abs/2112.10508}.

\bibitem[Moosa et~al.(2022)Moosa, Akhter, and
  Habib]{DBLP:journals/corr/abs-2201-12501}
Ibraheem~Muhammad Moosa, Mahmud~Elahi Akhter, and Ashfia~Binte Habib.
\newblock Does transliteration help multilingual language modeling?
\newblock \emph{arXiv preprint}, 2022.
\newblock URL \url{https://arxiv.org/abs/2201.12501}.

\bibitem[Nguyen et~al.(2009)Nguyen, Vu, Nguyen, Nguyen, and
  Le]{nguyen-etal-2009-building}
Phuong-Thai Nguyen, Xuan-Luong Vu, Thi-Minh-Huyen Nguyen, Van-Hiep Nguyen, and
  Hong-Phuong Le.
\newblock Building a large syntactically-annotated corpus of {V}ietnamese.
\newblock In \emph{Proceedings of the Third Linguistic Annotation Workshop
  ({LAW} {III})}, pp.\  182--185, Suntec, Singapore, August 2009. Association
  for Computational Linguistics.
\newblock URL \url{https://aclanthology.org/W09-3035}.

\bibitem[Nivre et~al.(2020)Nivre, de~Marneffe, Ginter, Haji{\v{c}}, Manning,
  Pyysalo, Schuster, Tyers, and Zeman]{nivre-etal-2020-universal}
Joakim Nivre, Marie-Catherine de~Marneffe, Filip Ginter, Jan Haji{\v{c}},
  Christopher~D. Manning, Sampo Pyysalo, Sebastian Schuster, Francis Tyers, and
  Daniel Zeman.
\newblock {U}niversal {D}ependencies v2: An evergrowing multilingual treebank
  collection.
\newblock In \emph{Proceedings of the 12th Language Resources and Evaluation
  Conference}, pp.\  4034--4043, Marseille, France, May 2020. European Language
  Resources Association.
\newblock ISBN 979-10-95546-34-4.
\newblock URL \url{https://aclanthology.org/2020.lrec-1.497}.

\bibitem[Palmer et~al.(2009)Palmer, Rambow, Bhatt, Sharma, Narasimhan, and
  Xia]{Palmer2009HindiSA}
Martha Palmer, Owen Rambow, Rajesh Bhatt, Dipti~Misra Sharma, Bhuvana
  Narasimhan, and F.~Xia.
\newblock Hindi syntax: Annotating dependency, lexical predicate-argument
  structure, and phrase structure.
\newblock In \emph{Proceedings of ICON-2009: 7th International Conference on
  Natural Language Processing}, India, 2009. Macmillan Publishers.
\newblock URL
  \url{http://cdn.iiit.ac.in/cdn/ltrc.iiit.ac.in/hutb_release/related_publications/ICON09.pdf}.

\bibitem[Paszke et~al.(2019)Paszke, Gross, Massa, Lerer, Bradbury, Chanan,
  Killeen, Lin, Gimelshein, Antiga, Desmaison, K{\"{o}}pf, Yang, DeVito,
  Raison, Tejani, Chilamkurthy, Steiner, Fang, Bai, and
  Chintala]{DBLP:conf/nips/PaszkeGMLBCKLGA19}
Adam Paszke, Sam Gross, Francisco Massa, Adam Lerer, James Bradbury, Gregory
  Chanan, Trevor Killeen, Zeming Lin, Natalia Gimelshein, Luca Antiga, Alban
  Desmaison, Andreas K{\"{o}}pf, Edward~Z. Yang, Zachary DeVito, Martin Raison,
  Alykhan Tejani, Sasank Chilamkurthy, Benoit Steiner, Lu~Fang, Junjie Bai, and
  Soumith Chintala.
\newblock Pytorch: An imperative style, high-performance deep learning library.
\newblock In Hanna~M. Wallach, Hugo Larochelle, Alina Beygelzimer, Florence
  d'Alch{\'{e}}{-}Buc, Emily~B. Fox, and Roman Garnett (eds.), \emph{Advances
  in Neural Information Processing Systems 32: Annual Conference on Neural
  Information Processing Systems 2019, NeurIPS 2019, December 8-14, 2019,
  Vancouver, BC, Canada}, pp.\  8024--8035, 2019.
\newblock URL
  \url{https://proceedings.neurips.cc/paper/2019/hash/bdbca288fee7f92f2bfa9f7012727740-Abstract.html}.

\bibitem[Patil et~al.(2022)Patil, Talukdar, and
  Sarawagi]{patil-etal-2022-overlap}
Vaidehi Patil, Partha Talukdar, and Sunita Sarawagi.
\newblock Overlap-based vocabulary generation improves cross-lingual transfer
  among related languages.
\newblock In \emph{Proceedings of the 60th Annual Meeting of the Association
  for Computational Linguistics (Volume 1: Long Papers)}, pp.\  219--233,
  Dublin, Ireland, May 2022. Association for Computational Linguistics.
\newblock \doi{10.18653/v1/2022.acl-long.18}.
\newblock URL \url{https://aclanthology.org/2022.acl-long.18}.

\bibitem[Pfeiffer et~al.(2021)Pfeiffer, Vuli{\'c}, Gurevych, and
  Ruder]{pfeiffer-etal-2021-unks}
Jonas Pfeiffer, Ivan Vuli{\'c}, Iryna Gurevych, and Sebastian Ruder.
\newblock {UNK}s everywhere: {A}dapting multilingual language models to new
  scripts.
\newblock In \emph{Proceedings of the 2021 Conference on Empirical Methods in
  Natural Language Processing}, pp.\  10186--10203, Online and Punta Cana,
  Dominican Republic, November 2021. Association for Computational Linguistics.
\newblock \doi{10.18653/v1/2021.emnlp-main.800}.
\newblock URL \url{https://aclanthology.org/2021.emnlp-main.800}.

\bibitem[Pires et~al.(2019)Pires, Schlinger, and
  Garrette]{pires-etal-2019-multilingual}
Telmo Pires, Eva Schlinger, and Dan Garrette.
\newblock How multilingual is multilingual {BERT}?
\newblock In \emph{Proceedings of the 57th Annual Meeting of the Association
  for Computational Linguistics}, pp.\  4996--5001, Florence, Italy, July 2019.
  Association for Computational Linguistics.
\newblock \doi{10.18653/v1/P19-1493}.
\newblock URL \url{https://aclanthology.org/P19-1493}.

\bibitem[Pruthi et~al.(2019)Pruthi, Dhingra, and
  Lipton]{pruthi-etal-2019-combating}
Danish Pruthi, Bhuwan Dhingra, and Zachary~C. Lipton.
\newblock Combating adversarial misspellings with robust word recognition.
\newblock In \emph{Proceedings of the 57th Annual Meeting of the Association
  for Computational Linguistics}, pp.\  5582--5591, Florence, Italy, July 2019.
  Association for Computational Linguistics.
\newblock \doi{10.18653/v1/P19-1561}.
\newblock URL \url{https://aclanthology.org/P19-1561}.

\bibitem[Radford et~al.(2021)Radford, Kim, Hallacy, Ramesh, Goh, Agarwal,
  Sastry, Askell, Mishkin, Clark, Krueger, and
  Sutskever]{DBLP:conf/icml/RadfordKHRGASAM21}
Alec Radford, Jong~Wook Kim, Chris Hallacy, Aditya Ramesh, Gabriel Goh,
  Sandhini Agarwal, Girish Sastry, Amanda Askell, Pamela Mishkin, Jack Clark,
  Gretchen Krueger, and Ilya Sutskever.
\newblock Learning transferable visual models from natural language
  supervision.
\newblock In Marina Meila and Tong Zhang (eds.), \emph{Proceedings of the 38th
  International Conference on Machine Learning, {ICML} 2021, 18-24 July 2021,
  Virtual Event}, volume 139 of \emph{Proceedings of Machine Learning
  Research}, pp.\  8748--8763. {PMLR}, 2021.
\newblock URL \url{http://proceedings.mlr.press/v139/radford21a.html}.

\bibitem[Raffel et~al.(2020)Raffel, Shazeer, Roberts, Lee, Narang, Matena,
  Zhou, Li, and Liu]{raffel-etal-2020-t5}
Colin Raffel, Noam Shazeer, Adam Roberts, Katherine Lee, Sharan Narang, Michael
  Matena, Yanqi Zhou, Wei Li, and Peter~J. Liu.
\newblock Exploring the limits of transfer learning with a unified text-to-text
  transformer.
\newblock \emph{Journal of Machine Learning Research}, 21\penalty0
  (140):\penalty0 1--67, 2020.
\newblock URL \url{http://jmlr.org/papers/v21/20-074.html}.

\bibitem[Rajpurkar et~al.(2016)Rajpurkar, Zhang, Lopyrev, and
  Liang]{rajpurkar-etal-2016-squad}
Pranav Rajpurkar, Jian Zhang, Konstantin Lopyrev, and Percy Liang.
\newblock {SQ}u{AD}: 100,000+ questions for machine comprehension of text.
\newblock In \emph{Proceedings of the 2016 Conference on Empirical Methods in
  Natural Language Processing}, pp.\  2383--2392, Austin, Texas, November 2016.
  Association for Computational Linguistics.
\newblock \doi{10.18653/v1/D16-1264}.
\newblock URL \url{https://aclanthology.org/D16-1264}.

\bibitem[Ramasamy \& {\v{Z}}abokrtsk{\'y}(2012)Ramasamy and
  {\v{Z}}abokrtsk{\'y}]{ramasamy-zabokrtsky-2012-prague}
Loganathan Ramasamy and Zden{\v{e}}k {\v{Z}}abokrtsk{\'y}.
\newblock {P}rague dependency style treebank for {T}amil.
\newblock In \emph{Proceedings of the Eighth International Conference on
  Language Resources and Evaluation ({LREC}'12)}, pp.\  1888--1894, Istanbul,
  Turkey, May 2012. European Language Resources Association (ELRA).
\newblock URL
  \url{http://www.lrec-conf.org/proceedings/lrec2012/pdf/456_Paper.pdf}.

\bibitem[Rust et~al.(2021)Rust, Pfeiffer, Vuli{\'c}, Ruder, and
  Gurevych]{rust-etal-2021-good}
Phillip Rust, Jonas Pfeiffer, Ivan Vuli{\'c}, Sebastian Ruder, and Iryna
  Gurevych.
\newblock How good is your tokenizer? on the monolingual performance of
  multilingual language models.
\newblock In \emph{Proceedings of the 59th Annual Meeting of the Association
  for Computational Linguistics and the 11th International Joint Conference on
  Natural Language Processing (Volume 1: Long Papers)}, pp.\  3118--3135,
  Online, August 2021. Association for Computational Linguistics.
\newblock \doi{10.18653/v1/2021.acl-long.243}.
\newblock URL \url{https://aclanthology.org/2021.acl-long.243}.

\bibitem[Salesky et~al.(2021)Salesky, Etter, and
  Post]{salesky-etal-2021-robust}
Elizabeth Salesky, David Etter, and Matt Post.
\newblock Robust open-vocabulary translation from visual text representations.
\newblock In \emph{Proceedings of the 2021 Conference on Empirical Methods in
  Natural Language Processing}, pp.\  7235--7252, Online and Punta Cana,
  Dominican Republic, November 2021. Association for Computational Linguistics.
\newblock \doi{10.18653/v1/2021.emnlp-main.576}.
\newblock URL \url{https://aclanthology.org/2021.emnlp-main.576}.

\bibitem[Sellam et~al.(2022)Sellam, Yadlowsky, Tenney, Wei, Saphra, D'Amour,
  Linzen, Bastings, Turc, Eisenstein, Das, and Pavlick]{sellam2022the}
Thibault Sellam, Steve Yadlowsky, Ian Tenney, Jason Wei, Naomi Saphra,
  Alexander D'Amour, Tal Linzen, Jasmijn Bastings, Iulia~Raluca Turc, Jacob
  Eisenstein, Dipanjan Das, and Ellie Pavlick.
\newblock The multi{BERT}s: {BERT} reproductions for robustness analysis.
\newblock In \emph{International Conference on Learning Representations}, 2022.
\newblock URL \url{https://openreview.net/forum?id=K0E_F0gFDgA}.

\bibitem[Sennrich et~al.(2016)Sennrich, Haddow, and
  Birch]{sennrich-etal-2016-neural}
Rico Sennrich, Barry Haddow, and Alexandra Birch.
\newblock Neural machine translation of rare words with subword units.
\newblock In \emph{Proceedings of the 54th Annual Meeting of the Association
  for Computational Linguistics (Volume 1: Long Papers)}, pp.\  1715--1725,
  Berlin, Germany, August 2016. Association for Computational Linguistics.
\newblock \doi{10.18653/v1/P16-1162}.
\newblock URL \url{https://aclanthology.org/P16-1162}.

\bibitem[Shen et~al.(2016)Shen, McDonald, Zeman, and
  Qi]{shen-etal-2016-chinese}
Mo~Shen, Ryan McDonald, Daniel Zeman, and Peng Qi.
\newblock Ud\_chinese-gsd.
\newblock \emph{GitHub repository}, 2016.
\newblock URL \url{https://github.com/UniversalDependencies/UD_Chinese-GSD}.

\bibitem[Silveira et~al.(2014)Silveira, Dozat, de~Marneffe, Bowman, Connor,
  Bauer, and Manning]{silveira-etal-2014-gold}
Natalia Silveira, Timothy Dozat, Marie-Catherine de~Marneffe, Samuel Bowman,
  Miriam Connor, John Bauer, and Chris Manning.
\newblock A gold standard dependency corpus for {E}nglish.
\newblock In \emph{Proceedings of the Ninth International Conference on
  Language Resources and Evaluation ({LREC}'14)}, pp.\  2897--2904, Reykjavik,
  Iceland, May 2014. European Language Resources Association (ELRA).
\newblock URL
  \url{http://www.lrec-conf.org/proceedings/lrec2014/pdf/1089_Paper.pdf}.

\bibitem[Smit et~al.(2014)Smit, Virpioja, Gr{\"o}nroos, and
  Kurimo]{smit-etal-2014-morfessor}
Peter Smit, Sami Virpioja, Stig-Arne Gr{\"o}nroos, and Mikko Kurimo.
\newblock {M}orfessor 2.0: Toolkit for statistical morphological segmentation.
\newblock In \emph{Proceedings of the Demonstrations at the 14th Conference of
  the {E}uropean Chapter of the Association for Computational Linguistics},
  pp.\  21--24, Gothenburg, Sweden, April 2014. Association for Computational
  Linguistics.
\newblock \doi{10.3115/v1/E14-2006}.
\newblock URL \url{https://aclanthology.org/E14-2006}.

\bibitem[So et~al.(2022)So, Byun, Kang, and Cho]{so2022jaquad}
ByungHoon So, Kyuhong Byun, Kyungwon Kang, and Seongjin Cho.
\newblock {JaQuAD}: Japanese question answering dataset for machine reading
  comprehension.
\newblock \emph{ArXiv preprint}, 2022.
\newblock URL \url{https://arxiv.org/abs/2202.01764}.

\bibitem[Socher et~al.(2013)Socher, Perelygin, Wu, Chuang, Manning, Ng, and
  Potts]{socher-etal-2013-recursive}
Richard Socher, Alex Perelygin, Jean Wu, Jason Chuang, Christopher~D. Manning,
  Andrew Ng, and Christopher Potts.
\newblock Recursive deep models for semantic compositionality over a sentiment
  treebank.
\newblock In \emph{Proceedings of the 2013 Conference on Empirical Methods in
  Natural Language Processing}, pp.\  1631--1642, Seattle, Washington, USA,
  October 2013. Association for Computational Linguistics.
\newblock URL \url{https://aclanthology.org/D13-1170}.

\bibitem[Sun et~al.(2019)Sun, Yang, Chi, Zhang, and Lin]{Sun2019SquaredEW}
Baohua Sun, Lin Yang, Catherine Chi, Wenhan Zhang, and Michael Lin.
\newblock Squared english word: A method of generating glyph to use super
  characters for sentiment analysis.
\newblock In \emph{AffCon@AAAI}, 2019.
\newblock URL \url{https://arxiv.org/abs/1902.02160}.

\bibitem[Sun et~al.(2020)Sun, Hashimoto, Yin, Asai, Li, Yu, and
  Xiong]{SunADV-BERT}
Lichao Sun, Kazuma Hashimoto, Wenpeng Yin, Akari Asai, Jia Li, Philip~S. Yu,
  and Caiming Xiong.
\newblock Adv-bert: {BERT} is not robust on misspellings! generating nature
  adversarial samples on {BERT}.
\newblock \emph{arXiv preprint}, 2020.
\newblock URL \url{https://arxiv.org/abs/2003.04985}.

\bibitem[Tay et~al.(2021)Tay, Tran, Ruder, Gupta, Chung, Bahri, Qin,
  Baumgartner, Yu, and Metzler]{tay2021charformer}
Yi~Tay, Vinh~Q Tran, Sebastian Ruder, Jai Gupta, Hyung~Won Chung, Dara Bahri,
  Zhen Qin, Simon Baumgartner, Cong Yu, and Donald Metzler.
\newblock Charformer: Fast character transformers via gradient-based subword
  tokenization.
\newblock In \emph{International Conference on Learning Representations}, 2021.
\newblock URL \url{https://openreview.net/forum?id=JtBRnrlOEFN}.

\bibitem[Taylor(2004)]{taylor-2004-pango}
Owen Taylor.
\newblock Pango, an open-source unicode text layout engine.
\newblock In \emph{Proceedings of the 25th Internationalization and Unicode
  Conference}, Washington, D.C., USA, 2004. The Unicode Consortium.
\newblock URL
  \url{https://people.redhat.com/otaylor/iuc25/pango-unicode-paper.pdf}.

\bibitem[Tjong Kim~Sang \& De~Meulder(2003)Tjong Kim~Sang and
  De~Meulder]{tjong-kim-sang-de-meulder-2003-introduction}
Erik~F. Tjong Kim~Sang and Fien De~Meulder.
\newblock Introduction to the {C}o{NLL}-2003 shared task: Language-independent
  named entity recognition.
\newblock In \emph{Proceedings of the Seventh Conference on Natural Language
  Learning at {HLT}-{NAACL} 2003}, pp.\  142--147, 2003.
\newblock URL \url{https://aclanthology.org/W03-0419}.

\bibitem[Touvron et~al.(2019)Touvron, Vedaldi, Douze, and
  Jegou]{touvron-etal-2019}
Hugo Touvron, Andrea Vedaldi, Matthijs Douze, and Herve Jegou.
\newblock Fixing the train-test resolution discrepancy.
\newblock In H.~Wallach, H.~Larochelle, A.~Beygelzimer, F.~d'Alch\'{e} Buc,
  E.~Fox, and R.~Garnett (eds.), \emph{Advances in Neural Information
  Processing Systems}, volume~32. Curran Associates, Inc., 2019.
\newblock URL
  \url{https://proceedings.neurips.cc/paper/2019/file/d03a857a23b5285736c4d55e0bb067c8-Paper.pdf}.

\bibitem[Turc et~al.(2021)Turc, Lee, Eisenstein, Chang, and
  Toutanova]{turc2021revisiting}
Iulia Turc, Kenton Lee, Jacob Eisenstein, Ming-Wei Chang, and Kristina
  Toutanova.
\newblock Revisiting the primacy of english in zero-shot cross-lingual
  transfer.
\newblock \emph{arXiv preprint}, 2021.
\newblock URL \url{https://arxiv.org/abs/2106.16171}.

\bibitem[van Esch et~al.(2019)van Esch, Sarbar, Lucassen, O'Brien, Breiner,
  Prasad, Crew, Nguyen, and Beaufays]{EschWritingAcrossWorld}
Daan van Esch, Elnaz Sarbar, Tamar Lucassen, Jeremy O'Brien, Theresa Breiner,
  Manasa Prasad, Evan Crew, Chieu Nguyen, and Fran{\c{c}}oise Beaufays.
\newblock Writing across the world's languages: Deep internationalization for
  gboard, the google keyboard.
\newblock \emph{Technical report}, 2019.
\newblock URL \url{http://arxiv.org/abs/1912.01218}.

\bibitem[Vaswani et~al.(2017)Vaswani, Shazeer, Parmar, Uszkoreit, Jones, Gomez,
  Kaiser, and Polosukhin]{DBLP:conf/nips/VaswaniSPUJGKP17}
Ashish Vaswani, Noam Shazeer, Niki Parmar, Jakob Uszkoreit, Llion Jones,
  Aidan~N. Gomez, Lukasz Kaiser, and Illia Polosukhin.
\newblock Attention is all you need.
\newblock In Isabelle Guyon, Ulrike von Luxburg, Samy Bengio, Hanna~M. Wallach,
  Rob Fergus, S.~V.~N. Vishwanathan, and Roman Garnett (eds.), \emph{Advances
  in Neural Information Processing Systems 30: Annual Conference on Neural
  Information Processing Systems 2017, December 4-9, 2017, Long Beach, CA,
  {USA}}, pp.\  5998--6008, 2017.
\newblock URL
  \url{https://proceedings.neurips.cc/paper/2017/hash/3f5ee243547dee91fbd053c1c4a845aa-Abstract.html}.

\bibitem[Vernikos \& Popescu-Belis(2021)Vernikos and
  Popescu-Belis]{vernikos-popescu-belis-2021-subword-mapping}
Giorgos Vernikos and Andrei Popescu-Belis.
\newblock Subword mapping and anchoring across languages.
\newblock In \emph{Findings of the Association for Computational Linguistics:
  EMNLP 2021}, pp.\  2633--2647, Punta Cana, Dominican Republic, November 2021.
  Association for Computational Linguistics.
\newblock \doi{10.18653/v1/2021.findings-emnlp.224}.
\newblock URL \url{https://aclanthology.org/2021.findings-emnlp.224}.

\bibitem[Wan(2022)]{wan2022fairness}
Ada Wan.
\newblock Fairness in representation for multilingual {NLP}: Insights from
  controlled experiments on conditional language modeling.
\newblock In \emph{International Conference on Learning Representations}, 2022.
\newblock URL \url{https://openreview.net/forum?id=-llS6TiOew}.

\bibitem[Wang et~al.(2018)Wang, Singh, Michael, Hill, Levy, and
  Bowman]{wang-etal-2018-glue}
Alex Wang, Amanpreet Singh, Julian Michael, Felix Hill, Omer Levy, and Samuel
  Bowman.
\newblock {GLUE}: A multi-task benchmark and analysis platform for natural
  language understanding.
\newblock In \emph{Proceedings of the 2018 {EMNLP} Workshop {B}lackbox{NLP}:
  Analyzing and Interpreting Neural Networks for {NLP}}, pp.\  353--355,
  Brussels, Belgium, November 2018. Association for Computational Linguistics.
\newblock \doi{10.18653/v1/W18-5446}.
\newblock URL \url{https://aclanthology.org/W18-5446}.

\bibitem[Wang et~al.(2020)Wang, Cho, and Gu]{DBLP:conf/aaai/WangCG20}
Changhan Wang, Kyunghyun Cho, and Jiatao Gu.
\newblock Neural machine translation with byte-level subwords.
\newblock In \emph{The Thirty-Fourth {AAAI} Conference on Artificial
  Intelligence, {AAAI} 2020, The Thirty-Second Innovative Applications of
  Artificial Intelligence Conference, {IAAI} 2020, The Tenth {AAAI} Symposium
  on Educational Advances in Artificial Intelligence, {EAAI} 2020, New York,
  NY, USA, February 7-12, 2020}, pp.\  9154--9160. {AAAI} Press, 2020.
\newblock URL \url{https://ojs.aaai.org/index.php/AAAI/article/view/6451}.

\bibitem[Wang et~al.(2022)Wang, Ruder, and Neubig]{wang-etal-2022-expanding}
Xinyi Wang, Sebastian Ruder, and Graham Neubig.
\newblock Expanding pretrained models to thousands more languages via
  lexicon-based adaptation.
\newblock In \emph{Proceedings of the 60th Annual Meeting of the Association
  for Computational Linguistics (Volume 1: Long Papers)}, pp.\  863--877,
  Dublin, Ireland, May 2022. Association for Computational Linguistics.
\newblock \doi{10.18653/v1/2022.acl-long.61}.
\newblock URL \url{https://aclanthology.org/2022.acl-long.61}.

\bibitem[Wei et~al.(2021)Wei, Liu, Guo, and
  Jiang]{DBLP:journals/corr/abs-2101-09469}
Junqiu Wei, Qun Liu, Yinpeng Guo, and Xin Jiang.
\newblock Training multilingual pre-trained language model with byte-level
  subwords.
\newblock \emph{arXiv preprint}, 2021.
\newblock URL \url{https://arxiv.org/abs/2101.09469}.

\bibitem[Wettig et~al.(2022)Wettig, Gao, Zhong, and
  Chen]{wettig-etal-2022-mask}
Alexander Wettig, Tianyu Gao, Zexuan Zhong, and Danqi Chen.
\newblock Should you mask 15\% in masked language modeling?
\newblock \emph{arXiv preprint}, 2022.
\newblock URL \url{https://arxiv.org/abs/2202.08005}.

\bibitem[Wolf et~al.(2020)Wolf, Debut, Sanh, Chaumond, Delangue, Moi, Cistac,
  Rault, Louf, Funtowicz, Davison, Shleifer, von Platen, Ma, Jernite, Plu, Xu,
  Le~Scao, Gugger, Drame, Lhoest, and Rush]{wolf-etal-2020-transformers}
Thomas Wolf, Lysandre Debut, Victor Sanh, Julien Chaumond, Clement Delangue,
  Anthony Moi, Pierric Cistac, Tim Rault, Remi Louf, Morgan Funtowicz, Joe
  Davison, Sam Shleifer, Patrick von Platen, Clara Ma, Yacine Jernite, Julien
  Plu, Canwen Xu, Teven Le~Scao, Sylvain Gugger, Mariama Drame, Quentin Lhoest,
  and Alexander Rush.
\newblock Transformers: State-of-the-art natural language processing.
\newblock In \emph{Proceedings of the 2020 Conference on Empirical Methods in
  Natural Language Processing: System Demonstrations}, pp.\  38--45, Online,
  October 2020. Association for Computational Linguistics.
\newblock \doi{10.18653/v1/2020.emnlp-demos.6}.
\newblock URL \url{https://aclanthology.org/2020.emnlp-demos.6}.

\bibitem[Wu \& Dredze(2020)Wu and Dredze]{wu-dredze-2020-languages}
Shijie Wu and Mark Dredze.
\newblock Are all languages created equal in multilingual {BERT}?
\newblock In \emph{Proceedings of the 5th Workshop on Representation Learning
  for NLP}, pp.\  120--130, Online, July 2020. Association for Computational
  Linguistics.
\newblock \doi{10.18653/v1/2020.repl4nlp-1.16}.
\newblock URL \url{https://aclanthology.org/2020.repl4nlp-1.16}.

\bibitem[Wu et~al.(2019)Wu, Meng, Wang, Han, Li, Li, Mei, Nie, Sun, and
  Li]{Wu2019GlyceGF}
Wei Wu, Yuxian Meng, Fei Wang, Qinghong Han, Muyu Li, Xiaoya Li, Jie Mei, Ping
  Nie, Xiaofei Sun, and Jiwei Li.
\newblock Glyce: Glyph-vectors for chinese character representations.
\newblock In \emph{Neural Information Processing Systems}, 2019.

\bibitem[Xue et~al.(2022)Xue, Barua, Constant, Al-Rfou, Narang, Kale, Roberts,
  and Raffel]{xue-etal-2022-byt5}
Linting Xue, Aditya Barua, Noah Constant, Rami Al-Rfou, Sharan Narang, Mihir
  Kale, Adam Roberts, and Colin Raffel.
\newblock {ByT5: Towards a Token-Free Future with Pre-trained Byte-to-Byte
  Models}.
\newblock \emph{Transactions of the Association for Computational Linguistics},
  10:\penalty0 291--306, 03 2022.
\newblock ISSN 2307-387X.
\newblock \doi{10.1162/tacl_a_00461}.
\newblock URL \url{https://doi.org/10.1162/tacl\_a\_00461}.

\bibitem[Zeldes \& Abrams(2018)Zeldes and Abrams]{zeldes-abrams-2018-coptic}
Amir Zeldes and Mitchell Abrams.
\newblock The {C}optic {U}niversal {D}ependency treebank.
\newblock In \emph{Proceedings of the Second Workshop on Universal Dependencies
  ({UDW} 2018)}, pp.\  192--201, Brussels, Belgium, November 2018. Association
  for Computational Linguistics.
\newblock \doi{10.18653/v1/W18-6022}.
\newblock URL \url{https://aclanthology.org/W18-6022}.

\bibitem[Zeman et~al.(2022)Zeman, Nivre, Abrams, Ackermann, Aepli, Aghaei,
  Agi{\'c}, Ahmadi, Ahrenberg, Ajede, Aleksandravi{\v c}i{\=u}t{\.e}, Alfina,
  Algom, Andersen, Antonsen, Aplonova, Aquino, Aragon, Aranes, Aranzabe,
  Ar{\i}can, Arnard{\'o}ttir, Arutie, Arwidarasti, Asahara, Aslan, Asmazo{\u
  g}lu, Ateyah, Atmaca, Attia, Atutxa, Augustinus, Badmaeva, Balasubramani,
  Ballesteros, Banerjee, Bank, Barbu~Mititelu, Barkarson, Basile, Basmov,
  Batchelor, Bauer, Bedir, Bengoetxea, Ben~Moshe, Berk, Berzak, Bhat, Bhat,
  Biagetti, Bick, Bielinskien{\.e}, Bjarnad{\'o}ttir, Blokland, Bobicev,
  Boizou, Borges~V{\"o}lker, B{\"o}rstell, Bosco, Bouma, Bowman, Boyd,
  Braggaar, Brokait{\.e}, Burchardt, Candito, Caron, Caron, Cassidy,
  Cavalcanti, Cebiro{\u g}lu~Eryi{\u g}it, Cecchini, Celano, {\v
  C}{\'e}pl{\"o}, Cesur, Cetin, {\c C}etino{\u g}lu, Chalub, Chauhan, Chi,
  Chika, Cho, Choi, Chun, Chung, Cignarella, Cinkov{\'a}, Collomb, {\c
  C}{\"o}ltekin, Connor, Corbetta, Courtin, Cristescu, Daniel, Davidson,
  Dehouck, de~Laurentiis, de~Marneffe, de~Paiva, Derin, de~Souza, Diaz~de
  Ilarraza, Dickerson, Dinakaramani, Di~Nuovo, Dione, Dirix, Dobrovoljc, Dozat,
  Droganova, Dwivedi, Eckhoff, Eiche, Eli, Elkahky, Ephrem, Erina, Erjavec,
  Etienne, Evelyn, Facundes, Farkas, Favero, Ferdaousi, Fernanda,
  Fernandez~Alcalde, Foster, Freitas, Fujita, Gajdo{\v s}ov{\'a}, Galbraith,
  Gamba, Garcia, G{\"a}rdenfors, Garza, Gerardi, Gerdes, Ginter, Godoy,
  Goenaga, Gojenola, G{\"o}k{\i}rmak, Goldberg, G{\'o}mez~Guinovart,
  Gonz{\'a}lez~Saavedra, Grici{\=u}t{\.e}, Grioni, Grobol, Gr{\=
  u}z{\={\i}}tis, Guillaume, Guillot-Barbance, G{\"u}ng{\"o}r, Habash,
  Hafsteinsson, Haji{\v c}, Haji{\v c}~jr., H{\"a}m{\"a}l{\"a}inen,
  H{\`a}~M{\~y}, Han, Hanifmuti, Harada, Hardwick, Harris, Haug, Heinecke,
  Hellwig, Hennig, Hladk{\'a}, Hlav{\'a}{\v c}ov{\'a}, Hociung, Hohle, Hwang,
  Ikeda, Ingason, Ion, Irimia, Ishola, Ito, Jannat, Jel{\'{\i}}nek, Jha,
  Johannsen, J{\'o}nsd{\'o}ttir, J{\o}rgensen, Juutinen, K, Ka{\c s}{\i}kara,
  Kaasen, Kabaeva, Kahane, Kanayama, Kanerva, Kara, Karah{\'o}ǧa, Katz,
  Kayadelen, Kenney, Kettnerov{\'a}, Kirchner, Klementieva, Klyachko, K{\"o}hn,
  K{\"o}ksal, Kopacewicz, Korkiakangas, K{\"o}se, Kotsyba, Kovalevskait{\.e},
  Krek, Krishnamurthy, K{\"u}bler, Kuyruk{\c c}u, Kuzgun, Kwak, Laippala, Lam,
  Lambertino, Lando, Larasati, Lavrentiev, Lee, \fontencoding{T5}\selectfont
  {Phương L{\^e}~H{\`{\^o}}ng}, Lenci, Lertpradit, Leung, Levina, Li, Li, Li,
  Li, Lim, Lima~Padovani, Lind{\'e}n, Ljube{\v s}i{\'c}, Loginova, Lusito,
  Luthfi, Luukko, Lyashevskaya, Lynn, Macketanz, Mahamdi, Maillard, Makazhanov,
  Mandl, Manning, Manurung, Mar{\c s}an, M{\u a}r{\u a}nduc, Mare{\v c}ek,
  Marheinecke, Markantonatou, Mart{\'{\i}}nez~Alonso,
  Mart{\'{\i}}n~Rodr{\'{\i}}guez, Martins, Ma{\v s}ek, Matsuda, Matsumoto,
  Mazzei, {McDonald}, {McGuinness}, Mendon{\c c}a, Merzhevich, Miekka,
  Mischenkova, Misirpashayeva, Missil{\"a}, Mititelu, Mitrofan, Miyao,
  Mojiri~Foroushani, Moln{\'a}r, Moloodi, Montemagni, More, Moreno~Romero,
  Moretti, Mori, Mori, Morioka, Moro, Mortensen, Moskalevskyi, Muischnek,
  Munro, Murawaki, M{\"u}{\"u}risep, Nainwani, Nakhl{\'e},
  Navarro~Hor{\~n}iacek, Nedoluzhko, Ne{\v s}pore-B{\=e}rzkalne, Nevaci,
  \fontencoding{T5}\selectfont{Lương Nguy{\~{\^e}}n Th{\d i}}, Nguy{\~{\^e}}n
  Th{\d i}~Minh, Nikaido, Nikolaev, Nitisaroj, Nourian, Nurmi, Ojala, Ojha,
  Ol{\'u}{\`o}kun, Omura, Onwuegbuzia, Ordan, Osenova, {\"O}stling, {\O}vrelid,
  {\"O}zate{\c s}, {\"O}z{\c c}elik, {\"O}zg{\"u}r, {\"O}zt{\"u}rk~Ba{\c
  s}aran, Paccosi, Palmero~Aprosio, Park, Partanen, Pascual, Passarotti,
  Patejuk, Paulino-Passos, Pedonese, Peljak-{\L}api{\'n}ska, Peng, Perez,
  Perkova, Perrier, Petrov, Petrova, Peverelli, Phelan, Piitulainen, Pirinen,
  Pitler, Plank, Poibeau, Ponomareva, Popel, Pretkalni{\c n}a, Pr{\'e}vost,
  Prokopidis, Przepi{\'o}rkowski, Puolakainen, Pyysalo, Qi, R{\"a}{\"a}bis,
  Rademaker, Rahoman, Rama, Ramasamy, Ramisch, Rashel, Rasooli, Ravishankar,
  Real, Rebeja, Reddy, Regnault, Rehm, Riabov, Rie{\ss}ler, Rimkut{\.e},
  Rinaldi, Rituma, Rizqiyah, Rocha, R{\"o}gnvaldsson, Romanenko, Rosa, Roșca,
  Rovati, Rozonoyer, Rudina, Rueter, R{\'u}narsson, Sadde, Safari, Sagot,
  Sahala, Saleh, Salomoni, Samard{\v z}i{\'c}, Samson, Sanguinetti, San{\i}yar,
  S{\"a}rg, Saul{\={\i}}te, Sawanakunanon, Saxena, Scannell, Scarlata,
  Schneider, Schuster, Schwartz, Seddah, Seeker, Seraji, Shahzadi, Shen,
  Shimada, Shirasu, Shishkina, Shohibussirri, Sichinava, Siewert,
  \fontencoding{T1}\selectfont{Einar Freyr Sigurðsson}, Silveira, Silveira,
  Simi, Simionescu, Simk{\'o}, {\v S}imkov{\'a}, Simov, Skachedubova, Smith,
  Soares-Bastos, Sourov, Spadine, Sprugnoli, Stamou, Steingr{\'{\i}}msson,
  Stella, Straka, Strickland, Strnadov{\'a}, Suhr, Sulestio, Sulubacak, Suzuki,
  Swanson, Sz{\'a}nt{\'o}, Taguchi, Taji, Takahashi, Tamburini, Tan, Tanaka,
  Tanaya, Tavoni, Tella, Tellier, Testori, Thomas, Tonelli, Torga, Toska,
  Trosterud, Trukhina, Tsarfaty, T{\"u}rk, Tyers, Uematsu, Untilov, Ure{\v
  s}ov{\'a}, Uria, Uszkoreit, Utka, Vagnoni, Vajjala, van~der Goot, Vanhove,
  van Niekerk, van Noord, Varga, Vedenina, Villemonte de~la Clergerie, Vincze,
  Vlasova, Wakasa, Wallenberg, Wallin, Walsh, Wang, Washington, Wendt, Widmer,
  Wigderson, Wijono, Williams, Wir{\'e}n, Wittern, Woldemariam, Wong,
  Wr{\'o}blewska, Yako, Yamashita, Yamazaki, Yan, Yasuoka, Yavrumyan, Yenice,
  Y{\i}ld{\i}z, Yu, Yuliawati, {\v Z}abokrtsk{\'y}, Zahra, Zeldes, Zhou, Zhu,
  Zhuravleva, and Ziane]{zeman-etal-2022-ud}
Daniel Zeman, Joakim Nivre, Mitchell Abrams, Elia Ackermann, No{\"e}mi Aepli,
  Hamid Aghaei, {\v Z}eljko Agi{\'c}, Amir Ahmadi, Lars Ahrenberg,
  Chika~Kennedy Ajede, Gabriel{\.e} Aleksandravi{\v c}i{\=u}t{\.e}, Ika Alfina,
  Avner Algom, Erik Andersen, Lene Antonsen, Katya Aplonova, Angelina Aquino,
  Carolina Aragon, Glyd Aranes, Maria~Jesus Aranzabe, Bilge~Nas Ar{\i}can, {\t
  H}{\'o}runn Arnard{\'o}ttir, Gashaw Arutie, Jessica~Naraiswari Arwidarasti,
  Masayuki Asahara, Deniz~Baran Aslan, Cengiz Asmazo{\u g}lu, Luma Ateyah,
  Furkan Atmaca, Mohammed Attia, Aitziber Atutxa, Liesbeth Augustinus, Elena
  Badmaeva, Keerthana Balasubramani, Miguel Ballesteros, Esha Banerjee,
  Sebastian Bank, Verginica Barbu~Mititelu, Starkaður Barkarson, Rodolfo
  Basile, Victoria Basmov, Colin Batchelor, John Bauer, Seyyit~Talha Bedir,
  Kepa Bengoetxea, Yifat Ben~Moshe, G{\"o}zde Berk, Yevgeni Berzak,
  Irshad~Ahmad Bhat, Riyaz~Ahmad Bhat, Erica Biagetti, Eckhard Bick, Agn{\.e}
  Bielinskien{\.e}, Krist{\'{\i}}n Bjarnad{\'o}ttir, Rogier Blokland, Victoria
  Bobicev, Lo{\"{\i}}c Boizou, Emanuel Borges~V{\"o}lker, Carl B{\"o}rstell,
  Cristina Bosco, Gosse Bouma, Sam Bowman, Adriane Boyd, Anouck Braggaar,
  Kristina Brokait{\.e}, Aljoscha Burchardt, Marie Candito, Bernard Caron,
  Gauthier Caron, Lauren Cassidy, Tatiana Cavalcanti, G{\"u}l{\c s}en Cebiro{\u
  g}lu~Eryi{\u g}it, Flavio~Massimiliano Cecchini, Giuseppe G.~A. Celano,
  Slavom{\'{\i}}r {\v C}{\'e}pl{\"o}, Neslihan Cesur, Savas Cetin, {\"O}zlem
  {\c C}etino{\u g}lu, Fabricio Chalub, Shweta Chauhan, Ethan Chi, Taishi
  Chika, Yongseok Cho, Jinho Choi, Jayeol Chun, Juyeon Chung, Alessandra~T.
  Cignarella, Silvie Cinkov{\'a}, Aur{\'e}lie Collomb, {\c C}a{\u g}r{\i} {\c
  C}{\"o}ltekin, Miriam Connor, Daniela Corbetta, Marine Courtin, Mihaela
  Cristescu, Philemon Daniel, Elizabeth Davidson, Mathieu Dehouck, Martina
  de~Laurentiis, Marie-Catherine de~Marneffe, Valeria de~Paiva, Mehmet~Oguz
  Derin, Elvis de~Souza, Arantza Diaz~de Ilarraza, Carly Dickerson, Arawinda
  Dinakaramani, Elisa Di~Nuovo, Bamba Dione, Peter Dirix, Kaja Dobrovoljc,
  Timothy Dozat, Kira Droganova, Puneet Dwivedi, Hanne Eckhoff, Sandra Eiche,
  Marhaba Eli, Ali Elkahky, Binyam Ephrem, Olga Erina, Toma{\v z} Erjavec,
  Aline Etienne, Wograine Evelyn, Sidney Facundes, Rich{\'a}rd Farkas, Federica
  Favero, Jannatul Ferdaousi, Mar{\'{\i}}lia Fernanda, Hector
  Fernandez~Alcalde, Jennifer Foster, Cl{\'a}udia Freitas, Kazunori Fujita,
  Katar{\'{\i}}na Gajdo{\v s}ov{\'a}, Daniel Galbraith, Federica Gamba, Marcos
  Garcia, Moa G{\"a}rdenfors, Sebastian Garza, Fabr{\'{\i}}cio~Ferraz Gerardi,
  Kim Gerdes, Filip Ginter, Gustavo Godoy, Iakes Goenaga, Koldo Gojenola,
  Memduh G{\"o}k{\i}rmak, Yoav Goldberg, Xavier G{\'o}mez~Guinovart, Berta
  Gonz{\'a}lez~Saavedra, Bernadeta Grici{\=u}t{\.e}, Matias Grioni, Lo{\"{\i}}c
  Grobol, Normunds Gr{\= u}z{\={\i}}tis, Bruno Guillaume, C{\'e}line
  Guillot-Barbance, Tunga G{\"u}ng{\"o}r, Nizar Habash, Hinrik Hafsteinsson,
  Jan Haji{\v c}, Jan Haji{\v c}~jr., Mika H{\"a}m{\"a}l{\"a}inen, Linh
  H{\`a}~M{\~y}, Na-Rae Han, Muhammad~Yudistira Hanifmuti, Takahiro Harada, Sam
  Hardwick, Kim Harris, Dag Haug, Johannes Heinecke, Oliver Hellwig, Felix
  Hennig, Barbora Hladk{\'a}, Jaroslava Hlav{\'a}{\v c}ov{\'a}, Florinel
  Hociung, Petter Hohle, Jena Hwang, Takumi Ikeda, Anton~Karl Ingason, Radu
  Ion, Elena Irimia, {\d O}l{\'a}j{\'{\i}}d{\'e} Ishola, Kaoru Ito, Siratun
  Jannat, Tom{\'a}{\v s} Jel{\'{\i}}nek, Apoorva Jha, Anders Johannsen, Hildur
  J{\'o}nsd{\'o}ttir, Fredrik J{\o}rgensen, Markus Juutinen, Sarveswaran K,
  H{\"u}ner Ka{\c s}{\i}kara, Andre Kaasen, Nadezhda Kabaeva, Sylvain Kahane,
  Hiroshi Kanayama, Jenna Kanerva, Neslihan Kara, Ritv{\'a}n Karah{\'o}ǧa,
  Boris Katz, Tolga Kayadelen, Jessica Kenney, V{\'a}clava Kettnerov{\'a},
  Jesse Kirchner, Elena Klementieva, Elena Klyachko, Arne K{\"o}hn, Abdullatif
  K{\"o}ksal, Kamil Kopacewicz, Timo Korkiakangas, Mehmet K{\"o}se, Natalia
  Kotsyba, Jolanta Kovalevskait{\.e}, Simon Krek, Parameswari Krishnamurthy,
  Sandra K{\"u}bler, O{\u g}uzhan Kuyruk{\c c}u, Asl{\i} Kuzgun, Sookyoung
  Kwak, Veronika Laippala, Lucia Lam, Lorenzo Lambertino, Tatiana Lando,
  Septina~Dian Larasati, Alexei Lavrentiev, John Lee,
  \fontencoding{T5}\selectfont {Phương L{\^e}~H{\`{\^o}}ng}, Alessandro
  Lenci, Saran Lertpradit, Herman Leung, Maria Levina, Cheuk~Ying Li, Josie Li,
  Keying Li, Yuan Li, {KyungTae} Lim, Bruna Lima~Padovani, Krister Lind{\'e}n,
  Nikola Ljube{\v s}i{\'c}, Olga Loginova, Stefano Lusito, Andry Luthfi, Mikko
  Luukko, Olga Lyashevskaya, Teresa Lynn, Vivien Macketanz, Menel Mahamdi, Jean
  Maillard, Aibek Makazhanov, Michael Mandl, Christopher Manning, Ruli
  Manurung, B{\"u}{\c s}ra Mar{\c s}an, C{\u a}t{\u a}lina M{\u a}r{\u a}nduc,
  David Mare{\v c}ek, Katrin Marheinecke, Stella Markantonatou, H{\'e}ctor
  Mart{\'{\i}}nez~Alonso, Lorena Mart{\'{\i}}n~Rodr{\'{\i}}guez, Andr{\'e}
  Martins, Jan Ma{\v s}ek, Hiroshi Matsuda, Yuji Matsumoto, Alessandro Mazzei,
  Ryan {McDonald}, Sarah {McGuinness}, Gustavo Mendon{\c c}a, Tatiana
  Merzhevich, Niko Miekka, Karina Mischenkova, Margarita Misirpashayeva, Anna
  Missil{\"a}, C{\u a}t{\u a}lin Mititelu, Maria Mitrofan, Yusuke Miyao,
  {AmirHossein} Mojiri~Foroushani, Judit Moln{\'a}r, Amirsaeid Moloodi,
  Simonetta Montemagni, Amir More, Laura Moreno~Romero, Giovanni Moretti,
  Keiko~Sophie Mori, Shinsuke Mori, Tomohiko Morioka, Shigeki Moro, Bjartur
  Mortensen, Bohdan Moskalevskyi, Kadri Muischnek, Robert Munro, Yugo Murawaki,
  Kaili M{\"u}{\"u}risep, Pinkey Nainwani, Mariam Nakhl{\'e}, Juan~Ignacio
  Navarro~Hor{\~n}iacek, Anna Nedoluzhko, Gunta Ne{\v s}pore-B{\=e}rzkalne,
  Manuela Nevaci, \fontencoding{T5}\selectfont{Lương Nguy{\~{\^e}}n Th{\d
  i}}, Huy{\`{\^e}}n Nguy{\~{\^e}}n Th{\d i}~Minh, Yoshihiro Nikaido, Vitaly
  Nikolaev, Rattima Nitisaroj, Alireza Nourian, Hanna Nurmi, Stina Ojala,
  Atul~Kr. Ojha, Ad{\'e}day{\`o} Ol{\'u}{\`o}kun, Mai Omura, Emeka Onwuegbuzia,
  Noam Ordan, Petya Osenova, Robert {\"O}stling, Lilja {\O}vrelid, {\c
  S}aziye~Bet{\"u}l {\"O}zate{\c s}, Merve {\"O}z{\c c}elik, Arzucan
  {\"O}zg{\"u}r, Balk{\i}z {\"O}zt{\"u}rk~Ba{\c s}aran, Teresa Paccosi, Alessio
  Palmero~Aprosio, Hyunji~Hayley Park, Niko Partanen, Elena Pascual, Marco
  Passarotti, Agnieszka Patejuk, Guilherme Paulino-Passos, Giulia Pedonese,
  Angelika Peljak-{\L}api{\'n}ska, Siyao Peng, Cenel-Augusto Perez, Natalia
  Perkova, Guy Perrier, Slav Petrov, Daria Petrova, Andrea Peverelli, Jason
  Phelan, Jussi Piitulainen, Tommi~A Pirinen, Emily Pitler, Barbara Plank,
  Thierry Poibeau, Larisa Ponomareva, Martin Popel, Lauma Pretkalni{\c n}a,
  Sophie Pr{\'e}vost, Prokopis Prokopidis, Adam Przepi{\'o}rkowski, Tiina
  Puolakainen, Sampo Pyysalo, Peng Qi, Andriela R{\"a}{\"a}bis, Alexandre
  Rademaker, Mizanur Rahoman, Taraka Rama, Loganathan Ramasamy, Carlos Ramisch,
  Fam Rashel, Mohammad~Sadegh Rasooli, Vinit Ravishankar, Livy Real, Petru
  Rebeja, Siva Reddy, Mathilde Regnault, Georg Rehm, Ivan Riabov, Michael
  Rie{\ss}ler, Erika Rimkut{\.e}, Larissa Rinaldi, Laura Rituma, Putri
  Rizqiyah, Luisa Rocha, Eir{\'{\i}}kur R{\"o}gnvaldsson, Mykhailo Romanenko,
  Rudolf Rosa, Valentin Roșca, Davide Rovati, Ben Rozonoyer, Olga Rudina, Jack
  Rueter, Kristj{\'a}n R{\'u}narsson, Shoval Sadde, Pegah Safari, Beno{\^{\i}}t
  Sagot, Aleksi Sahala, Shadi Saleh, Alessio Salomoni, Tanja Samard{\v
  z}i{\'c}, Stephanie Samson, Manuela Sanguinetti, Ezgi San{\i}yar, Dage
  S{\"a}rg, Baiba Saul{\={\i}}te, Yanin Sawanakunanon, Shefali Saxena, Kevin
  Scannell, Salvatore Scarlata, Nathan Schneider, Sebastian Schuster, Lane
  Schwartz, Djam{\'e} Seddah, Wolfgang Seeker, Mojgan Seraji, Syeda Shahzadi,
  Mo~Shen, Atsuko Shimada, Hiroyuki Shirasu, Yana Shishkina, Muh Shohibussirri,
  Dmitry Sichinava, Janine Siewert, \fontencoding{T1}\selectfont{Einar Freyr
  Sigurðsson}, Aline Silveira, Natalia Silveira, Maria Simi, Radu Simionescu,
  Katalin Simk{\'o}, M{\'a}ria {\v S}imkov{\'a}, Kiril Simov, Maria
  Skachedubova, Aaron Smith, Isabela Soares-Bastos, Shafi Sourov, Carolyn
  Spadine, Rachele Sprugnoli, Vivian Stamou, Stein{\t h}{\'o}r
  Steingr{\'{\i}}msson, Antonio Stella, Milan Straka, Emmett Strickland, Jana
  Strnadov{\'a}, Alane Suhr, Yogi~Lesmana Sulestio, Umut Sulubacak, Shingo
  Suzuki, Daniel Swanson, Zsolt Sz{\'a}nt{\'o}, Chihiro Taguchi, Dima Taji,
  Yuta Takahashi, Fabio Tamburini, Mary Ann~C. Tan, Takaaki Tanaka, Dipta
  Tanaya, Mirko Tavoni, Samson Tella, Isabelle Tellier, Marinella Testori,
  Guillaume Thomas, Sara Tonelli, Liisi Torga, Marsida Toska, Trond Trosterud,
  Anna Trukhina, Reut Tsarfaty, Utku T{\"u}rk, Francis Tyers, Sumire Uematsu,
  Roman Untilov, Zde{\v n}ka Ure{\v s}ov{\'a}, Larraitz Uria, Hans Uszkoreit,
  Andrius Utka, Elena Vagnoni, Sowmya Vajjala, Rob van~der Goot, Martine
  Vanhove, Daniel van Niekerk, Gertjan van Noord, Viktor Varga, Uliana
  Vedenina, Eric Villemonte de~la Clergerie, Veronika Vincze, Natalia Vlasova,
  Aya Wakasa, Joel~C. Wallenberg, Lars Wallin, Abigail Walsh, Jing~Xian Wang,
  Jonathan~North Washington, Maximilan Wendt, Paul Widmer, Shira Wigderson,
  Sri~Hartati Wijono, Seyi Williams, Mats Wir{\'e}n, Christian Wittern, Tsegay
  Woldemariam, Tak-sum Wong, Alina Wr{\'o}blewska, Mary Yako, Kayo Yamashita,
  Naoki Yamazaki, Chunxiao Yan, Koichi Yasuoka, Marat~M. Yavrumyan,
  Arife~Bet{\"u}l Yenice, Olcay~Taner Y{\i}ld{\i}z, Zhuoran Yu, Arlisa
  Yuliawati, Zden{\v e}k {\v Z}abokrtsk{\'y}, Shorouq Zahra, Amir Zeldes,
  He~Zhou, Hanzhi Zhu, Anna Zhuravleva, and Rayan Ziane.
\newblock Universal dependencies 2.10, 2022.
\newblock URL \url{http://hdl.handle.net/11234/1-4758}.
\newblock {LINDAT}/{CLARIAH}-{CZ} digital library at the Institute of Formal
  and Applied Linguistics ({{\'U}FAL}), Faculty of Mathematics and Physics,
  Charles University.

\bibitem[Zhang et~al.(2022)Zhang, Chaudhary, Goyal, Cross, Wenzek, Bansal, and
  Guzm{\'a}n]{zhang2022robust}
Shiyue Zhang, Vishrav Chaudhary, Naman Goyal, James Cross, Guillaume Wenzek,
  Mohit Bansal, and Francisco Guzm{\'a}n.
\newblock How robust is neural machine translation to language imbalance in
  multilingual tokenizer training?
\newblock \emph{arXiv preprint}, 2022.
\newblock URL \url{https://arxiv.org/abs/2204.14268}.

\bibitem[Zhu et~al.(2015)Zhu, Kiros, Zemel, Salakhutdinov, Urtasun, Torralba,
  and Fidler]{Zhu_2015_ICCV}
Yukun Zhu, Ryan Kiros, Rich Zemel, Ruslan Salakhutdinov, Raquel Urtasun,
  Antonio Torralba, and Sanja Fidler.
\newblock Aligning books and movies: Towards story-like visual explanations by
  watching movies and reading books.
\newblock In \emph{The IEEE International Conference on Computer Vision
  (ICCV)}, December 2015.
\newblock URL
  \url{https://www.cv-foundation.org/openaccess/content_iccv_2015/papers/Zhu_Aligning_Books_and_ICCV_2015_paper.pdf}.

\end{thebibliography}
\bibliographystyle{iclr2022_conference}

\clearpage
\appendix
\section{Abstract Reconstructions} 
\label{app:abstract_reconstructions}
\begin{figure*}[ht!]
    \begin{subfigure}[]{0.32\textwidth}
        \centering
        {%
        \setlength{\fboxsep}{0pt}%
        \setlength{\fboxrule}{1pt}%
        \fbox{\includegraphics[width=\textwidth]{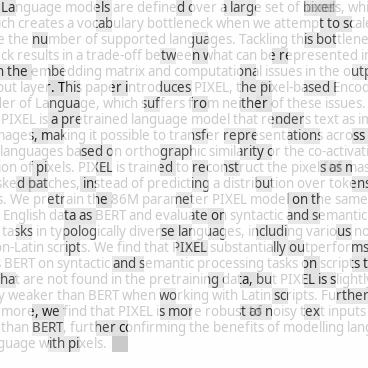}}%
        }%
    \end{subfigure}\hfill
    \vspace{0.25em}
    \begin{subfigure}[]{0.32\textwidth}
        \centering
        {%
        \setlength{\fboxsep}{0pt}%
        \setlength{\fboxrule}{1pt}%
        \fbox{\includegraphics[width=\textwidth]{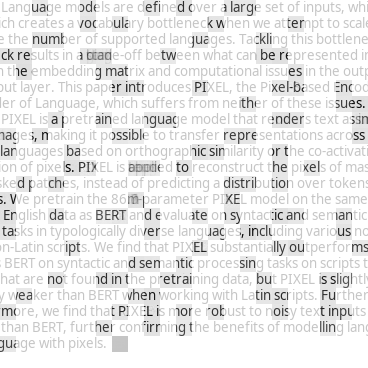}}%
        }%
    \end{subfigure}\hfill
    \vspace{0.25em}
    \begin{subfigure}[]{0.32\textwidth}
        \centering
        {%
        \setlength{\fboxsep}{0pt}%
        \setlength{\fboxrule}{1pt}%
        \fbox{\includegraphics[width=\textwidth]{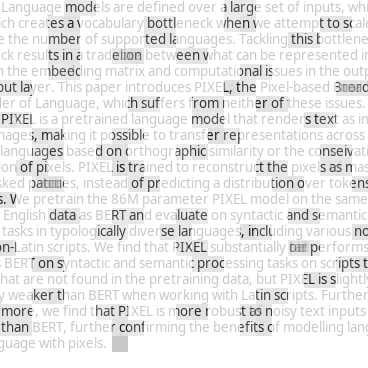}}%
        }%
    \end{subfigure}
    \vspace{0.5em}
        \begin{subfigure}[]{0.32\textwidth}
        \centering
        {%
        \setlength{\fboxsep}{0pt}%
        \setlength{\fboxrule}{1pt}%
        \fbox{\includegraphics[width=\textwidth]{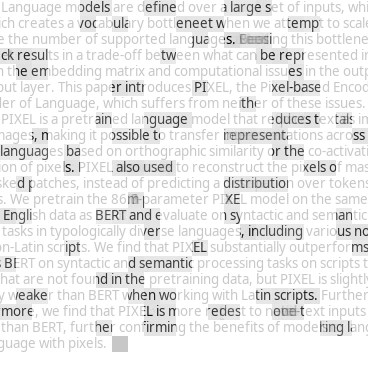}}%
        }%
    \end{subfigure}\hfill
    \begin{subfigure}[]{0.32\textwidth}
        \centering
        {%
        \setlength{\fboxsep}{0pt}%
        \setlength{\fboxrule}{1pt}%
        \fbox{\includegraphics[width=\textwidth]{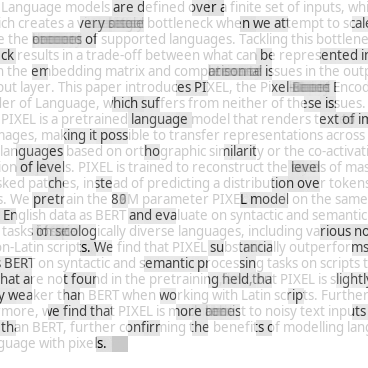}}%
        }%
    \end{subfigure}\hfill
    \begin{subfigure}[]{0.32\textwidth}
        \centering
        {%
        \setlength{\fboxsep}{0pt}%
        \setlength{\fboxrule}{1pt}%
        \fbox{\includegraphics[width=\textwidth]{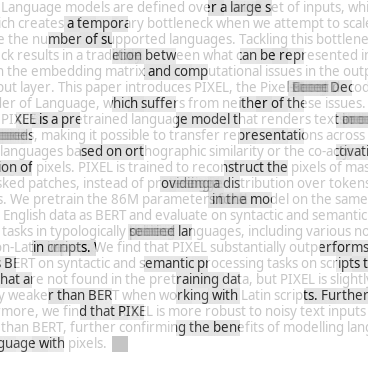}}%
        }%
    \end{subfigure}
    \vspace{0.5em}
    \begin{subfigure}[]{0.32\textwidth}
        \centering
        {%
        \setlength{\fboxsep}{0pt}%
        \setlength{\fboxrule}{1pt}%
        \fbox{\includegraphics[width=\textwidth]{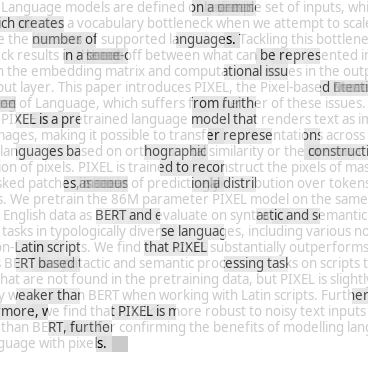}}%
        }%
    \end{subfigure}\hfill
    \begin{subfigure}[]{0.32\textwidth}
        \centering
        {%
        \setlength{\fboxsep}{0pt}%
        \setlength{\fboxrule}{1pt}%
        \fbox{\includegraphics[width=\textwidth]{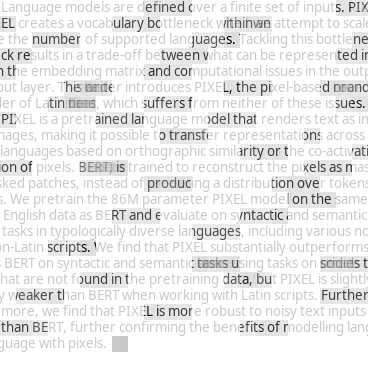}}%
        }%
    \end{subfigure}\hfill
    \begin{subfigure}[]{0.32\textwidth}
        \centering
        {%
        \setlength{\fboxsep}{0pt}%
        \setlength{\fboxrule}{1pt}%
        \fbox{\includegraphics[width=\textwidth]{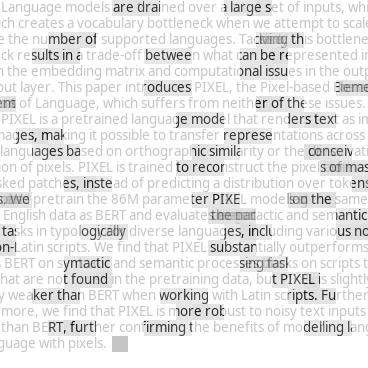}}%
        }%
    \end{subfigure}
    \vspace{0.5em}
        \begin{subfigure}[]{0.32\textwidth}
        \centering
        {%
        \setlength{\fboxsep}{0pt}%
        \setlength{\fboxrule}{1pt}%
        \fbox{\includegraphics[width=\textwidth]{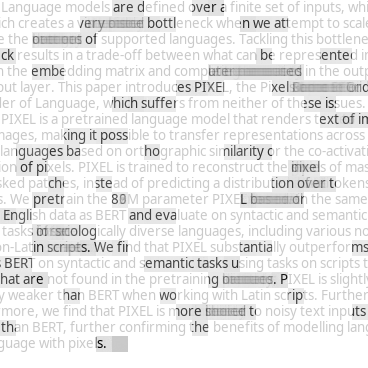}}%
        }%
    \end{subfigure}\hfill
    \begin{subfigure}[]{0.32\textwidth}
        \centering
        {%
        \setlength{\fboxsep}{0pt}%
        \setlength{\fboxrule}{1pt}%
        \fbox{\includegraphics[width=\textwidth]{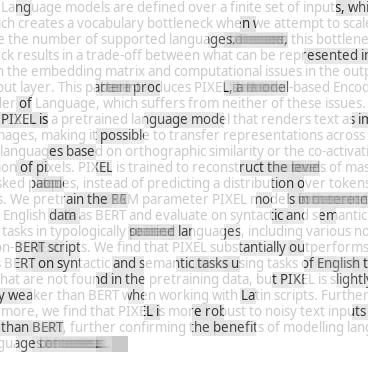}}%
        }%
    \end{subfigure}\hfill
    \begin{subfigure}[]{0.32\textwidth}
        \centering
        {%
        \setlength{\fboxsep}{0pt}%
        \setlength{\fboxrule}{1pt}%
        \fbox{\includegraphics[width=\textwidth]{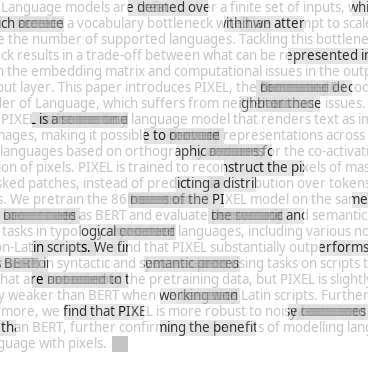}}%
        }%
    \end{subfigure}
    \caption{\model image reconstructions of the abstract with different span masks.}
    \label{fig:abstract_reconstructions}
\end{figure*}

\clearpage
\section{Web Text Reconstructions}
\label{app:steps_reconstructions}
\begin{figure*}[ht!]
    \begin{subfigure}[]{0.32\textwidth}
        \centering
        {%
        \setlength{\fboxsep}{0pt}%
        \setlength{\fboxrule}{1pt}%
        \fbox{\includegraphics[width=\textwidth]{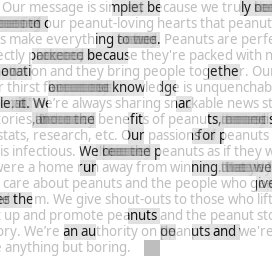}}%
        }%
    \end{subfigure}\hfill
    \vspace{0.25em}
    \begin{subfigure}[]{0.32\textwidth}
        \centering
        {%
        \setlength{\fboxsep}{0pt}%
        \setlength{\fboxrule}{1pt}%
        \fbox{\includegraphics[width=\textwidth]{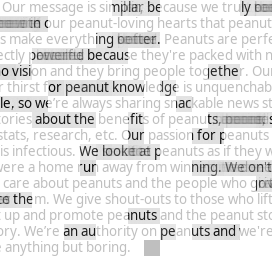}}%
        }%
    \end{subfigure}\hfill
    \vspace{0.25em}
    \begin{subfigure}[]{0.32\textwidth}
        \centering
        {%
        \setlength{\fboxsep}{0pt}%
        \setlength{\fboxrule}{1pt}%
        \fbox{\includegraphics[width=\textwidth]{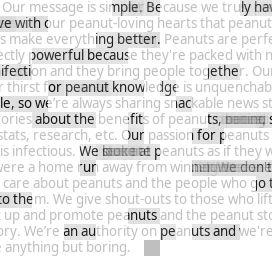}}%
        }%
    \end{subfigure}
    \vspace{0.5em}
        \begin{subfigure}[]{0.32\textwidth}
        \centering
        {%
        \setlength{\fboxsep}{0pt}%
        \setlength{\fboxrule}{1pt}%
        \fbox{\includegraphics[width=\textwidth]{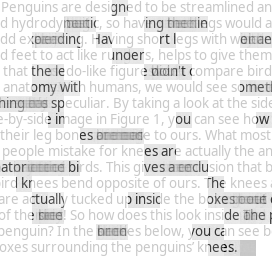}}%
        }%
    \end{subfigure}\hfill
    \begin{subfigure}[]{0.32\textwidth}
        \centering
        {%
        \setlength{\fboxsep}{0pt}%
        \setlength{\fboxrule}{1pt}%
        \fbox{\includegraphics[width=\textwidth]{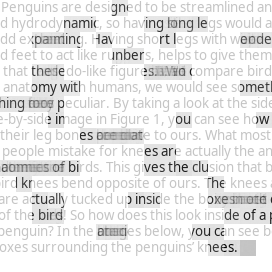}}%
        }%
    \end{subfigure}\hfill
    \begin{subfigure}[]{0.32\textwidth}
        \centering
        {%
        \setlength{\fboxsep}{0pt}%
        \setlength{\fboxrule}{1pt}%
        \fbox{\includegraphics[width=\textwidth]{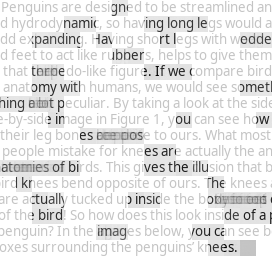}}%
        }%
    \end{subfigure}
    \vspace{0.5em}
    \begin{subfigure}[]{0.32\textwidth}
        \centering
        {%
        \setlength{\fboxsep}{0pt}%
        \setlength{\fboxrule}{1pt}%
        \fbox{\includegraphics[width=\textwidth]{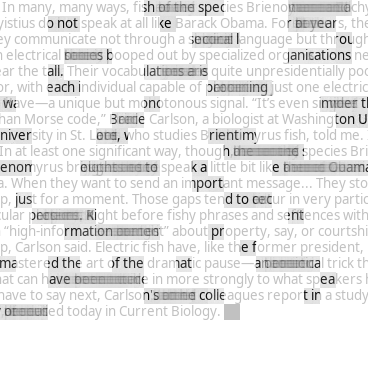}}%
        }%
        \caption{100k steps}
    \end{subfigure}\hfill
    \begin{subfigure}[]{0.32\textwidth}
        \centering
        {%
        \setlength{\fboxsep}{0pt}%
        \setlength{\fboxrule}{1pt}%
        \fbox{\includegraphics[width=\textwidth]{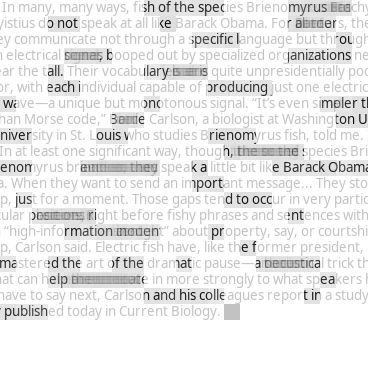}}%
        }%
        \caption{500k steps}
    \end{subfigure}\hfill
    \begin{subfigure}[]{0.32\textwidth}
        \centering
        {%
        \setlength{\fboxsep}{0pt}%
        \setlength{\fboxrule}{1pt}%
        \fbox{\includegraphics[width=\textwidth]{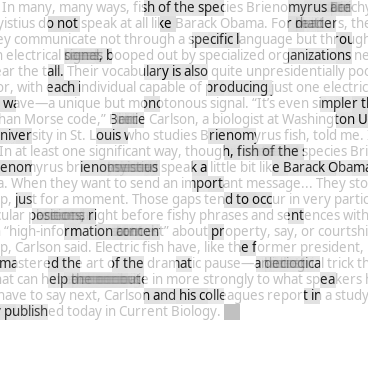}}%
        }%
        \caption{1M steps}
    \end{subfigure}
    \caption{\model image reconstructions after 100k, 500k, and 1M steps of pretraining. We overlay the masked original image with the model's predictions. Images are wrapped into squares and resized for visualization purposes only. The texts were not part of the training data. We see that the fully trained \model (1M) predicts masked spans more clearly and accurately. For longer spans with a larger possible prediction space, multiple predictions may appear together creating blurred text.}
    \label{fig:pretraining_dynamics}
\end{figure*}

Reconstructions of three sources of text\footnote{\url{https://www.nationalpeanutboard.org/peanut-info/our-message.htm}} \footnote{\url{https://www.penguinsinternational.org/2019/07/10/do-penguins-have-knees-and-other-frequently-asked-questions/}} \footnote{\url{https://www.theatlantic.com/science/archive/2021/05/electric-fish-pause/618993/}} after 100K, 500K and 1M pretraining steps. The figure also shows how \model (visually) expresses uncertainty, e.g.\ for reconstructions of long spans where the space of possible outputs is much larger than for short spans, and how it captures long-range dependencies. In the third row, we can for instance see that \model uses context from the beginning of a sequence (\emph{Barack Obama}) to correctly fill in a gap later in the sequence, and vice-versa (\emph{Brienomyrus}).
\section{Code}
\label{app:implementation}

\model is implemented in PyTorch \citep{DBLP:conf/nips/PaszkeGMLBCKLGA19} and built on HuggingFace transformers \citep{wolf-etal-2020-transformers}. We make our code available at \url{https://github.com/xplip/pixel}. Our pretrained \model model, including a large number of intermediate checkpoints, is available at \url{https://huggingface.co/Team-PIXEL/pixel-base} and our finetuned models, including multiple seeds each, are available through the model hub.

\section{Text Renderer Details}
\label{app:text_renderer}

\paragraph{Rendering backend} We experimented with different text rendering backends. Following \citet{salesky-etal-2021-robust}, our first implementation was based on PyGame,\footnote{\url{https://www.pygame.org/}} which \model was also pretrained with. Later on, we switched to a backend based on Pango \citep{taylor-2004-pango} and Cairographics,\footnote{\url{https://www.cairographics.org/}} which has native support for complex text layouts, making it possible to specify fallback fonts, and has faster rendering speed. Without fallback fonts, we would be limited to a maximum number of $2^{16} - 1$ glyphs that can fit into a single OpenType or TrueType font file due to a technical limitation.\footnote{See \url{https://en.wikipedia.org/wiki/Unicode_font} for an explanation.} By leveraging fallback fonts, we can theoretically cover all Unicode codepoints, including emojis. 

\paragraph{Fonts} We rely on the Google Noto Sans fonts collection,\footnote{\url{https://fonts.google.com/noto}} which covers the majority of Unicode codepoints and is actively growing.\footnote{See \url{https://notofonts.github.io/overview/} for an overview of Noto's Unicode coverage.}. Note, however, that \model is compatible with any font and can therefore encode anything that can be typeset on a computer screen. We used a font size of 8 at 120 DPI for pretraining with PyGame, which was selected manually to fit most scripts into a rendered height of 16px. It can, however, also be adjusted at finetuning time. For finetuning with PangoCairo, we use a font size of $8 \cdot (120/72) \approx 13.33$ which yields roughly the same outputs as the PyGame renderer. Due to how glyphs are shaped by the two backends, the outputs of the two renderers do not \emph{exactly} match. Because we did not employ data augmentation to make \model robust to such changes in font size,  we recommend using the PyGame renderer it was pretrained with for \emph{zero-shot} applications with \model. When finetuning, this minor mismatch in rendering outputs is easily overcome by \model, so we generally recommend using the PangoCairo renderer.

\paragraph{Characters versus glyphs} For extractive QA, it is necessary to obtain a mapping between the characters in the context paragraph and where they appear on the rendered image. Obtaining this mapping is not straightforward due to how text is rendered. The \emph{shaping} step in the rendering pipeline converts characters into glyphs.\footnote{See \url{https://docs.gtk.org/Pango/pango_rendering.html} for an overview of the rendering pipeline.} In ligatures, as common for instance in Arabic, a glyph is composed of multiple characters. Likewise, an emoji often consists of a base codepoint and a modifier codepoint (e.g.\ to change the emoji skin colour) which are represented by a single glyph. For accents, on the other hand, one character might yield multiple glyphs.\footnote{\url{https://docs.gtk.org/Pango/pango_fonts.html\#glyphs}} In practice, the renderer therefore uses grapheme clusters, whose logical boundaries in the rendered image we can map to the input characters.\footnote{\url{https://unicode.org/reports/tr29/\#Grapheme\_Cluster\_Boundaries}} For simplicity, we assign each codepoint of a grapheme cluster to the logical horizontal offset at which the cluster starts on the rendered image. Future work may investigate alternative mapping strategies.

\paragraph{RGB rendering} \model supports RGB rendering which may be useful to accurately represent colour emoji and for multimodal applications in the future. However, 24-bit RGB rendering is slightly slower than 8-bit grayscale rendering (see Table~\ref{tab:speed_comparison} below) for text written in Latin script, which is why we made RGB rendering an optional setting. In our pretraining and finetuning experiments we rendered text in grayscale, and we generally recommend doing so when not working with coloured inputs.

\paragraph{Right-to-left scripts} \model's renderer natively supports right-to-left (RTL) writing. In the default setting, the base text direction (which for instance determines on which side of a sentence punctuation marks are placed) is inferred automatically by the rendering backend based on the first ``strong directional'' character in a given paragraph.\footnote{See \url{https://unicode.org/reports/tr9/} for an overview of the Unicode bidi algorithm.} The mirroring of RTL characters is also handled automatically according to their Unicode bidi attributes.
Optionally, the base text direction can be set manually, which is useful when working on monolingual data, e.g.\ in Arabic or Hebrew, as the renderer does not have to go through the direction check. In \S \ref{app:limitations}, we describe limitations of how we currently handle RTL writing.

\begin{figure*}[ht]
  \centering
  \includegraphics[width=0.75\textwidth]{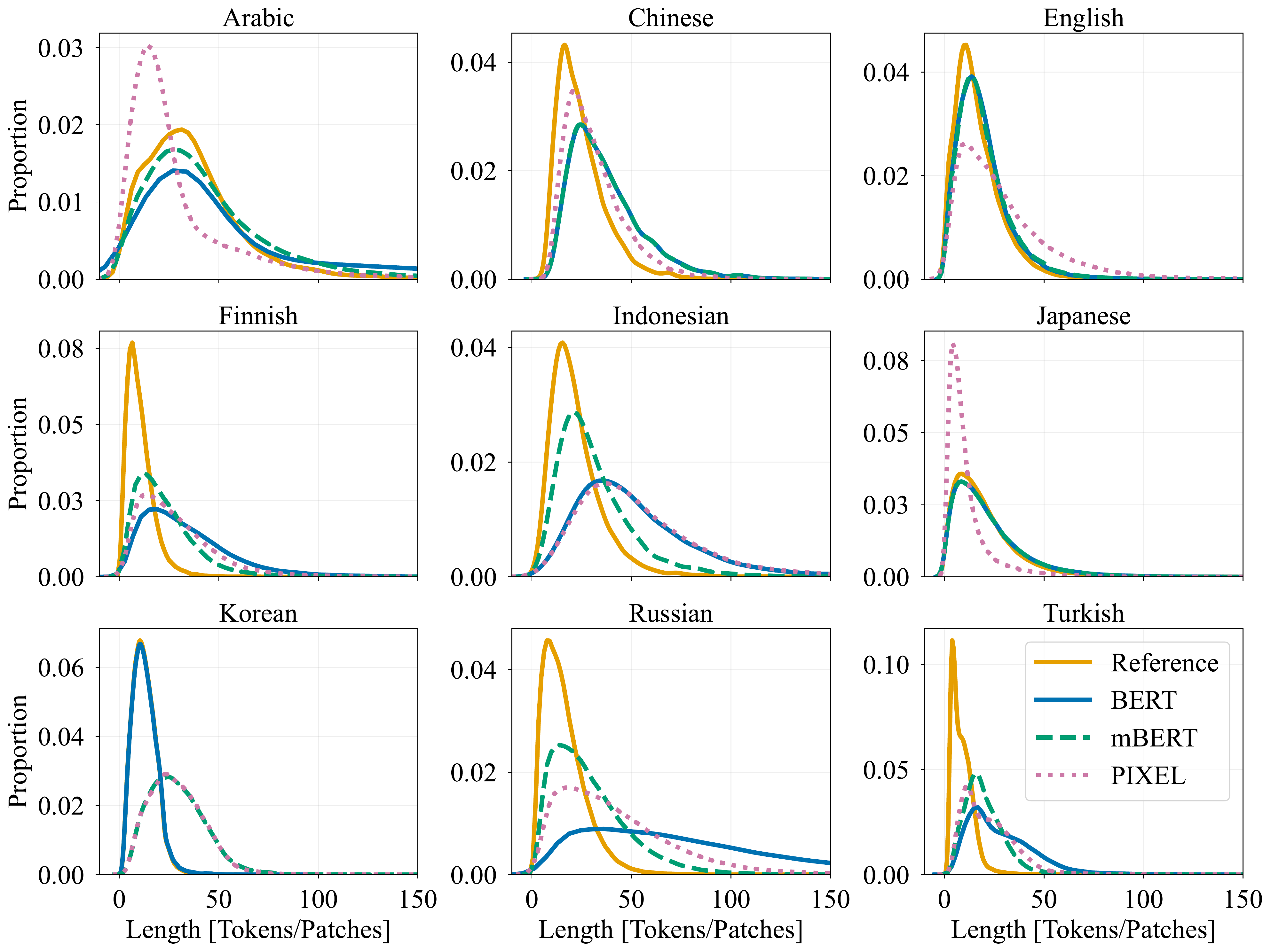}%
  \caption{Distributions of sentence lengths from monolingual UD corpora after tokenizing by \textsc{bert} and \textsc{mbert} and rendering by \model, compared to the reference by UD treebank annotators.}
  \label{fig:ud_sent_len}
\end{figure*}

\begin{table*}[h]
\centering
\resizebox{0.48\textwidth}{!}{%
\begin{tabular}{@{}lccc@{}}
\toprule
Processor & Batched & \multicolumn{2}{c}{Throughput [ex / s]} \\
 &  & ~~\textsc{eng} & ~~\textsc{zho} \\ \midrule
Renderer (Grayscale)   & \xmark  & ~~3944.1 & ~~6309.0            \\
Renderer (RGB)         & \xmark  & ~~3615.1 & ~~6849.5            \\
\midrule
\multirow{2}{*}{Tokenizer (Rust)}   & \cmark  & \textbf{19128.9} & \textbf{18550.5}  \\
                                         & \xmark  & ~~4782.9 & ~~5684.4           \\
\midrule
\multirow{2}{*}{Tokenizer (Python)} & \cmark  & ~~1286.6 &  ~~2637.1          \\
                                         & \xmark  &   ~~1286.8  &  ~~2580.9             \\ \bottomrule
\end{tabular}%
}
\caption{Throughput comparison between \model's PangoCairo renderer and the fast and slow \textsc{bert} tokenizers, implemented in Rust and Python respectively, from the HuggingFace tokenizers library. We estimate throughput, measured in examples per second, by how long it takes to process 1M lines of English (\textsc{eng}) and Chinese (\textsc{zho}) Wikipedia text on the same desktop workstation (AMD Ryzen 9 3900X 12-core CPU). We distinguish between tokenizing all lines individually (Batched = \xmark) and as one single batch (\cmark).
}

\label{tab:speed_comparison}
\end{table*}

\paragraph{Efficiency analysis}

We briefly analyze the text processing (rendering versus tokenizing) efficiency in terms of a) length of the processed sequence, which has a direct effect on GPU memory consumption and the time it takes to compute forward and backward passes, and b) processing throughput.

For a), we follow \cite{rust-etal-2021-good} and process the training and validation splits of all available UD v2.10 treebanks in various languages with the \model renderer and the tokenizers of \textsc{bert} and \textsc{mbert}. We plot the resulting sentence length distributions in Figure~\ref{fig:ud_sent_len}, including a comparison with the reference segmentations from the UD annotators. For English text, the \model renderer is slightly less efficient, i.e., it produces slightly longer sequences on average than the tokenizers. For other languages with Latin script, e.g. Finnish and Turkish, the renderer is more efficient than the \textsc{bert} tokenizer, albeit slightly less efficient than the \textsc{mbert} tokenizer. For non-Latin scripts such as Arabic and Japanese, we see that the renderer can be a lot more efficient than both tokenizers. The English \textsc{bert} tokenizer is technically fairly space-efficient for non-Latin scripts but this is misleading because it largely produces {\footnotesize{[\texttt{UNK}]}}s (recall right side of Table~\ref{res:syntactic_task_results}) and each {\footnotesize{[\texttt{UNK}]}} is a single token; the functionality of the \textsc{bert} model on a sequence of {\footnotesize{[\texttt{UNK}]}} is strongly compromised.

For b), we compare the processing throughput of HuggingFace's \textsc{bert} tokenizers and our \model renderer in Table~\ref{tab:speed_comparison}. We find that the Rust-based \textsc{bert} tokenizer with batch processing achieves the highest throughput by leveraging parallelization. When not using batch processing, it is comparable in throughput with \model's renderer, i.e. depending on the language or script, rendering can be slightly slower (\textsc{eng}) or faster (\textsc{zho}) than tokenizing. Since the rendering backend (PangoCairo) is implemented in C, we expect to achieve similar gains in rendering throughput by also leveraging parallelization for batch processing (in contrast to the Python-based tokenizer which is limited by Python's global interpreter lock (GIL)). We plan to implement batch rendering functionality in the future.

\section{Architecture \& Pretraining Details}
\label{app:architecture_details}

\begin{table}[H]
\centering
\resizebox{0.75\textwidth}{!}{%
\begin{tabular}{@{}ll@{}}
\toprule
\textsc{parameter}            & \textsc{value} \\ \midrule
Image size                    & (16, 8464, 3)  \\
Patch size $P$                   & 16             \\
Encoder hidden size $D_{\text{enc}}$ & 768            \\
Encoder intermediate size     & 3072           \\
Encoder num attention heads   & 12             \\
Encoder num layers $L$        & 12             \\
Decoder hidden size $D_{\text{dec}}$ & 512            \\
Decoder intermediate size     & 2048           \\
Decoder num attention heads   & 16             \\
Decoder num layers $K$        & 8              \\
Layer norm $\varepsilon$ \citep{DBLP:journals/corr/BaKH16}       & \num{1e-12}    \\
Span masking ratio $R$                & 0.25           \\
Span masking max length $S$ & 6 \\
Span masking cumulative weights $W$ & $\{0.2, 0.4, 0.6, 0.8, 0.9, 1\}$ \\
Span masking spacing & Dynamic \\
Dropout probability           & 0.1            \\ 
Hidden activation & GeLU \citep{DBLP:journals/corr/HendrycksG16} \\
Optimizer              & AdamW \citep{loshchilov2018decoupled, kingma-ba-2015-adam}                \\
Adam $\beta$     & (0.9, 0.999) \\
Adam $\varepsilon$       & \num{1e-8}     \\
Weight decay               & 0.05           \\
Peak learning rate         & \num{1.5e-4}   \\
Learning rate schedule & Cosine Decay \citep{DBLP:conf/iclr/LoshchilovH17}                         \\
Minimum learning rate      & \num{1e-5}     \\
Learning rate warmup ratio & 0.05           \\
Training steps             & 1M             \\
Batch size                 & 256       \\ \bottomrule
\end{tabular}%
}
\caption{\model pretraining settings}
\label{tab:pretraining_settings}

\end{table}

\begin{figure*}[t]
    \centering
    \includegraphics[width=\textwidth]{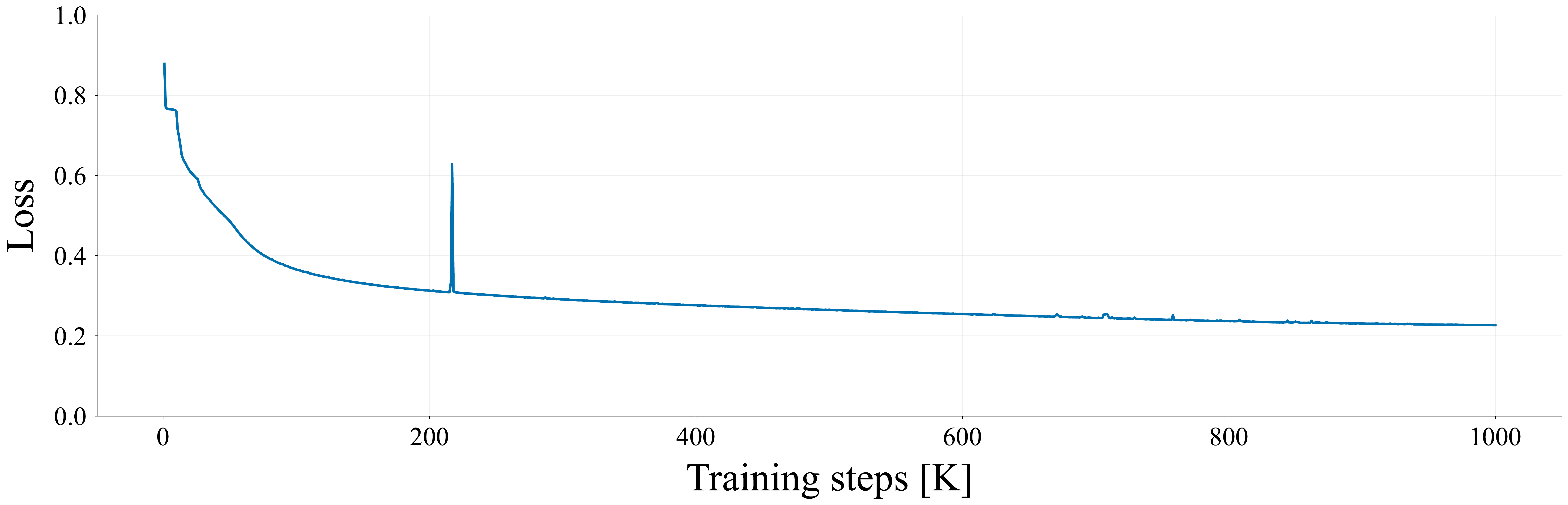}
    \caption{\model pretraining loss curve}
    \label{fig:training_loss}
\end{figure*}

\paragraph{Patch Embeddings} \model reshapes each image $\bm{x}$ into a sequence of $N=W/P$ non-overlapping flattened 2D patches ${\bm{x}_f\in \mathbb{R}^{N \times (P^2C)}}$, where $P=16$ is the patch size, and linearly projects them via ${\bm{E} \in \mathbb{R}^{(P^2C) \times D_{\text{enc}}}}$ to obtain patch embeddings $\bm{x}_p=(\bm{x}_f\bm{E}) \in \mathbb{R}^{N \times D_{\text{enc}}}$ with encoder hidden size ${D_{\text{enc}}=P^2C=768}$.\footnote{This is equivalent to projecting each rendered image $\bm{x} \in \mathbb{R}^{H \times W \times C}$ via a 2D-convolutional layer with $C$ input channels and $D_{\text{enc}}$ output channels and kernel size and stride both equal to the patch size $P$, which we do in practice.} Afterwards, fixed sinusoidal position embeddings $\bm{E}_{\text{pos}} \in \mathbb{R}^{(N + 1) \times D_{\text{enc}}}$ are added, leaving out the position vector in position~0 for a classification (CLS) embedding later: $\bm{\tilde{x}}_p=\bm{x}_p + [\bm{E}_{\text{pos}}^1, \ldots, \bm{E}_{\text{pos}}^{(N+1)}]$.

\paragraph{Span Masking} \model then masks out ${R=25\%}$ of the $N=529$ embedded patches via span masking with max span length $S=6$ and cumulative span weights $W=\{0.2,0.4,0.6,0.8,0.9,1\}$, i.e. $\mathbb{E}(s)=3.1$, as outlined in Algorithm~\ref{algo:span_masking}. Applying the mask $\mathcal{M}$, we obtain the unmasked patches ${\bm{\tilde{x}}_{\text{vis}} = \{ \bm{\tilde{x}}_p^{i} : i \notin \mathcal{M}\}_{i=0}^N}$.

\paragraph{Encoder} Following ViT-MAE \citep{he-etal-2022-mae}, the \model encoder only operates on unmasked patches (i.e., $\approx$ 396 patches at 25\% masking) and a special CLS embedding with its positional encoding ${\bm{c} = \bm{x}_{[\text{cls}]} + \bm{E}_{\text{pos}}^0 \in \mathbb{R}^{1 \times D_{\text{enc}}}}$ is prepended to the sequence: $\bm{h}_0 = [\bm{c}, \bm{\tilde{x}}_{\text{vis}}] \in \mathbb{R}^{(1 + \floor{R\cdot N}) \times D_{\text{enc}}}$.\footnote{In pretraining, no loss is computed for the CLS embedding but it can optionally be used when finetuning \model for sequence-level downstream tasks.} Let $\{\bm{h}_i\}_{i=1}^L$ be the encoder hidden states after each of the $L=12$ encoder transformer layers, and $\bm{h}_0$ denotes the input sequence. The outputs of each transformer layer are computed as detailed in \citep{DBLP:conf/nips/VaswaniSPUJGKP17}, \footnote{Note that encoder and decoder do not attend to the blank (padding) patches that appear after the {\footnotesize \texttt{EOS}} patch.} and the last layer's output $\bm{h}_L \in \mathbb{R}^{(1 + \floor{R\cdot N}) \times D_{\text{enc}}}$ is passed to the decoder.

\paragraph{Decoder} The \model decoder first projects the encoder outputs via $\bm{E}_{\text{dec}} \in \mathbb{R}^{D_{\text{enc}} \times D_{\text{dec}}}$ to obtain decoder embeddings $\bm{x}_d = \bm{h}_L\bm{E}_{\text{dec}} \in \mathbb{R}^{(1 + \floor{R\cdot N}) \times D_{\text{dec}}}$, where $D_{\text{dec}}=512$.
Next, mask embeddings $x_{[\text{mask}]} \in \mathbb{R}^{1\times D_{\text{dec}}}$ are inserted at the masked-out positions and fixed sinusoidal position embeddings are added to obtain
$\bm{d}_0 = [(\bm{x_d} \cup \{ x_{[\text{mask}]}: i \in \mathcal{M}\}_{i=0}^N) + \bm{E}_{\text{pos}}] \in \mathbb{R}^{(N + 1) \times D_{\text{\text{dec}}}}$. $\{\bm{d}_i\}_{i=1}^K$ are the decoder hidden states after each of the $K=8$ decoder transformer layers, computed in the same way as the encoder hidden states, and $\bm{d}_0$ denotes the input sequence. There is no encoder-decoder cross-attention. The decoder output $\bm{d}_K \in \mathbb{R}^{(N + 1) \times D_{\text{\text{dec}}}}$ is projected via $\bm{O} \in \mathbb{R}^{D_{\text{\text{dec}}} \times (P^2C)}$ to obtain patch-wise logits $\bm{o} = (\bm{d}_K \bm{O}) \in \mathbb{R}^{(N + 1) \times (P^2C)}$. Finally, the CLS logits are removed and a normalized mean squared error (MSE) pixel reconstruction loss is computed: $\bm{\mathcal{L}}_{\text{normpix}} = \frac{1}{ \lvert Q \rvert } \sum_{i \in Q}{\lvert \text{normalize}(\bm{x}_f^i) - \bm{o}^i\rvert^2}$ with $i$ denoting the indices in the set of \emph{masked, non-blank (text)} patches $Q = \{i: i \in (\mathcal{M} \cap \mathcal{T})\}_{i=0}^N$ and $\text{normalize}(\cdot)$ dividing the difference between the target patch and its mean by its standard deviation.

\section{Finetuning Details}
\label{app:finetuning_details}
Table~\ref{tab:language_overview} gives an overview of all languages used in our finetuning experiments, Table~\ref{tab:dataset_overview} links to our finetuning datasets, and Table~\ref{tab:treebank_overview} lists the UD treebanks we used.

We list our finetuning recipes in Table~\ref{tab:finetuning_details1} for \textsc{pos} tagging, dependency parsing, \textsc{ner}, \textsc{qa}, and \textsc{xnli} and in Table~\ref{tab:finetuning_details2} for the GLUE tasks. Due to compute limitations we did not run comprehensive hyperparameter sweeps. Instead, we relied on sensible priors from finetuning \textsc{bert} and made slight modifications as needed. In most cases, hyperparameters that work well for \textsc{bert} also work well for \model. For some of the semantic tasks, in particular \textsc{nli} and \textsc{sst-}{\footnotesize2}, we found that some random initializations did not converge. In those cases, minor tweaks to the learning rate or increasing the batch size usually helped. For \textsc{glue}, we found that \model performed slightly better on some tasks with the PangoCairo renderer, whereas for others, using the PyGame renderer (which \model was pretrained with) was more stable. We plan to further optimize the training recipes and study \model's convergence behaviour in the future. 

For word-level tasks, we add padding in order to render each word at the start of a new image patch and so create a bijective mapping between words and patches.
Doing so assumes that word boundaries are available. We note that subword-based and character-based models also make this assumption. In \textsc{bert}, for instance, word-level tasks are formulated such that a word's label is assigned to its first subword token, requiring word boundaries. During training, continuation tokens are then masked out when computing the loss. Consequently, predictions for continuation tokens also need to be masked out at inference time, which again requires word boundaries or aggregation strategies that may introduce errors. The same applies to character-based models. For \model, should this assumption be violated, it is still possible to render the text without adding spacing, although the mapping is then no longer bijective as multiple words can overlap on one image patch. In such cases, assigning the prediction for a patch to either word can cause loss of information. Although in practice this approach does not necessarily affect performance negatively, future work will investigate alternative approaches.

\begin{table*}[ht]
\centering
\resizebox{0.6\textwidth}{!}{%
\begin{tabular}{@{}llll@{}}
\toprule
Language     & ISO 639-3    & Language Family & Script       \\ \midrule
Amharic      & \amharic     & Afro-Asiatic    & Ge\fontencoding{T3}\selectfont{\textrevapostrophe}ez        \\
Arabic       & \arabi       & Afro-Asiatic    & Arabic       \\
Bengali      & \bengali     & Indo-European   & Bengali      \\
Bulgarian    & \bulgarian   & Indo-European   & Cyrillic     \\
Chinese      & \chinese     & Sino-Tibetan    & Chinese      \\
Coptic       & \coptic      & Afro-Asiatic    & Coptic       \\
English      & \english     & Indo-European   & Latin        \\
Finnish      & \finnish     & Uralic          & Latin        \\
French       & \french      & Indo-European   & Latin        \\
German       & \german      & Indo-European   & Latin        \\
Greek        & \greek       & Indo-European   & Greek        \\
Hausa        & \hausa       & Afro-Asiatic    & Latin        \\
Hindi        & \hindi       & Indo-European   & Devanagari   \\
Igbo         & \igbo        & Niger-Congo     & Latin        \\
Indonesian   & \indonesian  & Austronesian    & Latin        \\
Japanese     & \japanese    & Japonic         & Japanese     \\
Kinyarwanda  & \kinyarwanda & Niger-Congo     & Latin        \\
Korean       & \korean      & Koreanic        & Korean       \\
Luganda      & \luganda     & Niger-Congo     & Latin        \\
Luo          & \luo         & Nilo-Saharan    & Latin        \\
Naija Pidgin & \naija       & English Creole  & Latin        \\
Russian      & \russian     & Indo-European   & Cyrillic     \\
Spanish      & \spanish     & Indo-European   & Latin        \\
Swahili      & \swahili     & Niger-Congo     & Latin        \\
Tamil        & \tamil       & Dravidian       & Tamil        \\
Telugu       & \telugu      & Dravidian       & Telugu       \\
Thai         & \thai        & Kra-Dai         & Thai         \\
Turkish      & \turkish     & Turkic          & Latin        \\
Urdu         & \urdu        & Indo-European   & Perso-Arabic \\
Vietnamese   & \vietnamese  & Austro-Asiatic  & Latin        \\
Wolof        & \wolof       & Niger-Congo     & Latin        \\
Yorùbá       & \yoruba      & Niger-Congo     & Latin        \\ \bottomrule
\end{tabular}%
}
\caption{Overview of languages used in our experiments.}
\label{tab:language_overview}
\end{table*}

\begin{table*}[t]
\centering
\resizebox{\textwidth}{!}{%
\begin{tabular}{@{}lll@{}}
\toprule
Dataset      & Download Link                                                                         & Reference \\ \midrule
Universal Dependencies 2.10 & \url{https://lindat.mff.cuni.cz/repository/xmlui/handle/11234/1-4758} & \citep{zeman-etal-2022-ud, nivre-etal-2020-universal} \\
MasakhaNER                  & \url{https://github.com/masakhane-io/masakhane-ner/tree/main/data}       &  \citep{adelani-etal-2021-masakhaner} \\
GLUE         & \url{https://huggingface.co/datasets/glue}     & \citep{wang-etal-2018-glue}          \\
TyDiQA-GoldP & \url{https://huggingface.co/datasets/tydiqa} &  \citep{clark-etal-2020-tydi}         \\
SQuADv1.1    & \url{https://huggingface.co/datasets/squad}   &  \citep{rajpurkar-etal-2016-squad}         \\
KorQuAD 1.0                 & \url{https://huggingface.co/datasets/squad\_kor\_v1}                                   & \citep{lim-etal-2019-korquad} \\
JaQuAD                      & \url{https://huggingface.co/datasets/SkelterLabsInc/JaQuAD}                     &  \citep{so2022jaquad} \\
XNLI         & \url{https://huggingface.co/datasets/xnli}     & \citep{conneau-etal-2018-xnli} \\ \bottomrule          
\end{tabular}%
}
\caption{Links and references to the datasets we used in our finetuning experiments.}
\label{tab:dataset_overview}
\end{table*}

\begin{table*}[ht]
\centering
\resizebox{0.6\textwidth}{!}{%
\begin{tabular}{@{}llrl@{}}
\toprule
Language    & Treebank           & \#Sentences  & Reference \\ \midrule
\english    & English-EWT        & 16621        &  \citet{silveira-etal-2014-gold}\\
\arabi      & Arabic-PADT        & 7664         & \citet{hajivc2009prague}\\
\coptic     & Coptic-Scriptorium & 2011         & \citet{zeldes-abrams-2018-coptic}\\
\hindi      & Hindi-HDTB         & 16647        & \citet{Palmer2009HindiSA}\\
\japanese   & Japanese-GSD       & 8100         & \citet{asahara-etal-2018-universal}\\
\korean     & Korean-GSD         & 6339         & \citet{chun-etal-2018-building}\\
\tamil      & Tamil-TTB          & 600          & \citet{ramasamy-zabokrtsky-2012-prague}\\
\vietnamese & Vietnamese-VTB     & 3000         & \citet{nguyen-etal-2009-building}\\
\chinese    & Chinese-GSD        & 4997         & \citet{shen-etal-2016-chinese}\\ \bottomrule
\end{tabular}%
}
\caption{Overview of the Universal Dependencies v2.10 \citep{zeman-etal-2022-ud, nivre-etal-2020-universal} treebanks used in our POS tagging and dependency parsing experiments with the number of sentences in their respective training splits. As mentioned in \S\ref{sec:tasks_languages}, these treebanks were chosen with typological and script diversity in mind.}
\label{tab:treebank_overview}
\end{table*}

\begin{table}[ht]
\centering
\resizebox{0.65\textwidth}{!}{%
\begin{tabular}{@{}lccccc@{}}
\toprule
\textsc{parameter} & \textsc{pos} & \textsc{dp}                  & \textsc{ner} & \textsc{qa}                             & \textsc{xnli} \\ \midrule
Rendering backend                   & \multicolumn{5}{c}{PangoCairo}        \\
Classification head pooling & --- & --- & --- & --- & CLS \\
Optimizer                  & \multicolumn{5}{c}{AdamW}             \\
Adam $\beta$               & \multicolumn{5}{c}{(0.9, 0.999)}      \\
Adam $\varepsilon$            & \multicolumn{5}{c}{\num{1e-8}}        \\
Weight decay               & \multicolumn{5}{c}{0}                 \\
Learning rate      & \num{5e-5}   & $\{\num{5e-5}, \num{8e-5}\}$ & \num{5e-5}   & $\{\num{3e-5},\num{5e-5}, \num{7e-5}\}$ & \num{2e-5}    \\
Learning rate warmup steps & 100   & 100   & 100   & 100   & 1000  \\
Learning rate schedule     & \multicolumn{5}{c}{Linear decay}      \\
Max sequence length        & 256   & 256   & 196   & 400   & 196   \\
Stride                     & ---   & ---   & ---   & 160   & ---   \\
Batch size                 & 64    & 64    & 64    & 32    & 256   \\
Max steps                  & 15000 & 15000 & 15000 & 20000 & 50000 \\
Early stopping              & \multicolumn{5}{c}{\checkmark}                                      \\
Eval steps                 & 500   & 500   & 500   & 500   & 1000  \\
Dropout probability              & \multicolumn{5}{c}{0.1}               \\ \bottomrule
\end{tabular}%
}
\caption{Finetuning settings for \textsc{pos} tagging, dependency parsing (\textsc{dp}), \textsc{ner}, \textsc{qa}, and \textsc{xnli}. We did not run a comprehensive hyperparameter search due to compute limitations; these settings were manually selected based on a small number of preliminary runs. Maximum performance was often reached well before the specified number of max steps.}
\label{tab:finetuning_details1}
\end{table}

\begin{table}[ht]
\centering
\resizebox{\textwidth}{!}{%
\begin{tabular}{@{}lccccccccc@{}}
\toprule
\textsc{parameter} & \textsc{mnli} & \textsc{qqp} & \textsc{qnli} & \textsc{sst-2} & \textsc{cola} & \textsc{sts-b} & \textsc{mrpc} & \textsc{rte} & \textsc{wnli} \\ \midrule
Rendering backend           & PangoCairo    & PyGame       & PangoCairo    & PyGame         & PyGame        & PyGame         & PyGame        & PyGame       & PyGame        \\
Classification head pooling & \multicolumn{9}{c}{Mean}                                            \\
Optimizer                   & \multicolumn{9}{c}{AdamW}                                           \\
Adam $\beta$                & \multicolumn{9}{c}{(0.9, 0.999)}                                    \\
Adam $\varepsilon$             & \multicolumn{9}{c}{\num{1e-8}}                                      \\
Weight decay                & \multicolumn{9}{c}{0}                                               \\
Learning rate      & \num{3e-5}    & \num{3e-5}   & \num{3e-5}    & \num{3e-5}     & \num{2e-5}    & \num{2e-5}     & \num{3e-5}    & \num{3e-5}   & \num{1e-5}    \\
Learning rate warmup steps  & 100   & 100   & 100   & 100   & 200   & 100   & 100   & 200   & 100 \\
Learning rate schedule      & \multicolumn{9}{c}{Linear decay}                                    \\
Max sequence length         & \multicolumn{9}{c}{256}                                             \\
Batch size                  & 64    & 256   & 64    & 256   & 256   & 64    & 64    & 64    & 256 \\
Max steps                   & 15000 & 15000 & 15000 & 15000 & 15000 & 15000 & 15000 & 15000 & 400 \\
Early stopping              & \multicolumn{9}{c}{\checkmark}                                      \\
Eval interval      & 500 steps     & 500 steps    & 500 steps     & 500 steps      & 100 steps     & 100 steps      & 100 steps     & 250 steps    & 1 epoch       \\
Dropout probability                & \multicolumn{9}{c}{0.1}                                             \\ \bottomrule
\end{tabular}%
}
\caption{Finetuning settings for \textsc{glue} tasks. We did not run a comprehensive hyperparameter search due to compute limitations; these settings were manually selected based on a small number of preliminary runs. Increasing the batch size to 256 and switching to the PyGame renderer helped achieve more consistent convergence behaviour for some tasks. For the smaller datasets (to the right of \textsc{qqp}), maximum performance was reached well before the specified number of max steps.}
\label{tab:finetuning_details2}
\end{table}

\clearpage

\section{Examples of \textit{Zeroé} orthographic attacks}
\label{app:zeroe}
\begin{table}[h]
\vspace{-1em}
\centering
\resizebox{0.49\textwidth}{!}{
\setlength\tabcolsep{10pt} 
\begin{tabular}{ll}
    \toprule
    Attack & Sentence \\ 
    \midrule
    \tc{\textsc{None}} & \tc{Penguins are designed to be streamlined} \\
    \midrule
    \textsc{Confusable} & \includegraphics[width=0.42\textwidth]{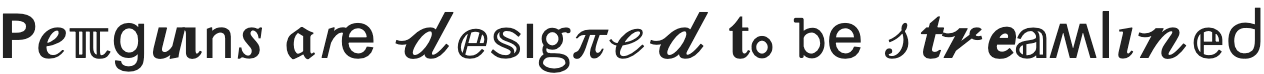} \\ 
    \textsc{Shuffle (inner)} & Pegnuins are dnesiged to be sieatrnmled \\
    \textsc{Shuffle (full)} & ngePnius rae dsgednei to be etimaslernd \\
    \textsc{Disemvowel} & Pngns r dsgnd to be strmlnd \\
    \textsc{Intrude} & Pe`nguins a\{re d)esigned t;o b*e stre\textless{}amlined \\
    \textsc{Keyboard typo} & Penguinz xre dwsigned ro ne streamllned \\
    \textsc{Natural noise} & Penguijs ard design4d ti bd streamlinfd \\
    \textsc{Truncate} & Penguin are designe to be streamline \\
    \textsc{Segmentation} & Penguinsaredesignedtobestreamlined \\
    \textsc{Phonetic} & Pengwains's ar dhiseind te be storimlignd \\
    \bottomrule
\end{tabular}
}
\caption{Examples of low-level orthographic attacks based on the \textit{Zeroé} benchmark.}
\label{tab:zeroe-examples} 
\end{table}

\begin{figure}[ht!]
  \centering
  \includegraphics[width=0.74\textwidth]{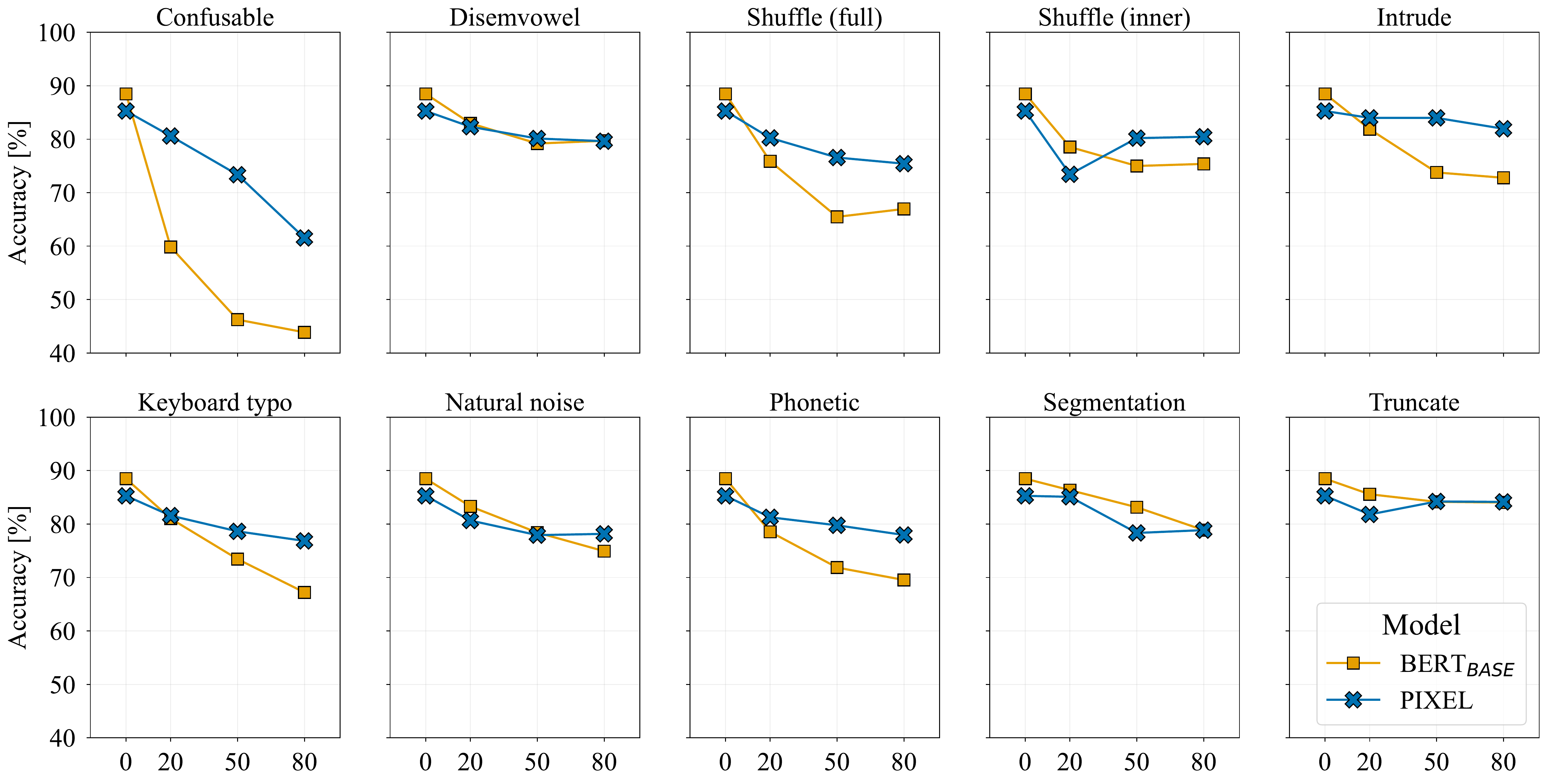}%
  \caption{Test set accuracy for a single run of \model{} and \textsc{bert} across different levels of noise introduced through various orthographic attacks in SNLI. The results show that \model is more robust than \textsc{bert} to most of these attacks.}
  \label{fig:robustness_snli}
\end{figure}

\begin{figure}[ht!]
  \centering
  \includegraphics[width=0.75\textwidth]{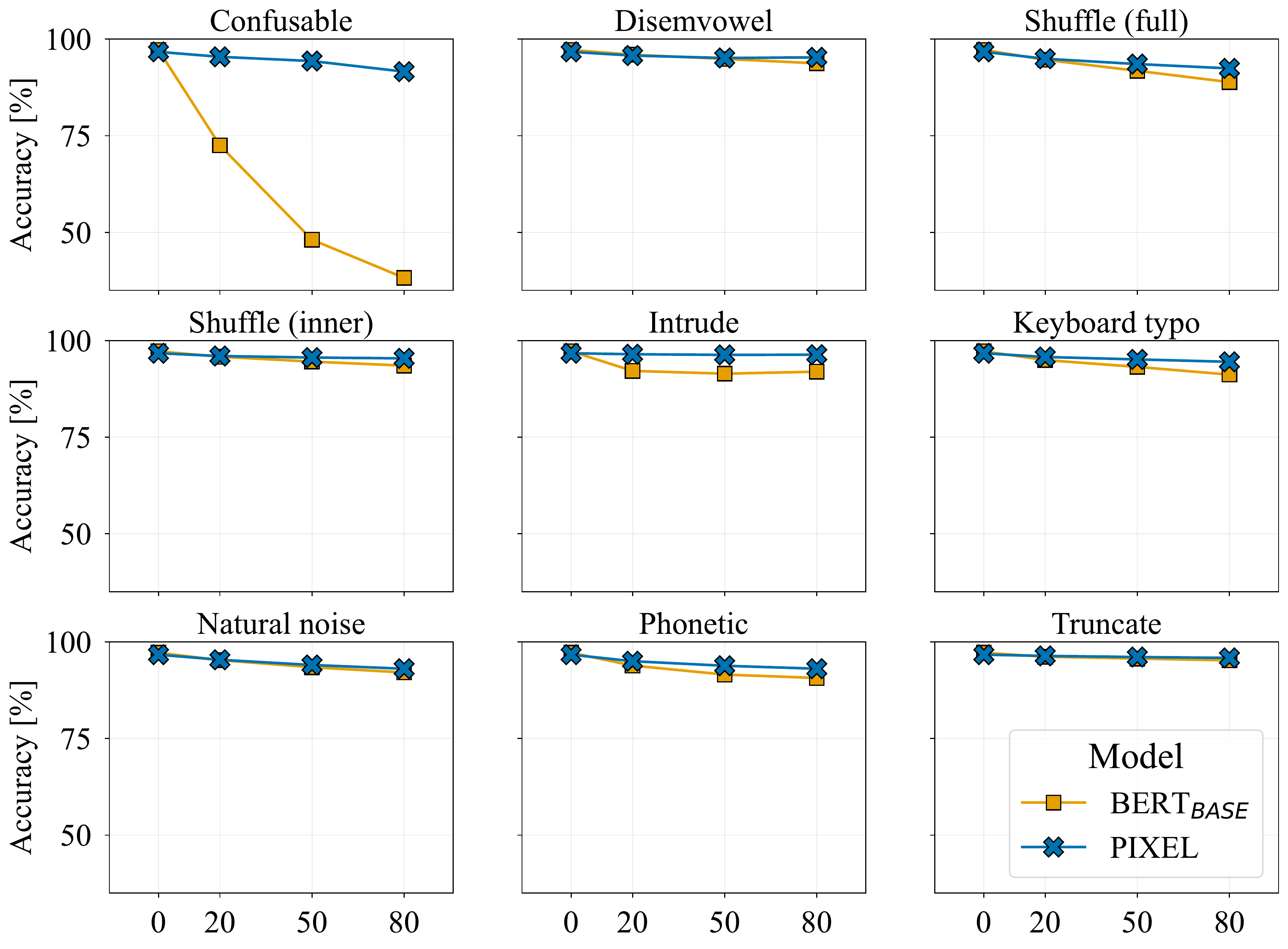}%
  \caption{Test set accuracy for a single run of \model{} and \textsc{bert} across different levels of noise introduced through various orthographic attacks in POS tagging. The results show that \model is more robust than \textsc{bert} to most of these attacks, especially when dealing with visually-confusable character substitutions. \textsc{Segmentation} is not applied to the task of POS tagging, since the joined words would not have a proper tag.}
  \label{fig:robustness_pos}
\end{figure}

\section{Font Transfer Analysis}
\label{app:font_transfer}

In this section, we analyse the adaptation capabilities of \model to new fonts at finetuning time. Specifically, we finetune \model models for POS tagging and dependency parsing on the UD\_English-EWT treebank and sentiment analysis on SST-2, once with a font similar to our  {\footnotesize \texttt{GoNotoCurrent / NotoSans-Regular}} pretraining font, {\footnotesize \texttt{NotoSerif-Regular}}, and once with a font strikingly different from it, {\footnotesize\texttt{JournalDingbats1}}. We compare the three fonts in Table~\ref{tab:font_transfer_examples} below:

\begin{table}[ht]
\centering
\resizebox{0.6\textwidth}{!}{%
\begin{tabular}{ll}
    \toprule
    Font & Rendered Example Sentence \\ 
    \midrule
    {\footnotesize \texttt{GoNotoCurrent}} & \includegraphics[width=0.5\textwidth]{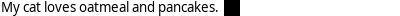} \\ 
    {\footnotesize \texttt{NotoSerif-Regular}} & \includegraphics[width=0.5\textwidth]{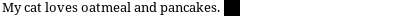} \\ 
    {\footnotesize\texttt{JournalDingbats1}} & \includegraphics[width=0.5\textwidth]{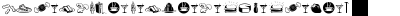} \\
    \bottomrule
\end{tabular}
}
\caption{An example sentence rendered in three different fonts.}
\label{tab:font_transfer_examples} 
\end{table}

\begin{table*}[ht]
\centering
\resizebox{0.6\textwidth}{!}{%
\begin{tabular}{@{}lccc@{}}
\toprule
      & {\footnotesize \texttt{GoNotoCurrent}} & {\footnotesize \texttt{NotoSerif-Regular}} & {\footnotesize\texttt{JournalDingbats1}} \\ \midrule
POS   & 96.7                    & 95.9                        & 93.9                          \\
DP    & 90.6                    &  88.1                           & 81.3                          \\
SST-2 & 89.6                    & 84.2                        & 72.9                          \\ \bottomrule
\end{tabular}%
}
\caption{Results for fine-tuning \model for POS tagging, dependency parsing (DP), and sentiment analysis on SST-2 with three different fonts: the font used in pretraining ({\footnotesize \texttt{GoNotoCurrent}}), a visually similar font ({\footnotesize \texttt{NotoSerif-Regular}}), and a highly dissimilar font ({\footnotesize\texttt{JournalDingbats1}}). We report test accuracy for POS, test LAS for DP, and validation accuracy for SST-2, each averaged over 5 runs.}
\label{tab:font_transfer_results}
\end{table*}

The font transfer results are shown in Table~\ref{tab:font_transfer_results}. We find that \model exhibits fairly high font transfer ability \emph{out-of-the-box}, i.e. without any font or image augmentation strategies employed during pretraining.\footnote{We believe such augmentation strategies would further improve robustness to font variations and leave this experiment to future work. Considering that we have full control over the font when working with NLP text datasets, robustness to font variations was not a primary goal in this work.} In line with our expectations, transfer to a visually similar font ({\footnotesize \texttt{NotoSerif-Regular}}) is easier than to a dissimilar font ({\footnotesize\texttt{JournalDingbats1}}). Nevertheless, \model is able to transfer surprisingly well to the {\footnotesize\texttt{JournalDingbats1}} font, in which every letter is simply mapped to the icon of an object or animal. 

\section{Further analysis}
\label{app:analysis}

\begin{figure*}[ht!]
  \centering
\includegraphics[scale=0.3]{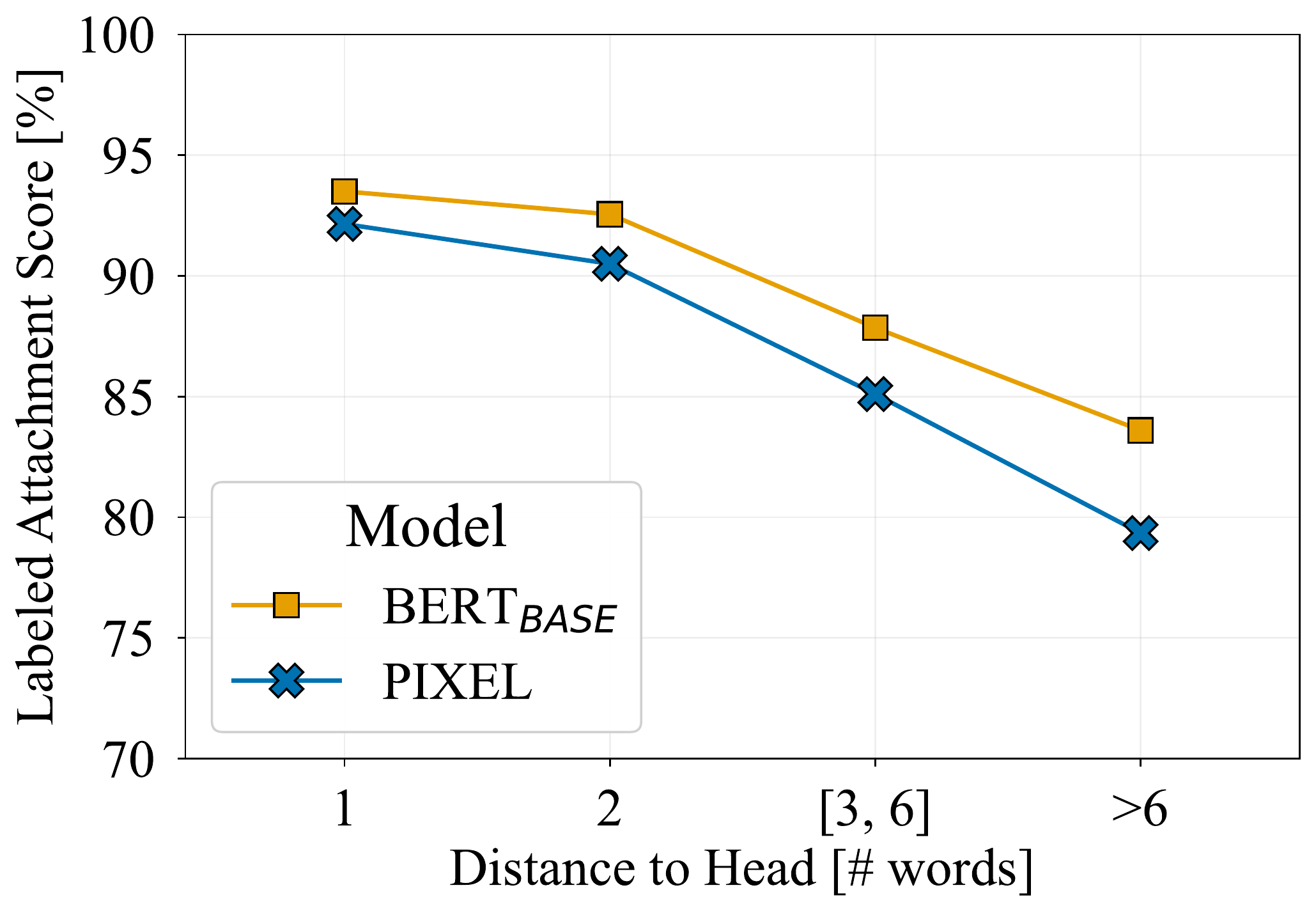} 
\caption{LAS scores (\english) across different dependency lengths averaged over 5 random intitializations of \textsc{bert} and \model. In \english, long syntactic dependencies are more challenging for \model.}
\label{fig:en_dep_len}
\vspace{2em}
\end{figure*}

To investigate where \model currently lags behind \textsc{bert}, we analyse the impact that dependency length has on both models in dependency parsing in \english. 
We can see in Figure~\ref{fig:en_dep_len} that the LAS gap between \textsc{bert} and \model increases with longer dependencies, indicating that \model struggles slightly more with long syntactic dependencies.

\begin{table*}[htp]
\centering
\resizebox{\textwidth}{!}{%
\begin{tabular}{lccccccccccccccccc}
\toprule
   &
  \textsc{\#l} &
  $\theta$ &
  \english &
  \arabi &
  \bulgarian &
  \german &
  \greek &
  \french &
  \hindi &
  \russian &
  \spanish &
  \swahili &
  \thai &
  \turkish &
  \urdu &
  \vietnamese &
  \chinese \\ \midrule
\tc{\textsc{mbert}} & \tc{104} & \tc{179M} & \tc{83.3}  & \tc{73.2}  & \tc{77.9}  & \tc{78.1}  & \tc{75.8}  & \tc{78.5}  & \tc{70.1}  & \tc{76.5}  & \tc{79.7}  & \tc{67.2} & \tc{67.7} & \tc{73.3} & \tc{66.1}  & \tc{77.2}  & \tc{77.7}           \\ \midrule
\textsc{bert} &
  \quad 1 &
  110M &
  \textbf{83.7} &
  \textbf{64.8} &
  \textbf{69.1} &
  \textbf{70.4} &
  \textbf{67.7} &
  \textbf{72.4} &
  \textbf{59.2} &
  \textbf{66.4} &
  \textbf{72.4} &
  \textbf{62.2} &
  35.7 &
  \textbf{66.3} &
  \textbf{54.5} &
  \textbf{67.6} &
  46.2 \\
\model         & \quad 1   & ~~86M  & 77.2 & 58.9 & 66.5 & 68.0 & 64.9 & 69.4 & 57.8 & 63.4 & 70.3 & 60.8 & \textbf{50.2} & 64.0 & 54.1 & 64.8 & \textbf{52.0} \\ \bottomrule
\end{tabular}%
}
\caption{Results for \model and \textsc{bert} finetuned on XNLI in the \emph{translate-train-all} setting where we train on the joint training data in all 15 languages, originally translated from \english by \cite{conneau-etal-2018-xnli}. We report test set accuracy averaged over 5 runs each. Despite the relatively large performance gap in favor of \textsc{bert} in \english (which is in line with the GLUE results in Table~\ref{res:glue_results}), the gap is much smaller for other languages, particularly those not using the Latin writing system. \model is overall more consistent across scripts, outperforming \textsc{bert} in \thai and \chinese.}
\label{res:xnli}
\end{table*}

\section{Limitations}
\label{app:limitations}
This paper introduces a new approach to processing written language as images, which removes the need for a finite vocabulary, providing a solution to the \emph{vocabulary bottleneck}. While our results show that \model is a promising approach in this direction, this is only the first step.
Here, we highlight current limitations and avenues for future work for pixel-based models:
\begin{itemize}
\itemsep-0.04em
\item \model is pretrained on predominantly English text written in the Latin script. The choice of English is driven by the scientific goal of comparing against a widely used model (English \textsc{bert}) but English may not be the best source language for cross-lingual transfer \citep{turc2021revisiting,blevins2022analyzing}.
We expect that \model{} trained on typologically diverse languages in multiple scripts would considerably surpass the cross-script and cross-lingual transferability of English-only \model{} but this remains to be verified, and training a model on large amounts of data will require large computational resources. 
\item \model currently seems to be less sample-efficient than subword-based PLMs. \model{} excels at syntactic tasks after being pretrained for the same number of steps/datapoints as \textsc{bert} (a challenging setup within an academic budget), but still lags behind in semantic processing.
As a consequence, it also requires more training steps  than \textsc{bert} to converge during finetuning.
Closing this gap might involve longer pretraining with additional (long-dependency) objectives.
\item There are challenges to be addressed when working with languages written right-to-left. \model{} currently processes sentences in such languages from the end to the beginning which may lead to learning inadequate features for sentence separation and position embeddings.
\item \model cannot be used for language generation tasks because it is not possible to produce discrete words from the pretrained decoder.

\item Rendering text as images requires more disk space than reading text from a file. This can be alleviated by caching the dataset in a compressed format, or rendering the images on-the-fly. Rendering images on-the-fly will create additional overhead when training for multiple epochs.

\end{itemize}

\end{document}